\documentclass[10pt,twocolumn,letterpaper]{article}

\PassOptionsToPackage{table,usenames,dvipsnames}{xcolor}

\usepackage{cvpr}              %

\newcommand{\red}[1]{{\color{red}#1}}

\definecolor{cvprblue}{rgb}{0.21,0.49,0.74}
\usepackage[pagebackref,breaklinks,colorlinks,allcolors=cvprblue]{hyperref}

\usepackage[accsupp]{axessibility}  %
\usepackage{graphicx}
\usepackage{amsmath}
\usepackage{amssymb}
\usepackage[utf8]{inputenc} %
\usepackage[T1]{fontenc}    %
\usepackage{url}            %
\usepackage{booktabs}       %
\usepackage{amsfonts}       %
\usepackage{nicefrac}       %
\usepackage{microtype}      %
\usepackage{colortbl}
\usepackage{makecell}
\usepackage{multirow}
\usepackage{subcaption}
\usepackage{float}
\usepackage{pifont}
\usepackage{arydshln} %

\usepackage{times}
\usepackage{epsfig}
\usepackage{colortbl}
\usepackage{booktabs} %
\usepackage{multirow}
\usepackage{makecell}
\usepackage{float}
\usepackage{enumitem} %
\usepackage{arydshln}
\usepackage{bbding}
\usepackage{animate}
\usepackage{grffile}
\usepackage{siunitx} %
\usepackage{tabularx}
\usepackage{bbm}
\usepackage{caption} 
\usepackage{subcaption}
\usepackage{pifont}     %
\usepackage{amsfonts}
\usepackage{tikz}
\usepackage{gensymb}

\newcommand{\project}{WildGS-SLAM}

\colorlet{colorFst}{Green!25}       %
\colorlet{colorSnd}{SpringGreen!45} %
\colorlet{colorTrd}{Yellow!30}      %
\colorlet{colorLow}{darkgray!30}    %
\newcommand{\fs}{\cellcolor{colorFst}\bf}   %
\newcommand{\nd}{\cellcolor{colorSnd}}      %
\newcommand{\rd}{\cellcolor{colorTrd}}      %

\def\paperID{12060} %
\def\confName{CVPR}
\def\confYear{2025}

\title{WildGS-SLAM: Monocular Gaussian Splatting SLAM in Dynamic Environments}

\author{
Jianhao Zheng$^{1*}$ \quad 
Zihan Zhu$^{2}$${\footnotemark}$ \quad 
Valentin Bieri$^{2}$ \quad 
Marc Pollefeys$^{2,3}$ \quad 
Songyou Peng$^{2}$ \quad 
Iro Armeni$^{1}$ \\
$^{1}$Stanford University\qquad
$^{2}$ETH Z\"urich \qquad
$^{3}$Microsoft \qquad
\\
\href{https://wildgs-slam.github.io/}{wildgs-slam.github.io}
}

\newcommand{\bomega}{\boldsymbol{\omega}}

\newcommand{\figref}[1]{Fig.~\ref{#1}}
\newcommand{\secref}[1]{Sec.~\ref{#1}}

\newcommand{\eqnref}[1]{Eq.~\eqref{#1}}
\newcommand{\tabref}[1]{Table~\ref{#1}}

\makeatletter
\DeclareRobustCommand\onedot{\futurelet\@let@token\@onedot}
\def\@onedot{\ifx\@let@token.\else.\null\fi\xspace}

\makeatother

\newcommand{\boldparagraph}[1]{\vspace{0.2em}\noindent{\bf #1.}}

\renewcommand{\paragraph}[1]{\boldparagraph{#1}}

\definecolor{darkgreen}{rgb}{0,0.7,0}
\definecolor{newyellow}{rgb}{1,0.8,0.05}
\definecolor{newgreen}{rgb}{0.2,0.8,0.2}

\newcommand{\ours}{Ours\xspace}

\definecolor{lightyellow}{RGB}{255,255,180} %

\newcommand{\cmark}{\color{ForestGreen}{\ding{51}}}%
\newcommand{\xmark}{\color{WildStrawberry}{\ding{55}}}%

\makeatletter
\def\adl@drawiv#1#2#3{%
        \hskip.5\tabcolsep
        \xleaders#3{#2.5\@tempdimb #1{1}#2.5\@tempdimb}%
                #2\z@ plus1fil minus1fil\relax
        \hskip.5\tabcolsep}
\newcommand{\cdashlinelr}[1]{%
  \noalign{\vskip\aboverulesep
           \global\let\@dashdrawstore\adl@draw
           \global\let\adl@draw\adl@drawiv}
  \cdashline{#1}
  \noalign{\global\let\adl@draw\@dashdrawstore
           \vskip\belowrulesep}}
\makeatother

\begin{document}

\twocolumn[{%
\renewcommand\twocolumn[1][]{#1}%
\maketitle
\vspace{-3em}
\begin{center}
    \includegraphics[width=1\textwidth]{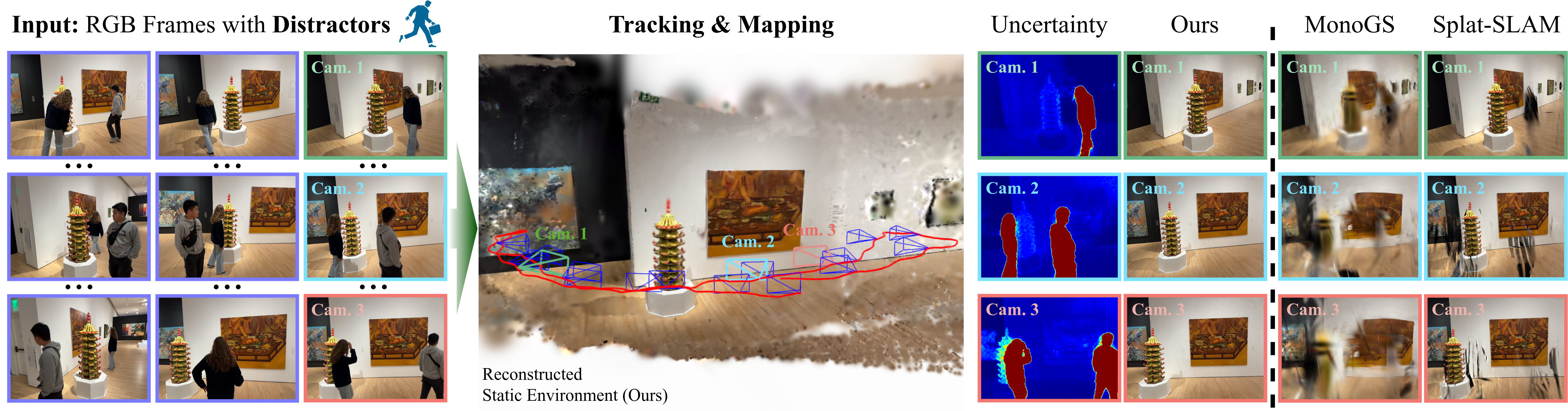}
    \vspace{-1.5em}
    \captionof{figure}{\textbf{\project{}.} Given a monocular video sequence captured in the wild with dynamic distractors, our method accurately tracks the camera trajectory and reconstructs a 3D Gaussian map for static elements, effectively removing all dynamic components. This approach enables high-fidelity rendering even in complex, dynamic scenes. The illustration presents the final 3D Gaussian map, the camera tracking trajectory (in \red{red}), and view synthesis comparisons with baseline methods.
    }
    \label{fig:teaser}
\end{center}
}]
\footnotetext{* Equal contribution.}

\begin{abstract}
We present \project{}, a robust and efficient monocular RGB SLAM system designed to handle dynamic environments by leveraging uncertainty-aware geometric mapping. Unlike traditional SLAM systems, which assume static scenes, our approach integrates depth and uncertainty information to enhance tracking, mapping, and rendering performance in the presence of moving objects. We introduce an uncertainty map, predicted by a shallow multi-layer perceptron and DINOv2 features, to guide dynamic object removal during both tracking and mapping.
This uncertainty map enhances dense bundle adjustment and Gaussian map optimization, improving reconstruction accuracy. Our system is evaluated on multiple datasets and demonstrates artifact-free view synthesis. Results showcase \project{}'s superior performance in dynamic environments compared to state-of-the-art methods.
\end{abstract}
\vspace{-2.0em}
    
\section{Introduction}
\label{sec:intro}

Simultaneous Localization and Mapping (SLAM) in dynamic environments is a fundamental challenge in computer vision, with broad applications in autonomous navigation, augmented reality, and robotics. Traditional SLAM systems ~\cite{Mur2015TRO,Engel2017PAMI,teed2021droid} rely on assumptions of scene rigidity, making them vulnerable to tracking errors in dynamic scenes where objects move independently. Although some recent approaches~\cite{cheng2022sg,palazzolo2019iros,xu2024dgslam} incorporate motion segmentation, semantic information, and depth-based cues to handle dynamic content, they often struggle to generalize across scenes with varied and unpredictable motion patterns. This issue is especially acute in real-world scenarios where dynamic distractors, occlusions, and varying lighting conditions introduce significant ambiguity for SLAM systems.

Uncertainty-aware methods have recently gained attention in scene reconstruction and view synthesis, particularly for handling complex environments with partial occlusions, dynamic objects, and noisy observations. For instance, NeRF \textit{On-the-go}~\cite{ren2024nerf} and WildGaussians~\cite{kulhanek2024wildgaussians} introduced uncertainty estimation to improve the rendering quality of neural radiance fields in real-world scenarios, enabling enhanced view synthesis in the presence of motion and varying light conditions. Such approaches provide valuable insights into modeling ambiguities and have shown strong results in highly dynamic environments. 
However, they focus on sparse-view settings and require camera poses as input.

To address these limitations, we propose a novel SLAM approach, namely \textbf{\project{}}, that leverages a 3D Gaussian Splatting (3DGS) representation, designed to perform robustly in highly dynamic environments using only monocular RGB input. Similar to~\cite{ren2024nerf,kulhanek2024wildgaussians}, our method takes a purely geometric approach. It integrates uncertainty-aware tracking and mapping, which removes dynamic distractors effectively without requiring explicit depth or semantic labels. 
This approach enhances tracking, mapping, and rendering while achieving strong generalizability and robustness across diverse real-world scenarios.
Results showcase its improved performance over prior work in both indoor and outdoor scenes, supporting artifact-free rendering and high-fidelity novel view synthesis, even in challenging settings.

Specifically, we train a shallow multi-layer perceptron (MLP) given 3D-aware, pre-trained DINOv2~\cite{yue2025improving} features to predict per-pixel uncertainty. The MLP is trained incrementally as input frames are streamed into the system, allowing it to dynamically adapt to incoming scene data. We leverage this uncertainty information to enhance tracking, guiding dense bundle adjustment (DBA) to prioritize reliable areas. Additionally, during mapping, the uncertainty predictions inform the rendering loss in Gaussian map optimization, helping to refine the quality of the reconstructed scene. By optimizing the map and the uncertainty MLP independently, we ensure maximal performance for each component. To evaluate our method in diverse and challenging scenarios, we collect a new dataset including indoor and outdoor scenes.

Our main contributions are as follows:
\begin{itemize}
\item A monocular SLAM framework, namely \project{}, utilizing a 3D Gaussian representation that operates robustly in highly dynamic environments, outperforming existing dynamic SLAM methods on a variety of dynamic datasets and on both indoor and outdoor scenarios.
\item An uncertainty-aware tracking and mapping pipeline that enables the accurate removal of dynamic distractors without depth or explicit semantic segmentation, achieving high-fidelity scene reconstructions and tracking.
\item A new dataset, namely Wild-SLAM Dataset, featuring diverse indoor and outdoor scenes, enables SLAM evaluation in unconstrained, real-world conditions. This dataset supports comprehensive benchmarking for dynamic environments with varied object motions and occlusions.
\end{itemize}

\section{Related Work}
\label{sec:related_work}

\begin{figure*}[ht!]
\centering
 \includegraphics[width=\linewidth]{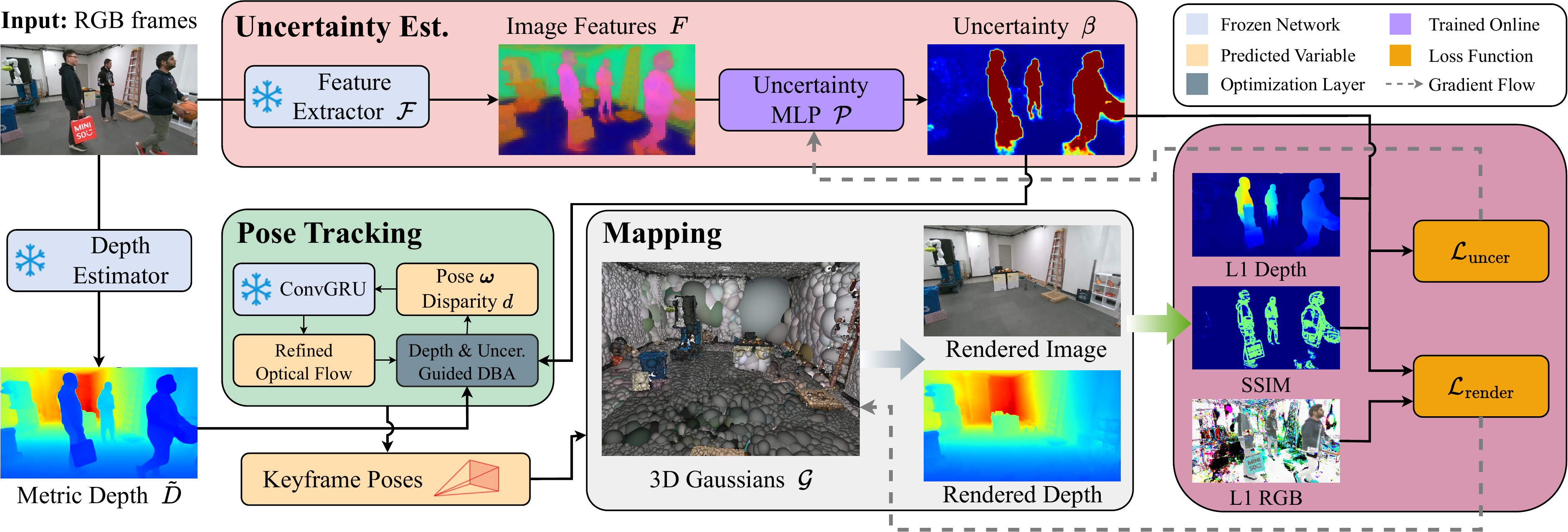}\\
\vspace{-2mm}
\caption{\textbf{System Overview.} \project{} takes a sequence of RGB images as input and simultaneously estimates the camera poses while building a 3D Gaussian map $\mathcal{G}$ of the static scene. Our method is more robust to the dynamic environment due to the uncertainty estimation module, where a pretrained DINOv2 model~\cite{yue2025improving} is first used to extract the image features. An uncertainty MLP $\mathcal{P}$ then utilizes the extracted features to predict per-pixel uncertainty. During the tracking, we leverage the predicted uncertainty as the weight in the dense bundle adjustment (DBA) layer to mitigate the impact of dynamic distractors. We further use monocular metric depth to facilitate the pose estimation. In the mapping module, the predicted uncertainty is incorporated into the rendering loss to update $\mathcal{G}$. Moreover, the uncertainty loss is computed in parallel to train $\mathcal{P}$. Note that $\mathcal{P}$ and $\mathcal{G}$ are optimized independently, as illustrated by the gradient flow in the gray dashed line. Faces are blurred to ensure anonymity. 
}
\vspace{-15pt}
\label{fig:pipeline}
\end{figure*}

\subsection{Traditional Visual SLAM}
Most traditional visual SLAM~\cite{Mur2015TRO,Mur2017TRO,Engel2017PAMI,Klein2007ISMAR} methods assume static scenes, however, the presence of dynamic objects can disrupt feature matching and photometric consistency, leading to substantial tracking drift.
To address this, many approaches enhance robustness in dynamic environments by detecting and filtering out dynamic regions, focusing on reconstructing the static parts of the scene. 
Common approaches to detect dynamic objects include warping or reprojection techniques~\cite{palazzolo2019iros, scona2018staticfusion, cheng2019improving}, off-the-shelf optical flow estimators~\cite{zhang2020flowfusion, sun2018motion}, predefined class priors for object detection or semantic segmentation~\cite{kaneko2018mask, soares2021crowd}, or hybrids of these strategies~\cite{cheng2022sg, wu2022yolo, bescos2018dynaslam, shen2024robust}.
Notably, ReFusion~\cite{palazzolo2019iros} requires RGB-D input and uses a TSDF~\cite{curless1996volumetric} map representation, leveraging depth residuals to filter out dynamic objects. DynaSLAM~\cite{bescos2018dynaslam} supports RGB, RGB-D, and stereo inputs, leveraging Mask R-CNN~\cite{he2017mask} for semantic segmentation with predefined movable object classes, and detects unknown dynamic objects in RGB-D mode via multi-view geometry. 

To our knowledge, no existing traditional SLAM methods support monocular input without relying on prior class information, likely due to the sparse nature of traditional monocular SLAM, which limits the use of purely geometric cues for identifying dynamic regions. Our SLAM approach, however, leverages a 3D Gaussian scene representation to provide dense mapping, enabling support for monocular input without prior semantic information.

\subsection{Neural Implicit and 3DGS SLAM}
Recently, Neural Implicit Representations and 3D Gaussian Splatting (3DGS) have gained substantial interest in SLAM research, as they offer promising advancements in enhancing dense reconstruction and novel view synthesis.
Early SLAM systems like iMAP~\cite{Sucar2021ICCV} and NICE-SLAM~\cite{Zhu2022CVPR} pioneered the use of neural implicit representations, integrating mapping and camera tracking within a unified framework. Subsequent works have further advanced these methods by exploring various optimizations and extensions, including efficient representations~\cite{Johari2022ESLAM, Kruzhkov2022MESLAM, wang2023co}, monocular settings~\cite{zhu2024nicer, zhang2023hi, belos2022mod}, and the integration of semantic information~\cite{zhu2024sni, li2023dns, zhai2024nis}.
The emergence of 3D Gaussian Splatting (3DGS)~\cite{kerbl3Dgaussians} introduces an efficient and flexible alternative representation for SLAM and has been adopted in several recent studies~\cite{yan2024gs, huang2024photo, keetha2024splatam, li2025sgs, ha2024rgbd, hu2025cg, peng2024rtg, li2024gs3lam, zhu2024loopsplat}. Among these, MonoGS~\cite{matsuki2024gaussian} is the first near real-time monocular SLAM system to use 3D Gaussian Splatting as its sole scene representation. Another notable advancement is Splat-SLAM~\cite{sandstrom2024splat}, the state-of-the-art (SoTA) in monocular Gaussian Splatting SLAM, offering high-accuracy mapping with robust global consistency.
These methods excel in tracking and reconstruction but typically assume static scene conditions, limiting their robustness as performance degrades significantly in dynamic environments. 

Some methods have focused explicitly on handling dynamic environments. Most approaches extract a dynamic object mask for each frame before passing it to the tracking and mapping components. DG-SLAM~\cite{xu2024dgslam}, DynaMon~\cite{schischka2023dynamon}, and RoDyn-SLAM~\cite{jiang2024rodyn} combine segmentation masks with motion masks derived from optical flow.  
DDN-SLAM~\cite{li2024ddn} employs object detection combined with a Gaussian Mixture Model to distinguish between foreground and background, checking feature reprojection error to enhance tracking accuracy. 
However, these approaches rely heavily on prior knowledge of object classes and depend on object detection or semantic segmentation, limiting their generalizability in real-world settings where dynamic objects may be unknown a priori and difficult to segment.

In contrast, our method is purely geometric even with monocular input. While similar works such as NeRF \textit{On-the-go}~\cite{ren2024nerf} and WildGaussians~\cite{kulhanek2024wildgaussians} demonstrate distractor removal in dynamic environments, they are primarily designed for sparse-view settings with known camera poses. Inspired by these approaches, we also leverage the pre-trained 2D foundation model DINOv2~\cite{oquab2023dinov2} and use an MLP to decode them into an uncertainty map. We extend this framework to tackle the challenging sequential SLAM setting, integrating specific design components that enable robust tracking and high-fidelity mapping within our 3DGS backend.

A concurrent work, MonST3R~\cite{zhang2024monst3r}, introduced a feed-forward approach for estimating scene geometry in the presence of motion. 
It detects moving objects by thresholding the difference between the predicted optical flow~\cite{wang2025sea} and the reprojection flow, estimated using an extended version of DUSt3R~\cite{wang2024dust3r}. However, this approach is limited to short sequences, and its use of point clouds as the scene representation does not support view synthesis.

\section{Method}
\label{sec:method}

Given a sequence of RGB frames $\{I_{i}\}^{N}_{i=1}$ captured in a dynamic environment, \project{} tracks the camera pose while reconstructing the static part of the scene as a 3D Gaussian map (\secref{sec:scene_rep}). To mitigate the adverse impact of moving objects in tracking and eliminate them from the 3D reconstruction, we utilize DINOv2 features \cite{oquab2023dinov2} and a shallow MLP to decode them to per-pixel uncertainty (\secref{sec:uncertainty}).
We further introduce how this uncertainty is integrated into the optical-flow-based tracking component (\secref{sec:tracking}). In parallel with tracking, we leverage this uncertainty to progressively expand and optimize the 3D Gaussian map with uncertainty-aware loss functions (\secref{sec:mapping}). The overview of \project{} is in \figref{fig:pipeline}.

\subsection{Preliminary on 3D Gaussian Splatting}
\label{sec:scene_rep}
We utilize a 3D Gaussian representation \cite{kerbl3Dgaussians} to reconstruct the static part of the scanned environment. The scene is represented by a set of anisotropic Gaussians $\mathcal{G}=\{ g_{i}\}^{K}_{i=1}$. Each Gaussian $g_{i}$ contains color $\boldsymbol{c}_{i} \in \mathbb{R}^{3}$, opacity $o_{i} \in [0,1]$, mean $\boldsymbol{\mu}_{i} \in \mathbb{R}^{3}$, and covariance matrix $\boldsymbol{\Sigma}_{i} \in \mathbb{R}^{3 \times 3}$.

\noindent \textbf{Rendering.} We follow the same rendering approach as in the original 3DGS~\cite{kerbl3Dgaussians} but omit spherical harmonics to speed up optimization, as in \cite{matsuki2024gaussian, yugay2023gaussian}. Given a camera-to-world pose $\bomega$ and the projection function $\Pi_{c}$ that maps 3D points onto the image frame, the 3D Gaussians can be "splatted" onto the 2D image plane by projecting the mean $\boldsymbol{\mu}$ and covariance matrix $\boldsymbol{\Sigma}$ as $\boldsymbol{\mu}' = \Pi_{c}\left(\bomega^{-1}\boldsymbol{\mu}\right)$ and $\boldsymbol{\Sigma}'=\boldsymbol{J}\boldsymbol{R}\boldsymbol{\Sigma}\boldsymbol{R}^{T}\boldsymbol{J}^{T}$, where $\boldsymbol{J}$ is the Jacobian of the linear approximation of the projective transformation and $\boldsymbol{R}$ is the rotation component of $\boldsymbol{\omega}$. The opacity of a Gaussian $g_{i}$ at pixel $\boldsymbol{x}'$ is:
\small
\begin{equation}
  \alpha_{i} = o_{i}\exp (-\frac{1}{2}(\boldsymbol{x}'-\boldsymbol{\mu}_{i}')^{T}\boldsymbol{\Sigma}_{i}'^{-1}(\boldsymbol{x}'-\boldsymbol{\mu}_{i}')) \enspace.
\end{equation}
\normalsize
The rendered color $\hat{I}$ and depth $\hat{D}$ at pixel $\boldsymbol{x}'$ are obtained by blending the 3D Gaussians $\mathcal{G'}$ overlapping with this pixel, sorted by their depth relative to the camera plane:
\small
\begin{align}
\hat{I} = \sum_{i \in \mathcal{G'}} \boldsymbol{c}_i \alpha_i \prod_{j=1}^{i-1}\left(1 - \alpha_j\right), 
\hat{D} = \sum_{i \in \mathcal{G'}} \hat{d}_i \alpha_i \prod_{j=1}^{i-1}\left(1 - \alpha_j\right)
\label{eq:render}
\end{align}
\normalsize
where $\hat{d}_{i}$ is the z-axis depth of the center of $g_{i}$. 
This process is fully differentiable, enabling incremental map updates as new frames are streamed (we discuss details in~\secref{sec:mapping}).

\subsection{Uncertainty Prediction}
\label{sec:uncertainty}
\project{}'s main contribution is to eliminate the impact of moving distractors in both mapping and tracking. To achieve this, we use an uncertainty prediction component inspired by \cite{ren2024nerf,kulhanek2024wildgaussians} while additionally incorporating our own custom depth uncertainty loss during training.
For each input frame, we extract DINOv2 \cite{oquab2023dinov2} features and utilize an uncertainty MLP, trained on-the-fly with streamed frames, to predict a per-pixel uncertainty map that mitigates the impact of distractors in both tracking and mapping.

\noindent  \textbf{Feed-Forward Uncertainty Estimation.} 
Given an input image $I_{i}$, we use a pre-trained DINOv2 feature extractor $\mathcal{F}$ to derive image features, $F_{i} = \mathcal{F}(I_{i})$. Instead of the original DINOv2 model \cite{oquab2023dinov2}, we use the finetuned version from \cite{yue2025improving}, which injects 3D awareness into the model. The features are used as input to a shallow uncertainty MLP $\mathcal{P}$ to predict an uncertainty map $\beta_{i} = \mathcal{P}(F_{i})$.
We bilinearly upsample $\beta_{i}$ to the original input frame resolution, which is then used in both tracking (\secref{sec:tracking}) and mapping (\secref{sec:mapping}).

\noindent \textbf{Uncertainty Loss Function.}
For the uncertainty loss functions, we adopt the modified SSIM loss and two regularization terms from NeRF \textit{On-the-go}~\cite{ren2024nerf}, along with the L1 depth loss term:
\begin{equation}
\mathcal{L}_\text{depth} = | \hat{D}_{i}-\Tilde{D}_{i}|_{1},
\label{eq:depth_uncer_loss}
\end{equation}
where $\mathcal{L}_\text{depth}$ represents the L1 loss between the rendered depth $\hat{D}_{i}$ and the metric depth $\tilde{D}_{i}$, as estimated by Metric3D v2~\cite{hu2024metric3d}. We find that this additional depth signal effectively improves the model's ability to distinguish distractors, enhancing the training of the uncertainty MLP.
Therefore the total uncertainty loss is:
\begin{equation}
\mathcal{L}_{\text {uncer}} = \frac{\mathcal{L}_{\text {SSIM }}' + \lambda_1 \mathcal{L}_{\text {uncer\_D}}}{\beta_{i}^2}+\lambda_2 \mathcal{L}_{\text {reg\_V}}+\lambda_3 \mathcal{L}_{\text {reg\_U}},
\label{eq:uncer_loss}
\end{equation}
where $\lambda_{*}$ are hyperparameters, $\mathcal{L}_{\text {SSIM }}'$ is the modified SSIM loss, $\mathcal{L}_{\text {reg\_V}}$ minimizes the variance of predicted uncertainty for features having high similarity, and the last term  $\mathcal{L}_{\text {reg\_U}}=\log \beta_{i}$ prevents $\beta_{i}$ from being infinitely large.
Please refer to NeRF \textit{On-the-go}~\cite{ren2024nerf} for details on $\mathcal{L}_{\text {SSIM }}'$, $\mathcal{L}_{\text {reg\_V}}$, and $\mathcal{L}_{\text {reg\_U}}$.
We use $\mathcal{L}_{\text {uncer}}$ to train $\mathcal{P}$ in parallel with map optimization (\secref{sec:mapping}).

\subsection{Tracking}
\label{sec:tracking}
Our tracking component is based on the recent method DROID-SLAM~\cite{teed2021droid} with the incorporation of depth and uncertainty into the DBA to make the system robust in dynamic environments.
The original DROID-SLAM~\cite{teed2021droid} uses a pretrained recurrent optical flow model coupled with a DBA layer to jointly optimize keyframe camera poses and disparities.
This optimization is performed over a frame graph, denoted as \( G = (V, E) \), where \( V \) represents the selected keyframes and \( E \) represents the edges between keyframes. 
Following \cite{sandstrom2024splat, zhang2024glorie}, we incorporate loop closure and online global BA to reduce pose drift over long sequences.

\noindent \textbf{Depth and Uncertainty Guided DBA.} 
Different from \cite{teed2021droid,sandstrom2024splat,zhang2024glorie}, we integrate the uncertainty map estimated by $\mathcal{P}$ into the BA optimization objective to deal with the moving distractors. 
In addition, we utilize the metric depth estimated by Metric3D V2 \cite{hu2024metric3d} to stabilize the DBA layer, since $\mathcal{P}$ is trained online and can not always give accurate uncertainty estimation, especially during the early stages of tracking. 
For each newly inserted keyframe $I_{i}$, we first estimate its monocular metric depth $\Tilde{D}_{i}$ and add it to the DBA objective alongside optical flow:

\small
\begin{equation}
\begin{aligned}
\underset{\bomega, d}{\arg \min } \sum_{(i, j) \in E} &\left\|\tilde{p}_{i j}-\Pi_c\left( \bomega_j^{-1}\bomega_i \Pi_c^{-1}\left(p_i, d_i\right)\right)\right\|_{\Sigma_{i j}/\beta_{i}^{2}}^2\\
& + \lambda_{4} \sum_{i \in V}\left\|M_{i}\left(d_i-1 / \Tilde{D}_i\right)\right\|^2,
\label{eq:dba}
\end{aligned}
\end{equation}

\normalsize The first term is the uncertainty-aware DBA objective where $\tilde{p}_{i j}$ is the predicted pixel position of pixels $p_{i}$ projected into keyframe $j$ by the estimated optical flow; this is iteratively updated by a Convolutional Gated Recurrent Unit (ConvGRU) \cite{teed2021droid}. $\Pi_{c}$ represents the projection from 3D points to 2D image planes, $\bomega_{i}$ is the camera-to-world transformation for keyframe $i$, $d_{i}$ is the optimized disparity, and $\|\cdot\| _{\Sigma_{i j}/\beta_{i}^{2}}$ is the Mahalanobis distance~\cite{mclachlan1999mahalanobis} which weighs the error terms by confidence matrix $\Sigma_{i j}$ from the flow estimator \cite{teed2021droid} and our uncertainty map $\beta_{i}$. 
As a result, pixels associated with moving objects will have minimal impact on the optimization in DBA.

\normalsize The second term is a disparity regularization term to encourage $1/d_{i}$ to be close to the predicted depth for all $i$ in the graph nodes $V$.
We find that this regularization term can stabilize the pose estimation, especially when the uncertainty MLP $\mathcal{P}$ has not converged to provide reliable uncertainty $\beta_{i}$ while moving objects are dominant in the image frames.
$M_{i}$ is a binary mask that deactivates disparity regularization in regions where $\tilde{D}_{i}$ is unreliable, computed via multi-view depth consistency (details provided in the supplementary).

\subsection{Mapping}
\label{sec:mapping}
After the tracking module predicts the pose of a newly inserted keyframe, its RGB image $I$, metric depth $\tilde{D}$, and estimated pose $\bomega$ will be utilized in the mapping module to expand and optimize the 3DGS map.
Given a new keyframe processed in the tracking module, we expand the Gaussian map to cover newly explored areas, using $\tilde{D}_{i}$ as proxy depth, following the RGBD strategy of MonoGS~\cite{matsuki2024gaussian}.
Before optimization, we also actively deform the 3D Gaussian map if the poses of previous keyframes are updated by loop closure or global BA, as in Splat-SLAM~\cite{sandstrom2024splat}.

\noindent \textbf{Map update.}
After the map is expanded, we optimize the Gaussians for a fixed number of iterations. We maintain a local window of keyframes selected by inter-frame covisibility, similar to MonoGS \cite{matsuki2024gaussian}. At each iteration, we randomly sample a keyframe with at least \(50\%\) probability evenly distributed for the keyframes in the local window, while all the other keyframes share the remaining probability equally. 
For a selected keyframe, we render the color $\hat{I}$ and depth $\hat{D}$ image by \eqnref{eq:render}. The Gaussian map $\mathcal{G}$ is optimized by minimizing the render loss $\mathcal{L}_{\text{render}}$: 
\begin{equation}
\mathcal{L}_{\text {render}} = \frac{\lambda_{5}\mathcal{L}_\text{color} + \lambda_6\mathcal{L}_\text{depth}}{\beta^2}+\lambda_7 \mathcal{L}_{\text{iso}},
\label{eq:render_loss}
\end{equation}
where the color loss $\mathcal{L}_\text{color}$, combines L1 and SSIM losses as follows:
\begin{equation}
\mathcal{L}_\text{color} = (1-\lambda_{\text{ssim}}) \|\hat{I}-I\|_{1} + \lambda_{\text{ssim}}\mathcal{L}_{\text{ssim}}.
\label{eq:render_color_loss}
\end{equation}
Unlike the loss function for static scenes, here we incorporate the uncertainty map $\beta$, which serves as a weighting factor for $\mathcal{L}_\text{color}$ and $\mathcal{L}_\text{depth}$, minimizing the influence of distractors during mapping optimization.
Additionally, isotropic regularization loss $\mathcal{L}_{\text{iso}}$ \cite{matsuki2024gaussian} constrains 3D Gaussians to prevent excessive elongation in sparsely observed regions.

At each iteration, we also compute $\mathcal{L}_\text{uncer}$ given the rendered color and depth image as in \eqnref{eq:uncer_loss}. $\mathcal{L}_\text{uncer}$ is then used to train the uncertainty MLP $\mathcal{P}$ in parallel to the map optimization. As shown in \cite{ren2024nerf}, it is crucial to separately optimize the 3D Gaussian map and the uncertainty MLP. Therefore, we detach the gradient flow from $\mathcal{L}_\text{uncer}$ to the Gaussians $\mathcal{G}$, as well as from $\mathcal{L}_{\text {render}}$ to $\mathcal{P}$.

\section{Experiments}
\label{sec:exp}

\begin{table*}[t]
\centering
\footnotesize
\setlength{\tabcolsep}{4.5pt}
{
\begin{tabular}{lrrrrrrrrrrrr}
\toprule
Method & \texttt{ANYmal1} & \texttt{ANYmal2} & \texttt{Ball} & \texttt{Crowd} & \texttt{Person} & \texttt{Racket} & \texttt{Stones} & \texttt{Table1} & \texttt{Table2} & \texttt{Umbrella} & Avg. \\
\midrule
\multicolumn{12}{l}{\cellcolor[HTML]{EEEEEE}{\textit{RGB-D}}} \\ 
Refusion~\cite{palazzolo2019iros} & 4.2 & 5.6 & 5.0 & 91.9 & 5.0 & 10.4 & 39.4 & 99.1 & 101.0 & 10.7 & 37.23\\
DynaSLAM (N+G)~\cite{bescos2018dynaslam} & 1.6 & \rd 0.5 & \rd 0.5 & 1.7 & \nd 0.5 & \rd 0.8 & 2.1 & \rd 1.2 &  34.8 & 34.7 & 7.84 \\
NICE-SLAM~\cite{Zhu2022CVPR} & F & 123.6 & 21.1 & F & 150.2 & F & 134.4 & 138.4 & F & 23.8 & -\\

\hdashline
\noalign{\vskip 1pt}
\multicolumn{12}{l}{\cellcolor[HTML]{EEEEEE}{\textit{Monocular}}} \\ 
DSO~\cite{Engel2017PAMI} & 12.0 & 2.5 & 1.0 & 88.6 & 9.3 & 3.1 & 41.5 & 50.6 & 85.3 & 26.0 & 32.99 \\
DROID-SLAM~\cite{teed2021droid} & \rd 0.6 & 4.7 & 1.2 & 2.3 & \rd 0.6 & 1.5 & 3.4 & 48.0 & 95.6 &  3.8 & 16.17\\
DynaSLAM (RGB)~\cite{bescos2018dynaslam} & \rd 0.6 & \rd 0.5 & \rd 0.5 & \nd 0.5 & \fs 0.4 & \nd 0.6 & \nd 1.7 & 1.8 & 42.1 & \rd 1.2 & \rd 5.19\\
MonoGS~\cite{matsuki2024gaussian}& 8.8 & 51.6 & 7.4 & 70.3 & 55.6 & 67.6 & 39.9 & 24.9 & 118.4 & 35.3 & 47.99\\
Splat-SLAM~\cite{sandstrom2024splat}& \nd 0.4 & \nd 0.4 & \nd 0.3 & \rd 0.7 &  0.8 & \nd 0.6 & \rd 1.9 & 2.5 & 73.6 & 5.9 & 8.71\\
MonST3R-SW~\cite{zhang2024monst3r} & 3.5 & 21.6 & 6.1 & 14.4 & 7.2 & 13.2 & 11.2 & 4.8 & \rd 33.7 & 5.5 & 12.12 \\
MegaSaM~\cite{li2024megasam} & \rd 0.6 & 2.7 & 0.6 & 1.0 & 3.2 & 1.6 & 3.2 & \nd 1.0 &  \nd 9.4 &  \nd 0.6 & \nd 2.40\\
{\textbf{\project{} (\ours)}} & \fs 0.2 & \fs 0.3 & \fs 0.2 & \fs 0.3 & 0.8 & \fs 0.4 & \fs 0.3 & \fs 0.6 & \fs 1.3 & \fs 0.2 & \fs 0.46\\
\bottomrule
\end{tabular}

}
\vspace{-2mm}

\caption{\textbf{Tracking Performance on our Wild-SLAM MoCap Dataset} (ATE RMSE $\downarrow$ [cm]). Best results are highlighted as\colorbox{colorFst}{\bf first},\colorbox{colorSnd}{second}, and\colorbox{colorTrd}{third}. All baseline methods were run using their publicly available code. For DynaSLAM (RGB), initialization is time-consuming for certain sequences, and only keyframe poses are generated and evaluated. `F' denotes tracking failure.} 
\label{tab:mocap_tracking}
\vspace{-5pt}
\end{table*}

\begin{figure*}[t!]
  \centering
  \footnotesize
  \setlength{\tabcolsep}{1.5pt}
  \newcommand{\sz}{0.155}
  \newcommand{\sza}{0.18} %
  \begin{tabular}{ccc:ccc:c}
  \raisebox{1.8\normalbaselineskip}[0pt][0pt]{\rotatebox[origin=c]{90}{\scriptsize ANYmal2}} &
  \includegraphics[width=\sz\linewidth]{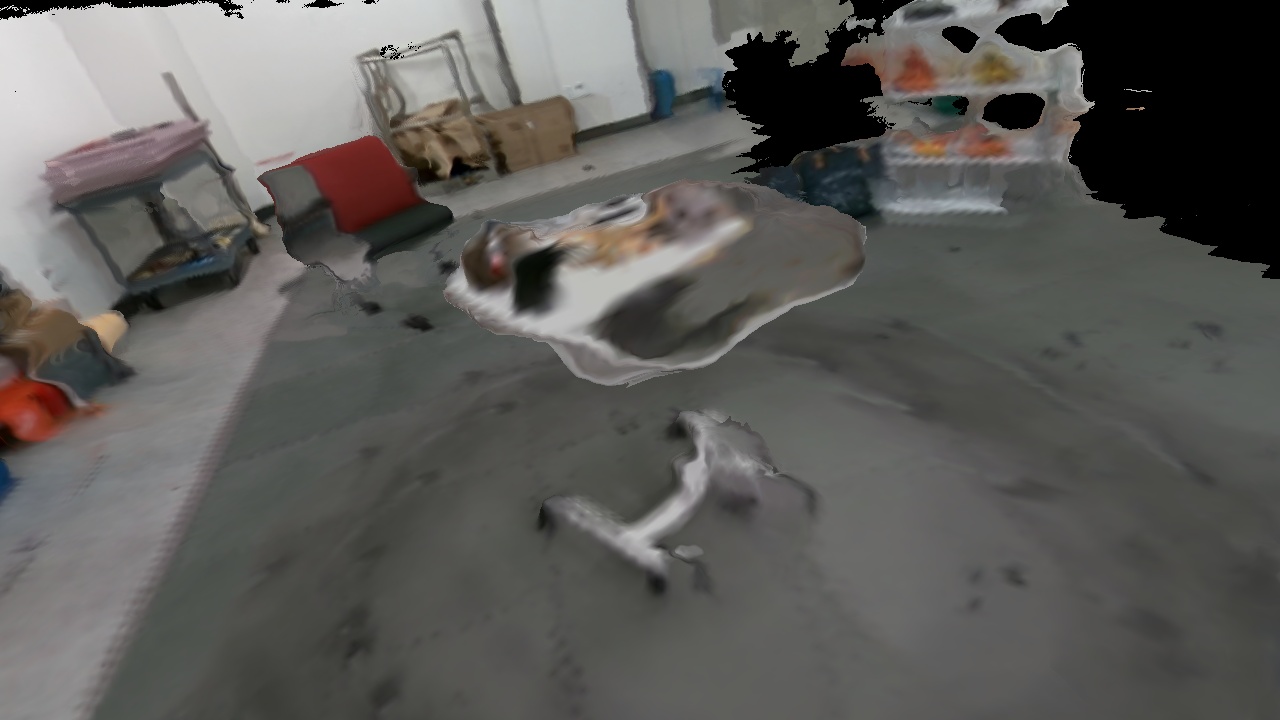} &
  \includegraphics[width=\sz\linewidth]{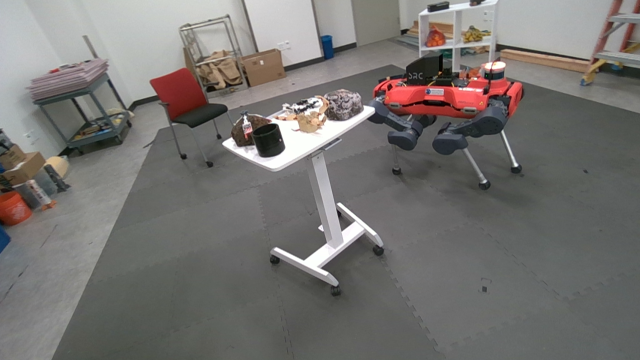} &
  \includegraphics[width=\sz\linewidth]{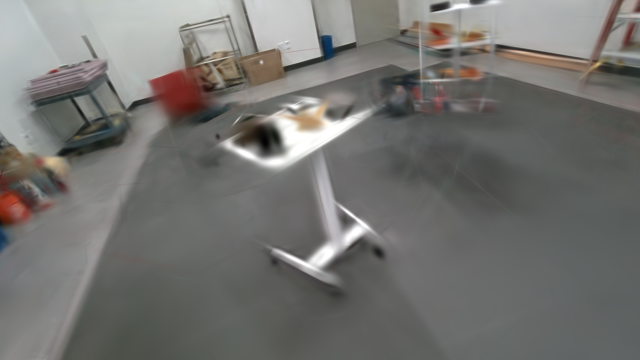} &
  \includegraphics[width=\sz\linewidth]{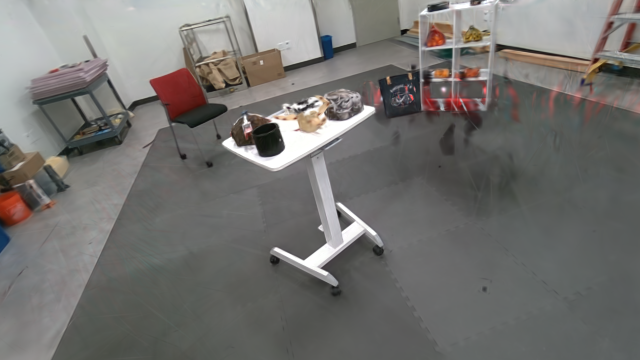} &
  \includegraphics[width=\sz\linewidth]{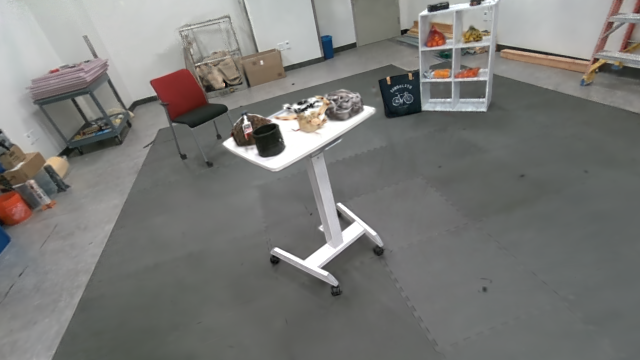}  &
  \includegraphics[width=\sz\linewidth]{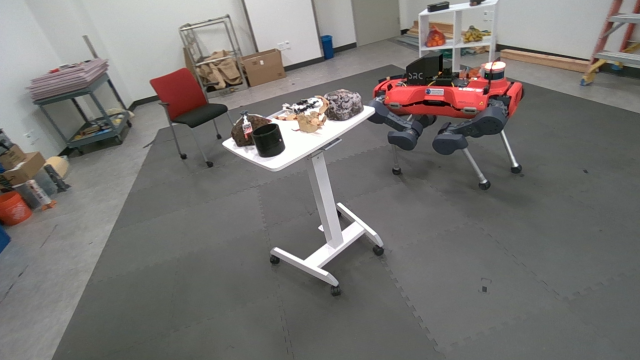} 
  \\
  \raisebox{1.8\normalbaselineskip}[0pt][0pt]{\rotatebox[origin=c]{90}{Stones}} &
  \includegraphics[width=\sz\linewidth]{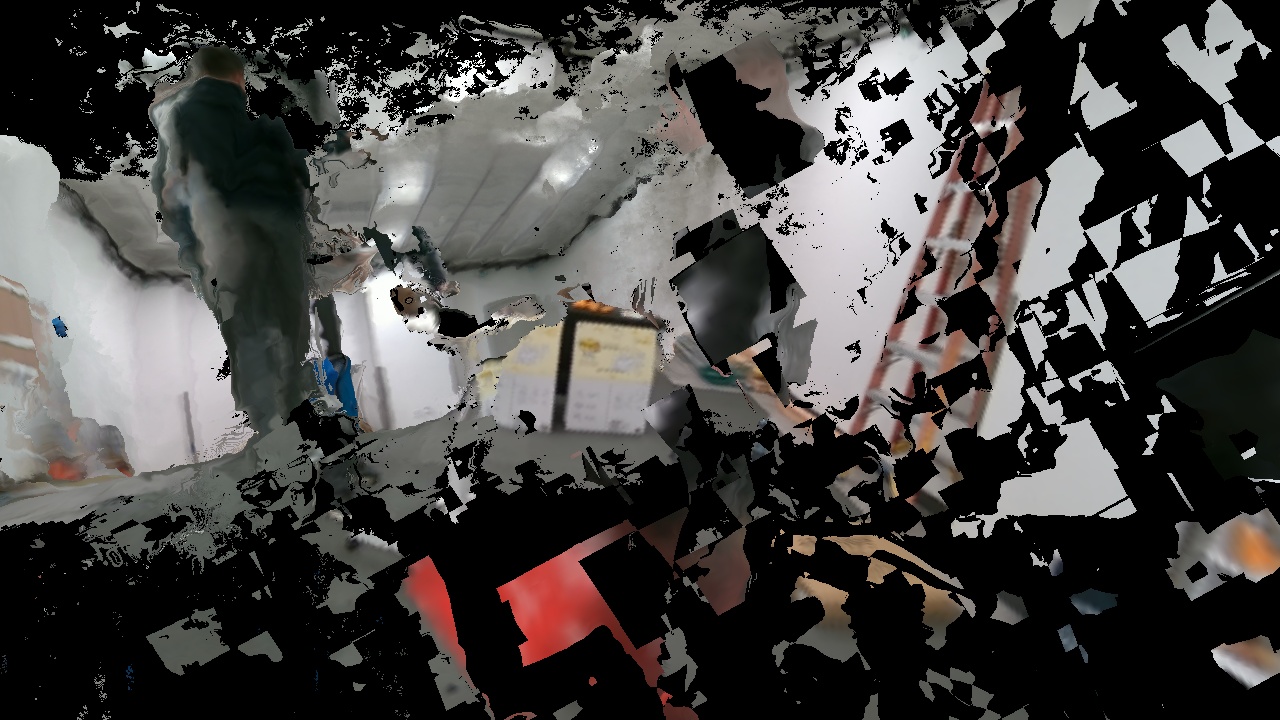} &
  \includegraphics[width=\sz\linewidth]{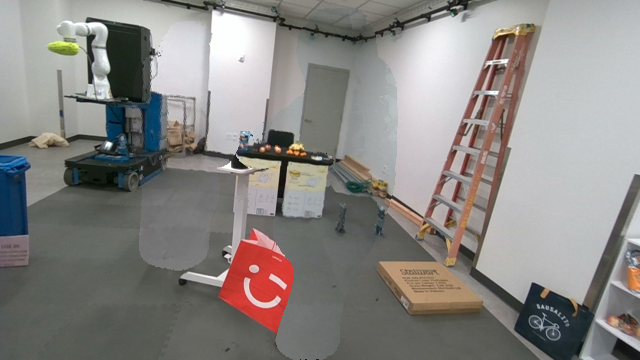} &
  \includegraphics[width=\sz\linewidth]{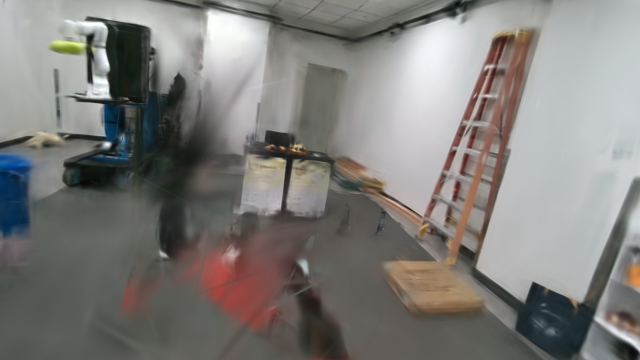} &
  \includegraphics[width=\sz\linewidth]{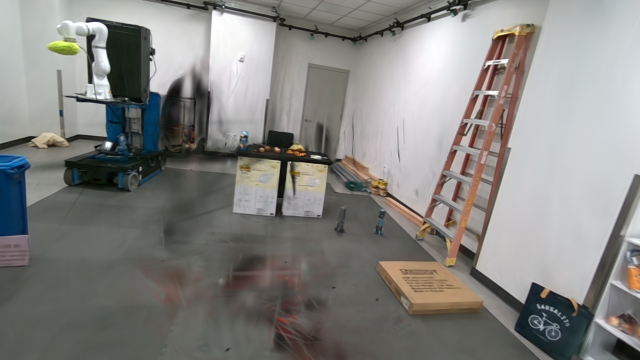} &
  \includegraphics[width=\sz\linewidth]{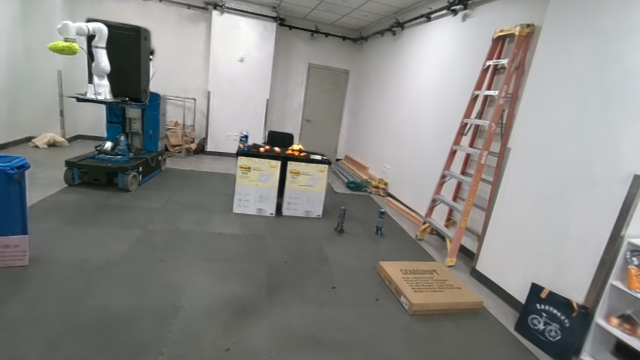}  &
  \includegraphics[width=\sz\linewidth]{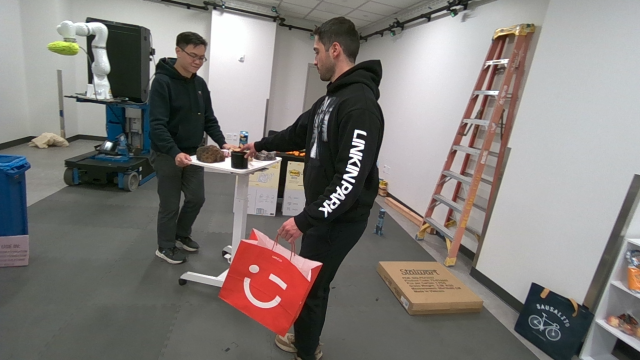} 
  \\
  \raisebox{1.8\normalbaselineskip}[0pt][0pt]{\rotatebox[origin=c]{90}{Umbrella}} &
  \includegraphics[width=\sz\linewidth]{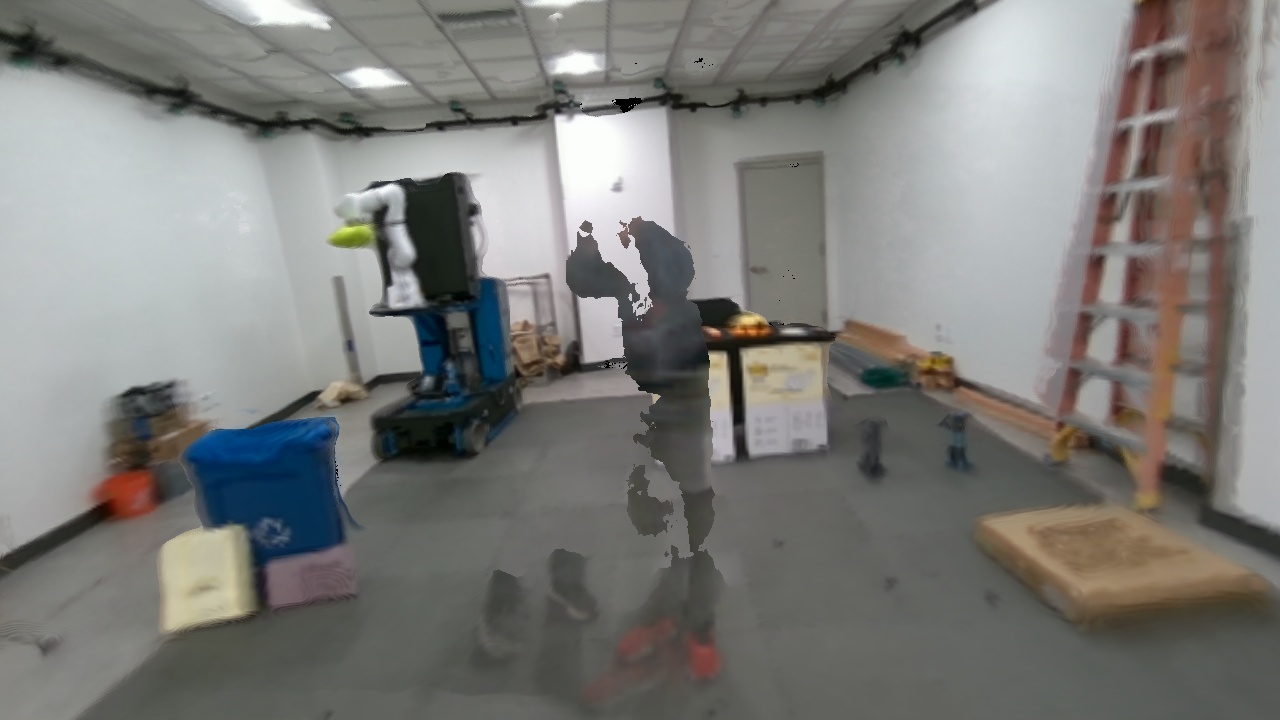} &
  \includegraphics[width=\sz\linewidth]{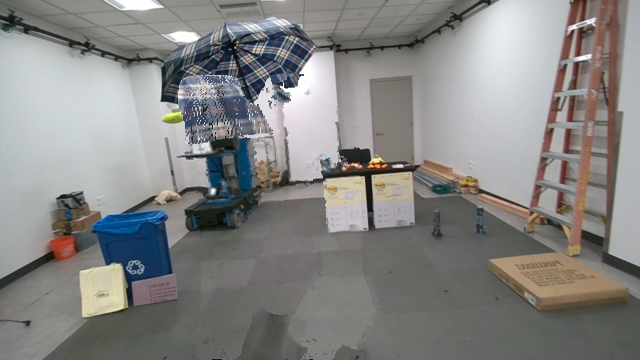} &
  \includegraphics[width=\sz\linewidth]{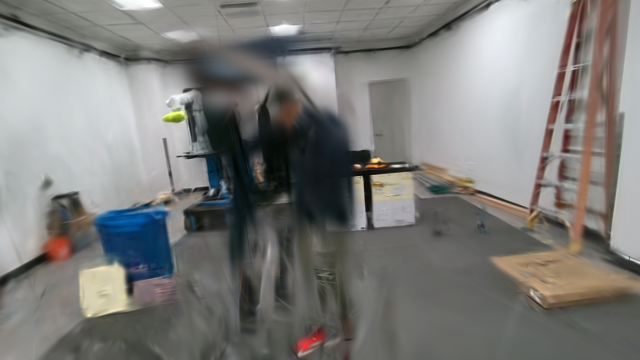} &
  \includegraphics[width=\sz\linewidth]{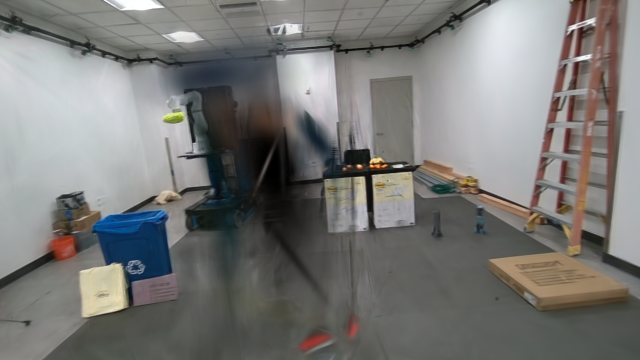} &
  \includegraphics[width=\sz\linewidth]{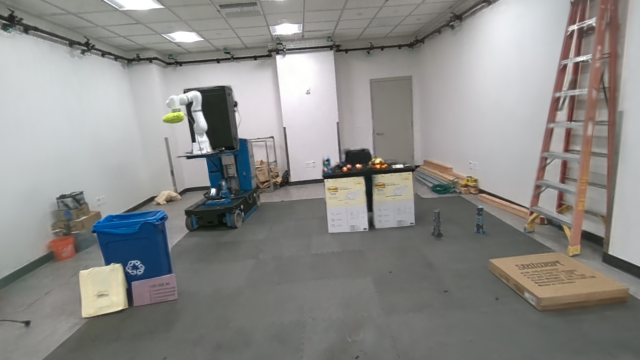}  &
  \includegraphics[width=\sz\linewidth]{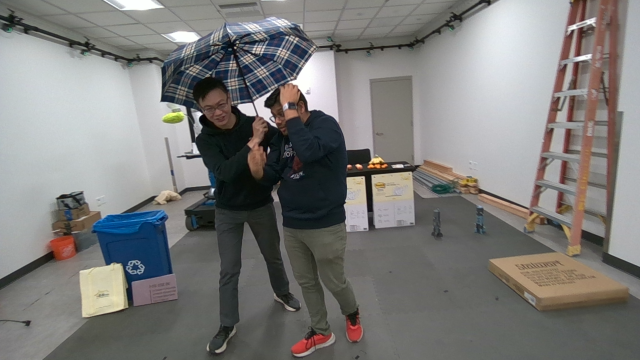} 
  \\
  & {ReFusion~\cite{palazzolo2019iros}}   & {DynaSLAM (N+G)~\cite{bescos2018dynaslam}} &  {MonoGS~\cite{matsuki2024gaussian}} & {Splat-SLAM~\cite{sandstrom2024splat}} & {\textbf{\project{} (\ours)}}  &  {Input} \\
\multicolumn{3}{c:}{\textbf{\textit{RGB-D input}}} & \multicolumn{3}{c:}{\textbf{\textit{Monocular input}}} & \\ 
     
  \end{tabular} 
  
  \vspace{-2mm}
  \caption{\textbf{\textit{Input} View Synthesis Results on our Wild-SLAM MoCap Dataset.} Regardless of the distractor type, our method is able to remove distractors and render realistic images. Faces are blurred to ensure anonymity. 
  }
  \label{fig:mocap_rendering}
\vspace{-5pt}
\end{figure*}

\begin{table*}[!t]
  \centering
  \footnotesize
  \setlength{\tabcolsep}{2pt}          %
  \resizebox{0.95\linewidth}{!}{
    \begin{tabular}{llccccccccccc}
      \toprule
           & & \texttt{ANYmal1} & \texttt{ANYmal2} & \texttt{Ball} & \texttt{Crowd} & \texttt{Person} & \texttt{Racket} & \texttt{Stones} & \texttt{Table1} & \texttt{Table2} & \texttt{Umbrella} & Avg. \\  
      \midrule 
      \multicolumn{13}{l}{\cellcolor[HTML]{EEEEEE}{\textit{Monocular}}} \\

    \multirow{3}{*}{\rotatebox[origin=c]{0}{Splat-SLAM~\cite{sandstrom2024splat}}} & 
    
          PSNR $\uparrow$ 
          & 19.71 & 20.32 & 17.68 & 16.00 & 18.58 & 16.45 & 17.90 & 17.54 & 11.45 & 16.65 & 17.23\\
          & SSIM $\uparrow$ 
          & 0.786 & 0.800 & 0.702 & 0.693 & 0.754 & 0.699 & 0.711 & 0.717 & 0.458 & 0.667 & 0.699\\
          & LPIPS $\downarrow$ 
          & 0.313 & 0.278 & 0.294 & 0.356 & 0.298 & 0.301 & 0.291 & 0.312 & 0.650 & 0.362 & 0.346\\
      \midrule
    \multirow{3}{*}{\rotatebox[origin=c]{0}{\textbf{\project{} (\ours)}}} & 
    
          PSNR $\uparrow$ 
          & \textbf{21.85} & \textbf{21.46} & \textbf{20.06} & \textbf{21.28} & \textbf{20.31} & \textbf{20.87} & \textbf{20.52} & \textbf{20.33} & \textbf{19.16} & \textbf{20.03} & \textbf{20.59}\\
          & SSIM $\uparrow$ 
          & \textbf{0.807} & \textbf{0.832} & \textbf{0.754} & \textbf{0.802} & \textbf{0.801} & \textbf{0.785} & \textbf{0.768} & \textbf{0.788} & \textbf{0.728} & \textbf{0.766} & \textbf{0.783}\\
          & LPIPS $\downarrow$ 
          & \textbf{0.211} & \textbf{0.230} & \textbf{0.191} & \textbf{0.176} & \textbf{0.189} & \textbf{0.186} & \textbf{0.185} & \textbf{0.209} & \textbf{0.303} & \textbf{0.210} & \textbf{0.209}\\
      
      \bottomrule
    \end{tabular}}%
    \vspace{-2mm}
    \caption{\textbf{\textit{Novel} View Synthesis Evaluation on our Wild-SLAM MoCap Dataset.} Best results are in bold. 
    }
\label{tab:mocap_rendering}
\vspace{-15pt}
\end{table*}

\input{figs_tex/mocap_nvs}
\begin{figure*}[t!]
  \centering
  \footnotesize
  \setlength{\tabcolsep}{1.5pt}
  \newcommand{\sz}{0.132}
  \newcommand{\sza}{0.18} %
  \begin{tabular}{cccccccc}
  \raisebox{2.8\normalbaselineskip}[0pt][0pt]{\rotatebox[origin=c]{90}{Parking}} &
  \includegraphics[width=\sz\linewidth]{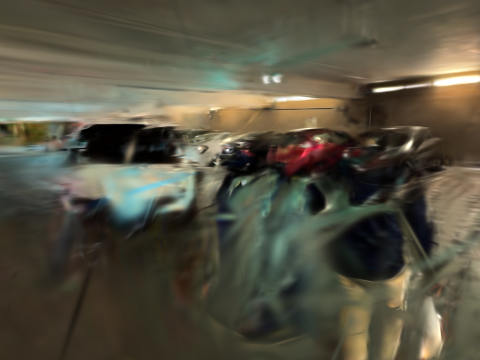} &
  \includegraphics[width=\sz\linewidth]{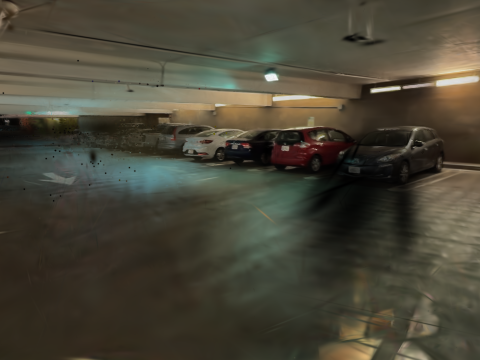} &
  \includegraphics[width=\sz\linewidth]{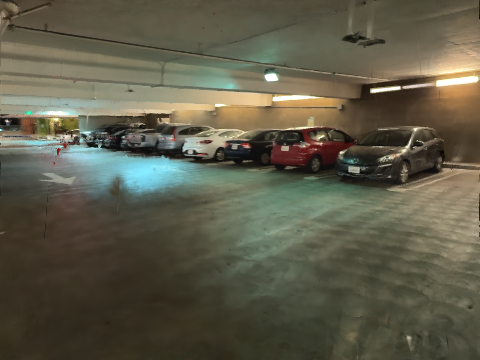} &
  \includegraphics[width=\sz\linewidth]{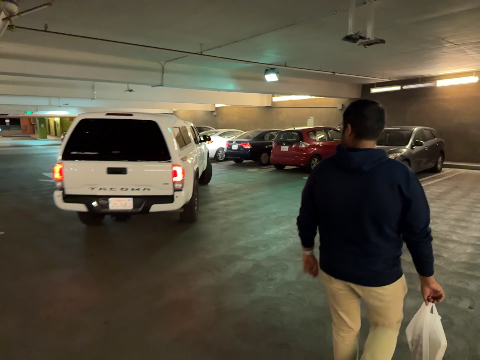} &
  \includegraphics[width=\sz\linewidth]{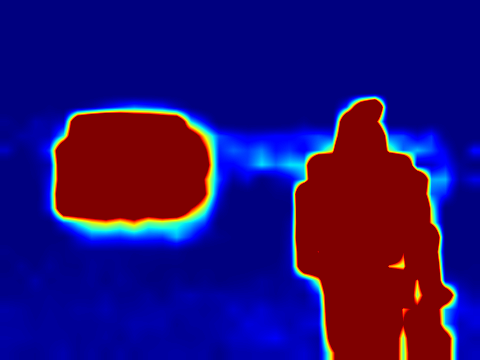}  &
  \includegraphics[width=\sz\linewidth]{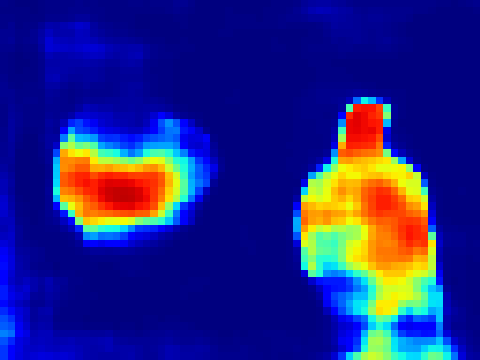} 
  &
  \includegraphics[width=\sz\linewidth]{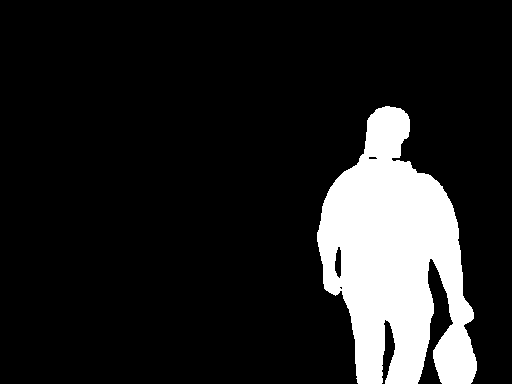} 
  \\
  \raisebox{2.5\normalbaselineskip}[0pt][0pt]{\rotatebox[origin=c]{90}{Piano}} &
  \includegraphics[width=\sz\linewidth]{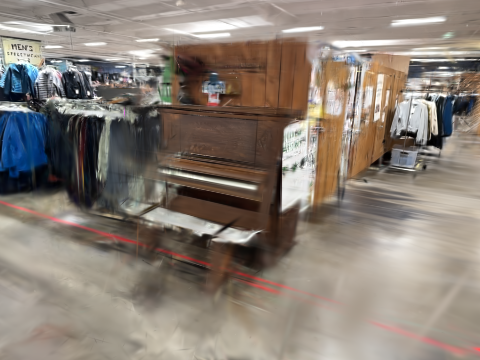} &
  \includegraphics[width=\sz\linewidth]{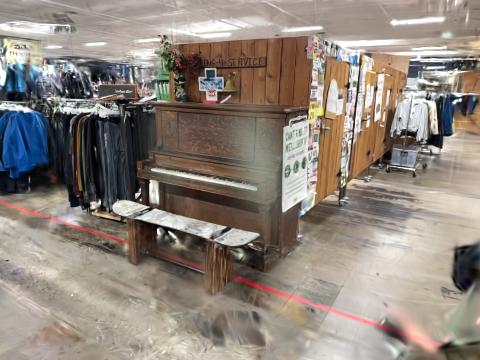} &
  \includegraphics[width=\sz\linewidth]{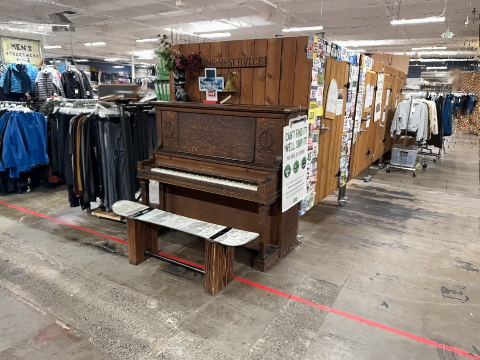} &
  \includegraphics[width=\sz\linewidth]{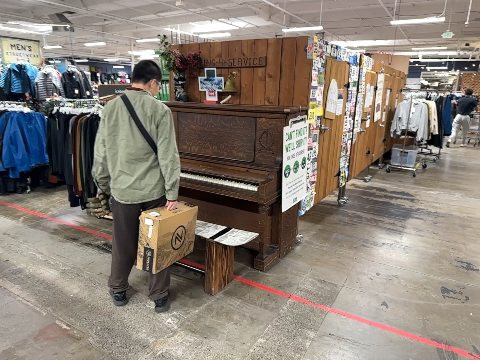} &
  \includegraphics[width=\sz\linewidth]{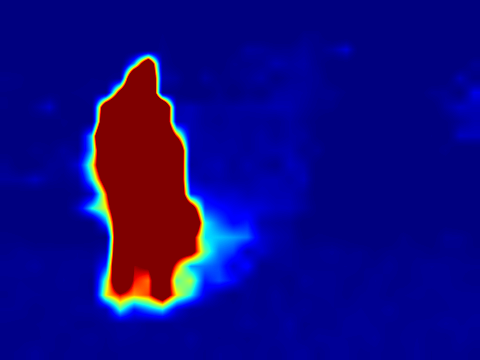}   &
  \includegraphics[width=\sz\linewidth]{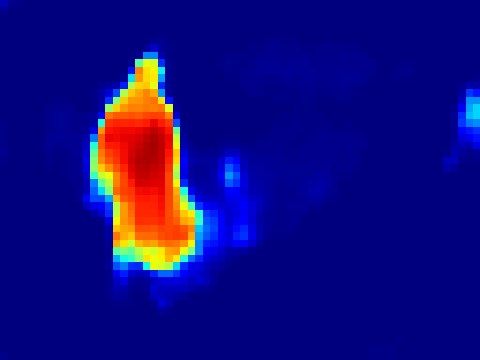} 
  &
  \includegraphics[width=\sz\linewidth]{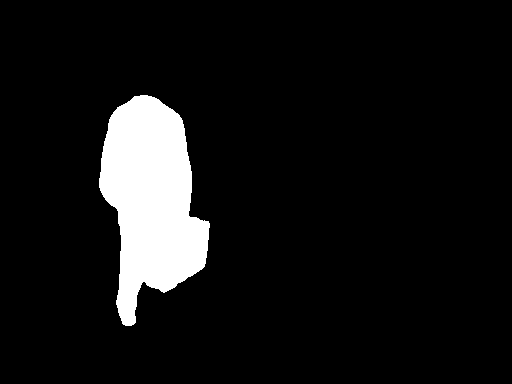} 
  \\
  \raisebox{2.5\normalbaselineskip}[0pt][0pt]{\rotatebox[origin=c]{90}{Street}} &
  \includegraphics[width=\sz\linewidth]{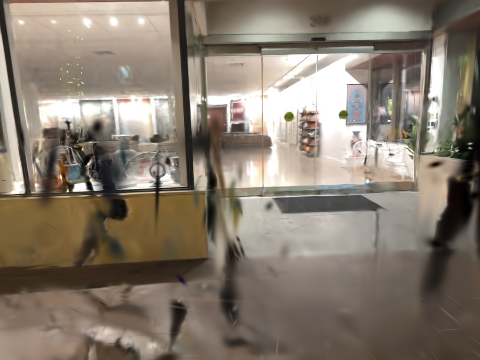} &
  \includegraphics[width=\sz\linewidth]{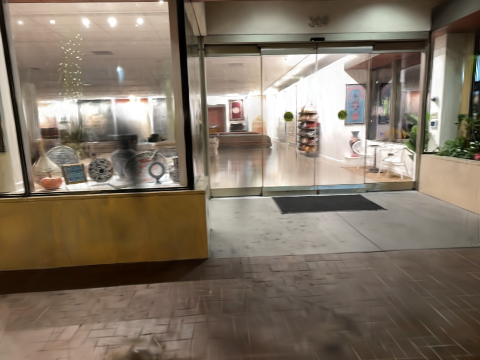} &
  \includegraphics[width=\sz\linewidth]{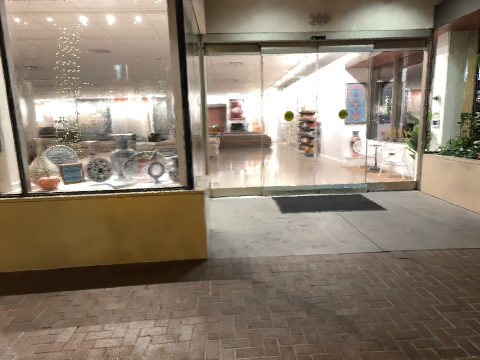} &
  \includegraphics[width=\sz\linewidth]{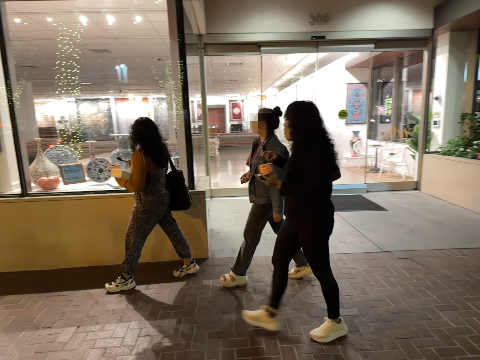} &
  \includegraphics[width=\sz\linewidth]{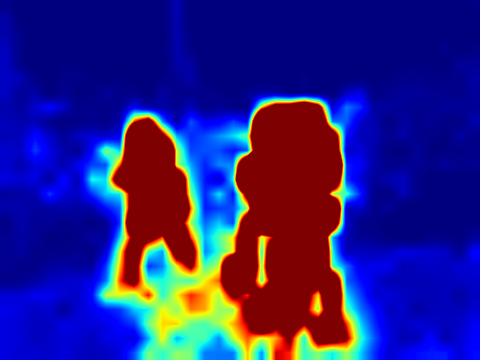}   &
  \includegraphics[width=\sz\linewidth]{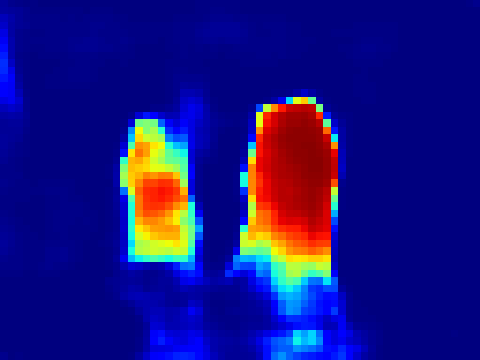} &
  \includegraphics[width=\sz\linewidth]{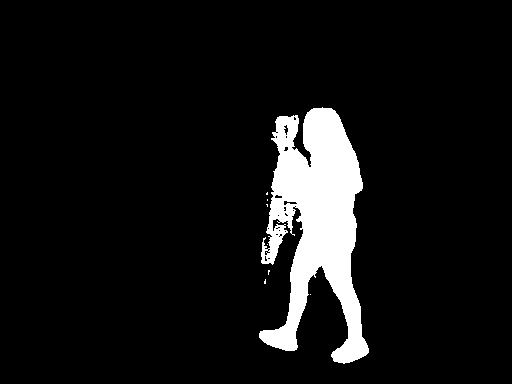} 
  \\
  \raisebox{2.5\normalbaselineskip}[0pt][0pt]{\rotatebox[origin=c]{90}{Tower}} &
  \includegraphics[width=\sz\linewidth]{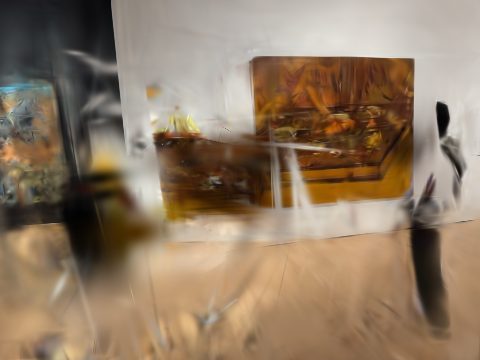} &
  \includegraphics[width=\sz\linewidth]{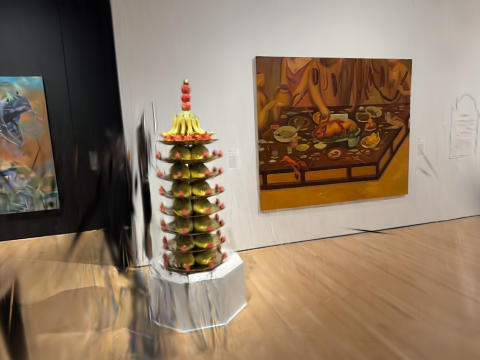} &
  \includegraphics[width=\sz\linewidth]{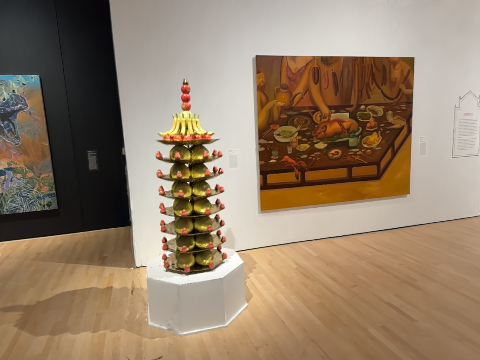} &
  \includegraphics[width=\sz\linewidth]{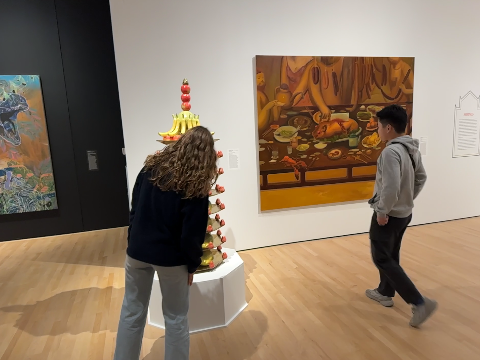} &
  \includegraphics[width=\sz\linewidth]{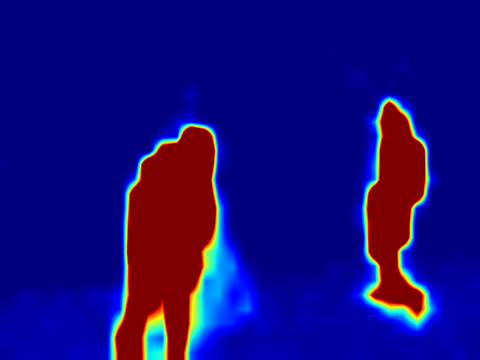}   &
  \includegraphics[width=\sz\linewidth]{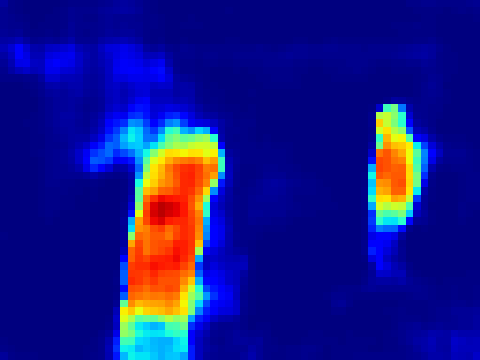} &
  \includegraphics[width=\sz\linewidth]{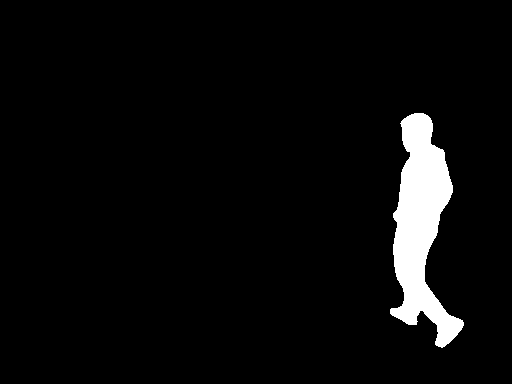}  
  \\

  &  {\fontsize{7}{8} \selectfont MonoGS~\cite{matsuki2024gaussian}} & {\fontsize{7}{8} \selectfont Splat-SLAM~\cite{sandstrom2024splat}} & {\fontsize{7}{8} \selectfont \textbf{\project{} (\ours)}}  &  {\fontsize{7}{8} \selectfont Input} & {\fontsize{7}{8} \selectfont Uncertainty $\beta$ (\textbf{\ours})}& {\fontsize{7}{8} \selectfont 
 MegaSaM~\cite{li2024megasam} Mask } & {\fontsize{7}{8} \selectfont MonST3R~\cite{zhang2024monst3r} Mask }\\
     
  \end{tabular}  
 \vspace{-2mm}
  \caption{\textbf{\textit{Input} View Synthesis Results on our Wild-SLAM iPhone Dataset.} We only show rendering results of monocular methods, as depth images are unavailable in this dataset. 
  Note that our uncertainty map appears blurry, as DINOv2 outputs feature maps at 1/14 of the original resolution, and for mapping we also downsample to 1/3 of the original resolution, in order to maintain SLAM system efficiency. For a high-resolution, sharper uncertainty map, the resolution can be increased at the cost of some efficiency; further details and results are provided in the supplementary materials. 
  Faces are blurred to ensure anonymity.
  }
  \label{fig:iphone_rendering}
\end{figure*} 

\vspace{1pt} \noindent

\subsection{Experimental Setup}
\paragraph{Datasets} 
We evaluate our approach on the Bonn RGB-D Dynamic Dataset~\cite{palazzolo2019iros} and TUM RGB-D Dataset~\cite{sturm2012benchmark}.
To further assess performance in unconstrained, real-world settings, we introduce the Wild-SLAM Dataset, comprising two subsets: Wild-SLAM MoCap and Wild-SLAM iPhone. The Wild-SLAM MoCap Dataset includes 10 RGB-D sequences recorded with an Intel RealSense D455 camera~\cite{intel_d455} in a room equipped with an OptiTrack~\cite{optitrack} motion capture system, providing ground truth trajectories. The Wild-SLAM iPhone Dataset comprises 7 non-staged RGB sequences recorded with an iPhone 14 Pro. Since ground truth trajectories are not available for this dataset, it is used solely for qualitative experiments.
The Wild-SLAM MoCap Dataset provides RGB-D frames at 720 $\times$ 1280 resolution, while the Wild-SLAM iPhone Dataset offers RGB frames at 1440 $\times$ 1920. For efficiency, in our experiments, we downsample these to 360 $\times$ 480 and 360 $\times$ 640, respectively.
Dataset details are offered in supplementary. 

\paragraph{Baselines}
We compare \project{} with the following 13 methods.
(a) \textit{Classic SLAM methods:}  DSO~\cite{Engel2017PAMI}, ORB-SLAM2~\cite{Mur2017TRO}, and DROID-SLAM~\cite{teed2021droid};
(b) \textit{Classic SLAM methods dealing with dynamic environments:} Refusion~\cite{palazzolo2019iros} and DynaSLAM~\cite{bescos2018dynaslam}; 
(c) \textit{Static neural implicit and 3DGS SLAM systems:} NICE-SLAM~\cite{Zhu2022CVPR}, MonoGS~\cite{matsuki2024gaussian}, and Splat-SLAM ~\cite{sandstrom2024splat};
(d) \textit{Concurrent neural implicit and 3DGS SLAM systems dealing with dynamic environments:} DG-SLAM~\cite{xu2024dgslam}, RoDyn-SLAM~\cite{jiang2024rodyn}, DDN-SLAM~\cite{li2024ddn}, and DynaMoN~\cite{schischka2023dynamon}; (e) the very recent \textit{feed-forward approach MonST3R}~\cite{zhang2024monst3r};
and (f) a concurrent deep SLAM framework for dynamic videos: MegaSaM~\cite{li2024megasam}. 
To address its substantial VRAM usage (65 frames requiring 33 GB), we adapt the model by integrating a custom sliding-window inference strategy, enabling SLAM-style sequential input processing  (referred to as MonST3R-SW; implementation details provided in the supplementary).
We include two versions of DynaSLAM~\cite{bescos2018dynaslam} in our experiments. The first, DynaSLAM (RGB), uses only monocular input without leveraging geometric information or performing inpainting. The second, DynaSLAM (N+G), utilizes RGB-D input, incorporates geometric information, and performs inpainting.
Since not all methods are open-sourced and run on all sequences, we provide detailed sources for each baseline method's metrics in supplementary.

\paragraph{Metrics} 
For camera tracking evaluation, we follow the standard monocular SLAM pipeline, aligning the estimated trajectory to the ground truth (GT) using \texttt{evo}~\cite{grupp2017evo} with Sim(3) Umeyama alignment~\cite{umeyama1991least}, and then evaluate (\textit{ATE RMSE})~\cite{sturm2012benchmark}.
Note that, although our tracking optimizations are performed solely on keyframe images, we recover camera poses for non-keyframes and evaluate the complete camera trajectory. Please refer to the supplementary for details.
Additionally, we employ PSNR, SSIM~\cite{ssim}, and LPIPS~\cite{lpips} metrics to evaluate the novel view synthesis quality.

\subsection{Mapping, Tracking, and Rendering}\label{sec:exp_evaluation}

\begin{table*}[t]
\centering
\footnotesize
\setlength{\tabcolsep}{4.5pt}
\resizebox{1.55\columnwidth}{!}
{
\begin{tabular}{lrrrrrrrrr}
\toprule
Method & \texttt{Balloon} & \texttt{Balloon2} & \texttt{Crowd} & \texttt{Crowd2} & \texttt{Person} & \texttt{Person2} & \texttt{Moving} & \texttt{Moving2} & Avg.\\
\midrule
\multicolumn{10}{l}{\cellcolor[HTML]{EEEEEE}{\textit{RGB-D}}} \\ 
ReFusion~\cite{palazzolo2019iros} & 17.5 & 25.4 & 20.4 & 15.5 & 28.9 & 46.3 & 7.1 & 17.9&22.38\\
ORB-SLAM2~\cite{Mur2017TRO} & 6.5 & 23.0 & 4.9 & 9.8 & 6.9 & 7.9 & 3.2 & 3.9 & 6.36\\
DynaSLAM (N+G)~\cite{bescos2018dynaslam}& \rd3.0& 2.9&\nd1.6&\rd3.1&6.1&7.8&23.2&3.9&6.45\\
NICE-SLAM~\cite{Zhu2022CVPR} & 24.4 & 20.2 & 19.3 & 35.8 & 24.5 & 53.6 & 17.7 & 8.3 & 22.74\\
DG-SLAM~\cite{xu2024dgslam} & 3.7 & 4.1 & - & - & 4.5 & 6.9 & - & 3.5 & - \\
RoDyn-SLAM~\cite{jiang2024rodyn}& 7.9 & 11.5 & - & - & 14.5 &13.8 & - &12.3 & - \\
DDN-SLAM (RGB-D)~\cite{li2024ddn} &\fs1.8&4.1&\rd1.8&\fs 2.3&4.3&3.8&2.0&3.2&\nd2.91\\
\hdashline
\noalign{\vskip 1pt}
\multicolumn{10}{l}{\cellcolor[HTML]{EEEEEE}{\textit{Monocular}}} \\ 
DSO~\cite{Engel2017PAMI}& 7.3 & 21.8 & 10.1 & 7.6 & 30.6 & 26.5 & 4.7 & 11.2 & 15.0\\
DROID-SLAM~\cite{teed2021droid} & 7.5 & 4.1 & 5.2 & 6.5 & 4.3 & 5.4 & 2.3 & 4.0 & 4.91\\
MonoGS~\cite{matsuki2024gaussian}& 15.3 & 17.3 & 11.3 & 7.3 & 26.4 & 35.2 & 22.2 & 47.2 & 22.8\\
Splat-SLAM~\cite{sandstrom2024splat}& 8.8 & 3.0 & 6.8 & F & 4.9 & 25.8 & 1.7 & 3.0 & -\\
DynaMoN (MS)~\cite{schischka2023dynamon} & 6.8 & 3.8 & 6.1&5.6&\fs2.4&\rd3.5&\nd1.4&\nd2.6&4.02\\
DynaMoN (MS\&SS)~\cite{schischka2023dynamon} & 
\nd 2.8&\rd2.7&3.5&\nd2.8&14.8&\fs 2.2&\fs1.3&\rd2.7&4.10\\
MonST3R-SW~\cite{zhang2024monst3r} & 5.4 & 7.2 & 5.4 & 6.9 & 11.9 & 11.1 & 3.3 & 7.4 & 7.3\\
MegaSaM~\cite{li2024megasam} & 3.7 & \nd 2.6 & \nd 1.6 & 7.2 & \rd 4.1 & 4.0 & \nd 1.4 & 3.4 & 
 \rd 3.51\\
{\textbf{\project{} (\ours)}} & \nd 2.8 & \fs 2.4 & \fs 1.5 & \fs 2.3 & \nd 3.1 & \nd 2.7 & \rd 1.6 & \fs 2.2 & \fs 2.31\\

\bottomrule
\end{tabular}
}
\vspace{-2mm}
\caption{\textbf{Tracking Performance on Bonn RGB-D Dynamic Dataset~\cite{palazzolo2019iros}} (ATE RMSE $\downarrow$ [cm]). 
DDN-SLAM~\cite{li2024ddn} is not open source and does not report its RGB mode results on this dataset. DynaSLAM (RGB)~\cite{bescos2018dynaslam} consistently fails to initialize or experiences extended tracking loss across all sequences and therefore cannot be included in the table. `F' indicates failure.} 
\label{tab:bonn_tracking}
\vspace{-10pt}
\end{table*}

\begin{figure*}[t!]
  \centering
  \footnotesize
  \setlength{\tabcolsep}{2pt}
  \newcommand{\sz}{0.157}
  \newcommand{\sza}{0.18} %
  \begin{tabular}{cc:ccc:c}

    \includegraphics[width=\sz\linewidth]{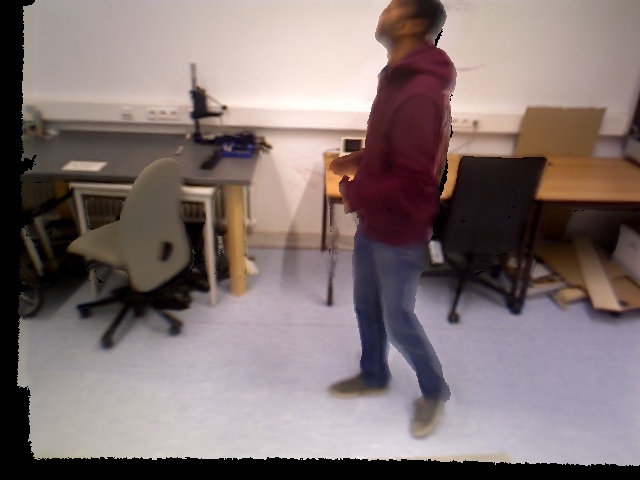} &
    \includegraphics[width=\sz\linewidth]{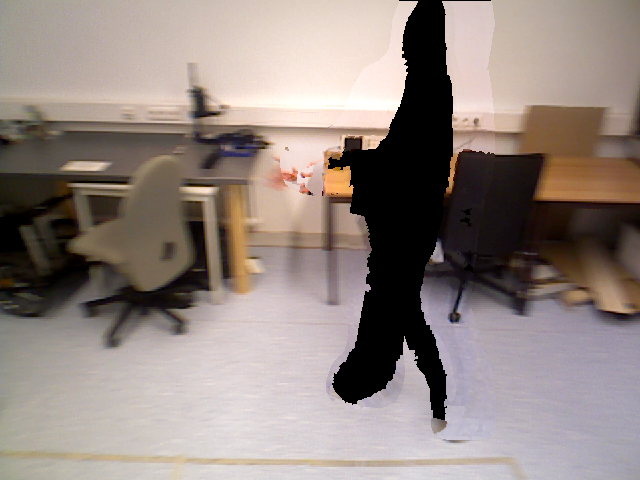} &
    \includegraphics[width=\sz\linewidth]{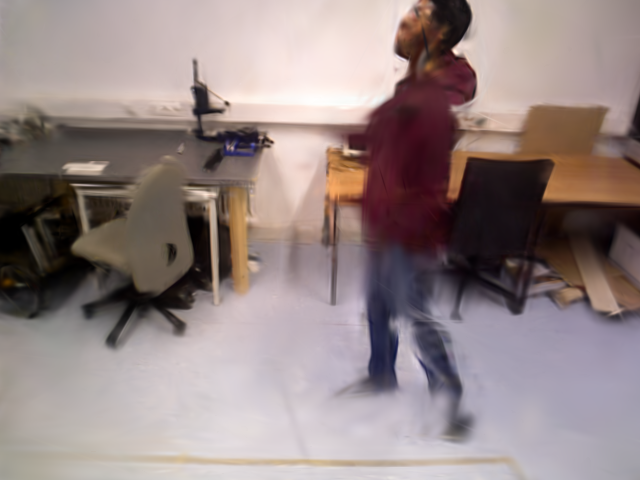} &
    \includegraphics[width=\sz\linewidth]{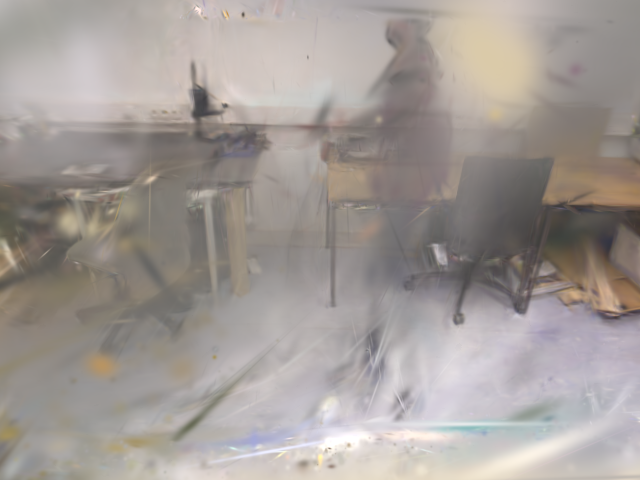} &
    \includegraphics[width=\sz\linewidth]{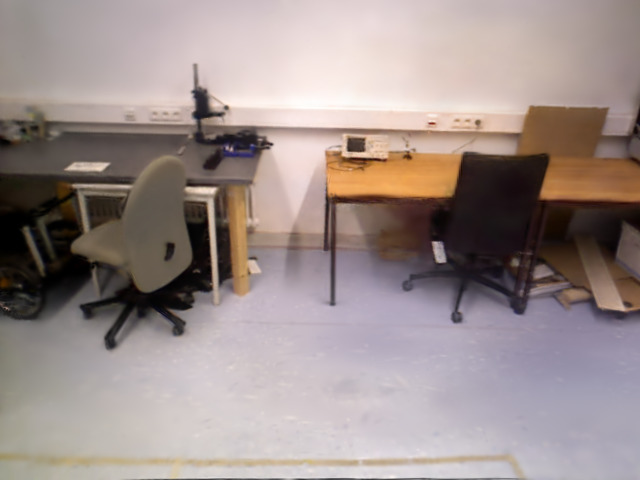}  &
    \includegraphics[width=\sz\linewidth]{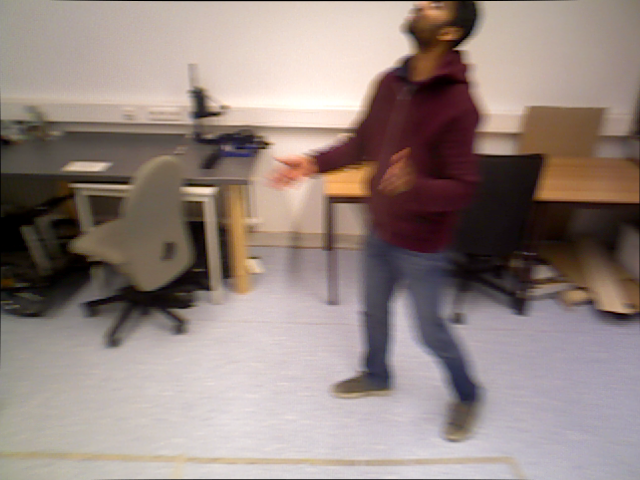} 
    \\
    \includegraphics[width=\sz\linewidth]{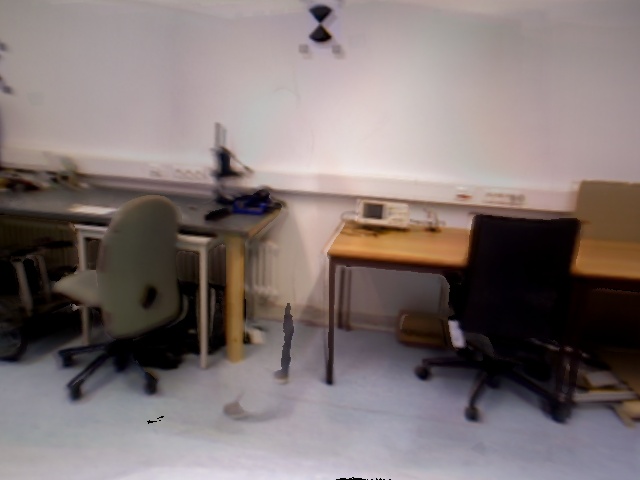} &
    \includegraphics[width=\sz\linewidth]{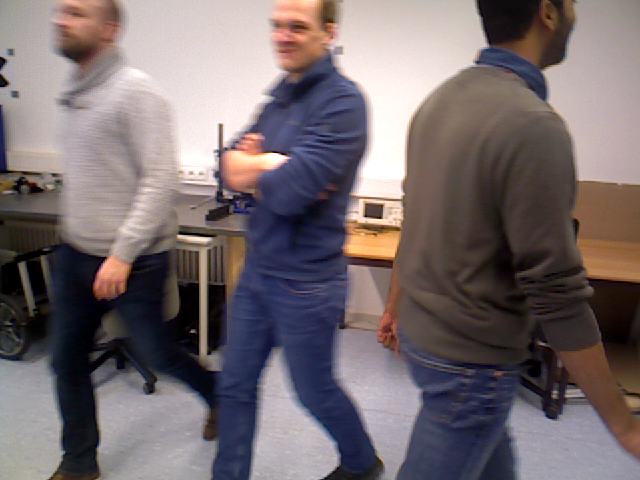} &
    \includegraphics[width=\sz\linewidth]{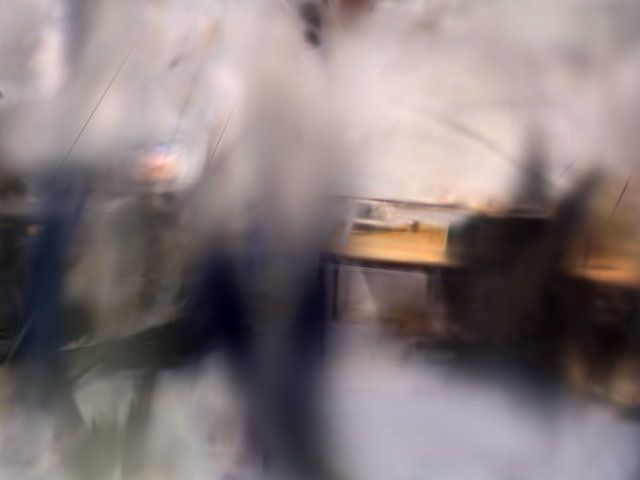} &
    \includegraphics[width=\sz\linewidth]{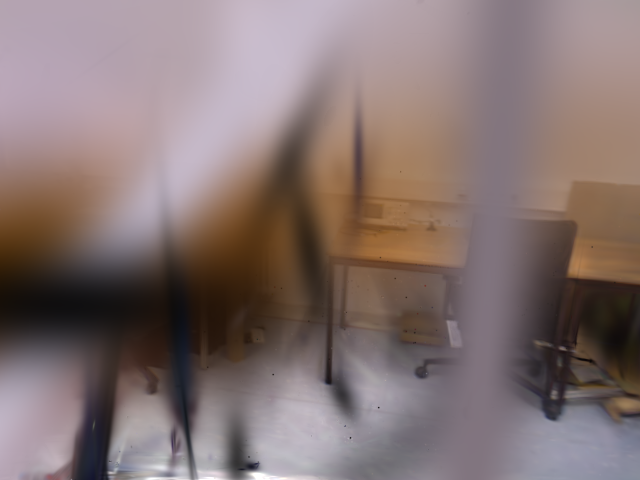} &
    \includegraphics[width=\sz\linewidth]{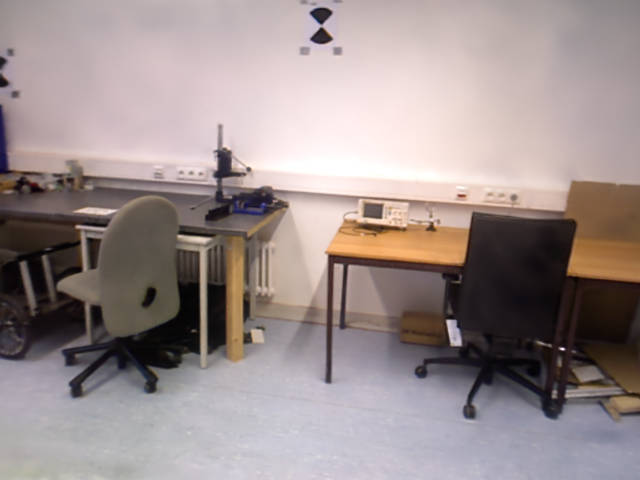}  &
    \includegraphics[width=\sz\linewidth]{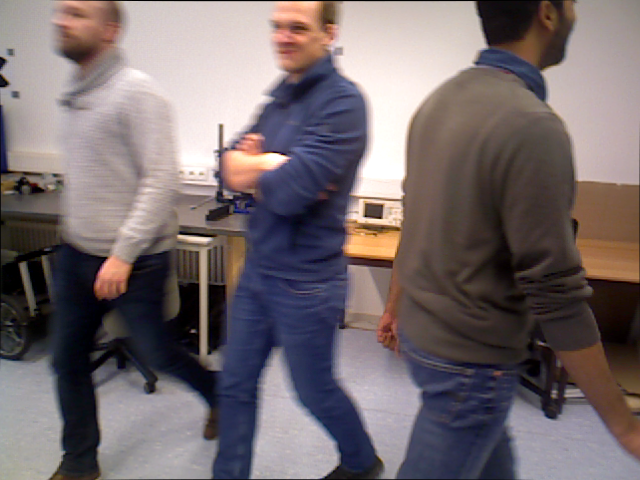} 
    \\
    {ReFusion~\cite{palazzolo2019iros}}   & {DynaSLAM (N+G)~\cite{bescos2018dynaslam}}  & {MonoGS~\cite{matsuki2024gaussian}} & {Splat-SLAM~\cite{sandstrom2024splat}} & {\textbf{\project{} (\ours)}}  &  {Input} \\

     \multicolumn{2}{c:}{\textbf{\textit{RGB-D input}}} & \multicolumn{3}{c:}{\textbf{\textit{Monocular input}}} & \\ 
  \end{tabular} 
  \vspace{-2mm}
  \caption{\textbf{View Synthesis Results on Bonn RGB-D Dynamic Dataset~\cite{palazzolo2019iros}.} We show results on the \texttt{Balloon} (first row) and \texttt{Crowd} (second row) sequences. 
  For \texttt{Balloon}, ReFusion~\cite{palazzolo2019iros} fails to remove the person from the TSDF, and DynaSLAM(N+G)\cite{bescos2018dynaslam} struggles with limited static information from multiple views, resulting in partial black masks. In \texttt{Crowd}, DynaSLAM(N+G)\cite{bescos2018dynaslam} cannot detect dynamic regions, defaulting the original image as the inpainted result. In contrast, ours achieves superior rendering even with motion blur in the input.
  }
  
  \label{fig:bonn_rendering}
\vspace{-10pt}
\end{figure*}

\begin{table}[t]
\centering
\footnotesize
\resizebox{\columnwidth}{!}
{
\begin{tabular}{lrrrrrrrr}
\toprule
Method & \texttt{f3/ws} & \texttt{f3/wx} & \texttt{f3/wr} & \texttt{f3/whs} & Avg.\\
\midrule
\multicolumn{6}{l}{\cellcolor[HTML]{EEEEEE}{\textit{RGB-D}}} \\ 
ReFusion~\cite{palazzolo2019iros} & 1.7 & 9.9 & 40.6* & 10.4 & 15.7*\\
ORB-SLAM2~\cite{Mur2017TRO} & 40.8 & 72.2 & 80.5 & 72.3 & 66.45\\
DynaSLAM (N+G)~\cite{bescos2018dynaslam} & \nd 0.6 & \rd 1.5 & \rd 3.5 & 2.5 & \rd 2.03\\
NICE-SLAM~\cite{Zhu2022CVPR} & 79.8 & 86.5 & 244.0 & 152.0 & 140.57\\
DG-SLAM~\cite{xu2024dgslam} & \nd 0.6 & 1.6 & 4.3 & - & - \\
RoDyn-SLAM~\cite{jiang2024rodyn} & 1.7 & 8.3 & - & 5.6 & -\\
DDN-SLAM (RGB-D)~\cite{li2024ddn} & 1.0 & \nd 1.4 & 3.9 & 2.3 & 2.15\\
\hdashline
\noalign{\vskip 1pt}
\multicolumn{6}{l}{\cellcolor[HTML]{EEEEEE}{\textit{Monocular}}} \\ 
DSO~\cite{Engel2017PAMI} & 1.5 & 12.9 & 13.8 & 40.7 & 17.23\\
DROID-SLAM~\cite{teed2021droid} & 1.2 & 1.6 & 4.0 & 2.2 & 2.25\\
MonoGS~\cite{matsuki2024gaussian} & 1.1 & 21.5 & 17.4 & 44.2 & 21.05\\
Splat-SLAM~\cite{sandstrom2024splat} & 2.3 & \fs 1.3 &  3.9 & 2.2 & 2.43\\
DynaMoN (MS) & 1.4 & \nd 1.4 &  3.9 &  2.0 & 2.18\\
DynaMoN (MS\&SS)~\cite{schischka2023dynamon} & \rd 0.7 & \nd 1.4 & 3.9 & \rd 1.9 & \nd 1.98\\
DDN-SLAM (RGB)~\cite{li2024ddn} & 2.5 & 2.8 & 8.9 & 4.1 & 4.58\\
MonST3R-SW~\cite{zhang2024monst3r} & 2.2 & 27.3 & 13.6 & 19.8 & 15.73 \\
MegaSaM~\cite{li2024megasam} & \nd 0.6 & \rd 1.5 & \fs 2.6 & \nd 1.8 & \fs 1.63\\
{\textbf{\project{} (\ours)}} & \fs 0.4 & \fs 1.3 & \nd 3.3 & \fs 1.6 & \fs 1.63\\ 
\bottomrule
\end{tabular}

}
\vspace{-2mm}
\caption{\textbf{Tracking Performance on TUM RGB-D Dataset~\cite{sturm2012benchmark}} (ATE RMSE $\downarrow$ [cm]). 
We present sequences with higher dynamics here; see the supplementary materials for tracking results on other dynamic sequences.
For methods without complete scene coverage in the original reports, results obtained by running their open-source code are marked with `*'. If open-source code is unavailable, scenes without results are marked with `-'.
DynaSLAM (RGB)~\cite{bescos2018dynaslam} consistently fails to initialize or experiences extended tracking loss across all sequences and therefore cannot be included in this table.
} 
\label{tab:tum_tracking}
\vspace{-15pt}
\end{table}

\noindent \textbf{Wild-SLAM Mocap Dataset.}
We begin our evaluation on the newly captured Wild-SLAM MoCap dataset. As shown in \tabref{tab:mocap_tracking}, our method significantly outperforms other baselines on average.
The only exception is a slight increase in tracking error in the \texttt{Person} sequence, where the single person moves in a simple pattern, making it easier for DynaSLAM~\cite{bescos2018dynaslam} to use semantic segmentation to mask out dynamic regions within the frames.
All other sequences contain various types of distractors beyond humans, with different forms of occlusion.
Although Refusion~\cite{palazzolo2019iros} and DynaSLAM (N+G)~\cite{bescos2018dynaslam} use geometric approaches leveraging raw depth information to identify dynamic objects, they still underperform compared to our \project{} with only monocular input.
While MonST3R~\cite{zhang2024monst3r} has demonstrated strong performance on short sequences, extending it to longer sequences with a sliding window approach (MonST3R-SW), even with substantial overlap, still leads to significant tracking errors.

We further evaluate rendering results in \figref{fig:mocap_rendering}, \tabref{tab:mocap_rendering}, and \figref{fig:mocap_nvs_rendering}. In \figref{fig:mocap_rendering},
we render from the input view to demonstrate our distractor removal capability. In comparison to other baselines, our method produces artifact-free, realistic renderings of the static scene. 
To evaluate novel view synthesis, we capture additional images of static scenes as part of the dataset. For each dynamic sequence, we select a subset of static scene images that match the coverage of the dynamic sequence.
The results in \tabref{tab:mocap_rendering} and \figref{fig:mocap_nvs_rendering} showcase that our method has the best novel view synthesis performance thanks to our uncertainty aware mapping.

\paragraph{Wild-SLAM iPhone Dataset}
We further evaluate our approach on in-the-wild sequences captured using an iPhone RGB camera. \figref{fig:iphone_rendering} presents rendering results comparisons, along with visualizations of the uncertainty map of our method and the dynamic mask from MonST3R~\cite{zhang2024monst3r}. Our method achieves the best rendering results with an accurate uncertainty map, even able to assign higher uncertainty to the shadows of distractors. 
In contrast, MonST3R~\cite{zhang2024monst3r} depends heavily on the performance of pretrained models, which may lead to missed detections of entire dynamic objects. 
On the other hand, MegaSAM~\cite{li2024megasam} leverages only neighboring frames, lacking enough multi-view information, which results in less reliable motion masks.

\vspace{1pt} \noindent
\textbf{Bonn RGB-D Dynamic Dataset~\cite{palazzolo2019iros}.} Tracking results, presented in \tabref{tab:bonn_tracking}, demonstrate that our method achieves the highest overall performance, underscoring its robustness. In contrast, DynaMoN~\cite{schischka2023dynamon} first performs initial camera tracking and then refines it offline, which is time-consuming. Meanwhile, DynaSLAM (N+G)~\cite{bescos2018dynaslam} and DDN-SLAM~\cite{li2024ddn} rely on depth sensor data combined with semantic segmentation. In \figref{fig:bonn_rendering}, we show rendered images from the input view. Our method successfully removes dynamic objects while achieving realistic rendering with minimal artifacts. 

\vspace{1pt} \noindent
\textbf{TUM RGB-D Dataset~\cite{sturm2012benchmark}.}
Our tracking method achieves the best performance on all sequences, as shown in \tabref{tab:tum_tracking}. Rendering results are included in the supplementary.

\subsection{Ablation Study}
\vspace{1pt} \noindent
We ablate our key design choices in \tabref{tab:ablation}.
For (c), we pass predefined distractor types to YOLOv8\footnote{\url{https://github.com/ultralytics/ultralytics.git}} to generate bounding boxes, then apply the Segment Anything Model (SAM)~\cite{kirillov2023segment} for segmentation within each box. 
It achieves similar results on Bonn and TUM datasets, as their distractors are primarily humans.
Our method outperforms all other variants, confirming the effectiveness of our design choices.

\begin{table}[t]
    \centering
    \footnotesize
    \newcommand{\sz}{0.195}
    \newcommand{\sza}{0.1729} 
    \setlength{\tabcolsep}{3pt}
    \resizebox{0.8\linewidth}{!}{
    \begin{tabular}{lrrrr}
    \toprule
& \texttt{Wild-SLAM} & \texttt{Bonn} & \texttt{TUM} & \\
\midrule
(a) w/o Uncertainty Mask $\beta$ & 3.89 & \rd5.11 & 1.91 \\
(b) w/o L1 Depth Loss in \eqnref{eq:uncer_loss} & \nd0.50 & \nd2.37 & \rd1.83\\
(c) YOLOv8 + SAM Mask & \rd 3.06 & \nd2.37 & \nd 1.65\\
(d) w/o Disparity Reg. in \eqnref{eq:dba} & 10.97 & F & 2.9\\
{\textbf{\project{} (\ours)}}  & \fs0.46 & \fs2.31 & \fs 1.63 \\ 
\bottomrule
\end{tabular}}
\vspace{-2mm}
\caption{\textbf{\project{} Ablation Study} (ATE RMSE $\downarrow$ [cm]). For each dataset, we report the average tracking error. 
`F' indicates that the method fails on at least one sequence within that dataset.
}
\label{tab:ablation}
\vspace{-15pt}
\end{table}

\section{Conclusion}
\label{sec:conclusions}
In this paper, we introduced \project{}, a novel SLAM approach designed to handle dynamic environments through a purely geometric framework. By leveraging a shallow MLP to predict per-pixel uncertainty based on pre-trained 3D-aware features, our method efficiently isolates static and dynamic scene elements, enabling robust tracking and rendering. Through extensive evaluations on both newly collected and existing datasets, we demonstrated that \project{} achieves state-of-the-art performance in dynamic SLAM tasks, excelling in both tracking and novel view synthesis.

\vspace{1pt} \noindent
\textbf{Limitation.} Our method's uncertainty predictor is trained on-the-fly with input frames, making it challenging to recognize distractors when a limited number of views capture the same regions. Introducing motion priors could improve handling of dynamic scenes and enhance tracking robustness.

\vspace{1pt} \noindent
\textbf{Acknowledgements.} 
The authors thank Sayan Deb Sarkar, Tao Sun, Ata Celen, Liyuan Zhu, Emily Steiner, Jikai Jin, Yiming Zhao, Matt VanCleave, Tess Ruby Horowitz Buckley, Stanley Wang for their help in Wild-SLAM data collection. We thank Aleesa Pitchamarn Alexander for granting permission to release the data collected from the \href{https://museum.stanford.edu/exhibitions/spirit-house}{Spirit House} exhibition. We also thank Aleesa Pitchamarn Alexander, Robert M. and Ruth L. Halperin for curating the exhibition, as well as all the participating artists, particularly Dominique Fung, Stephanie H. Shih, and Tammy Nguyen whose art works are prominently captured in the video data.

{
    \small
    \bibliographystyle{ieeenat_fullname}
    \bibliography{main}
}
\clearpage
\setcounter{page}{1}
\maketitlesupplementary

\begin{abstract}
In the supplementary material, we provide additional details about the following:
\begin{enumerate}
    \item More information about the Wild-SLAM dataset (\secref{sec:dataset}).
    \item Implementation details of \project{} and baseline methods (\secref{sec:implementation}).
    \item Additional results and ablations (\secref{sec:add_results}).
\end{enumerate}
\end{abstract}

\section{Wild-SLAM Dataset}
\label{sec:dataset}
\begin{figure}[t]
    \centering
    \begin{subfigure}[b]{0.48\linewidth}
        \centering
        \includegraphics[width=\linewidth]{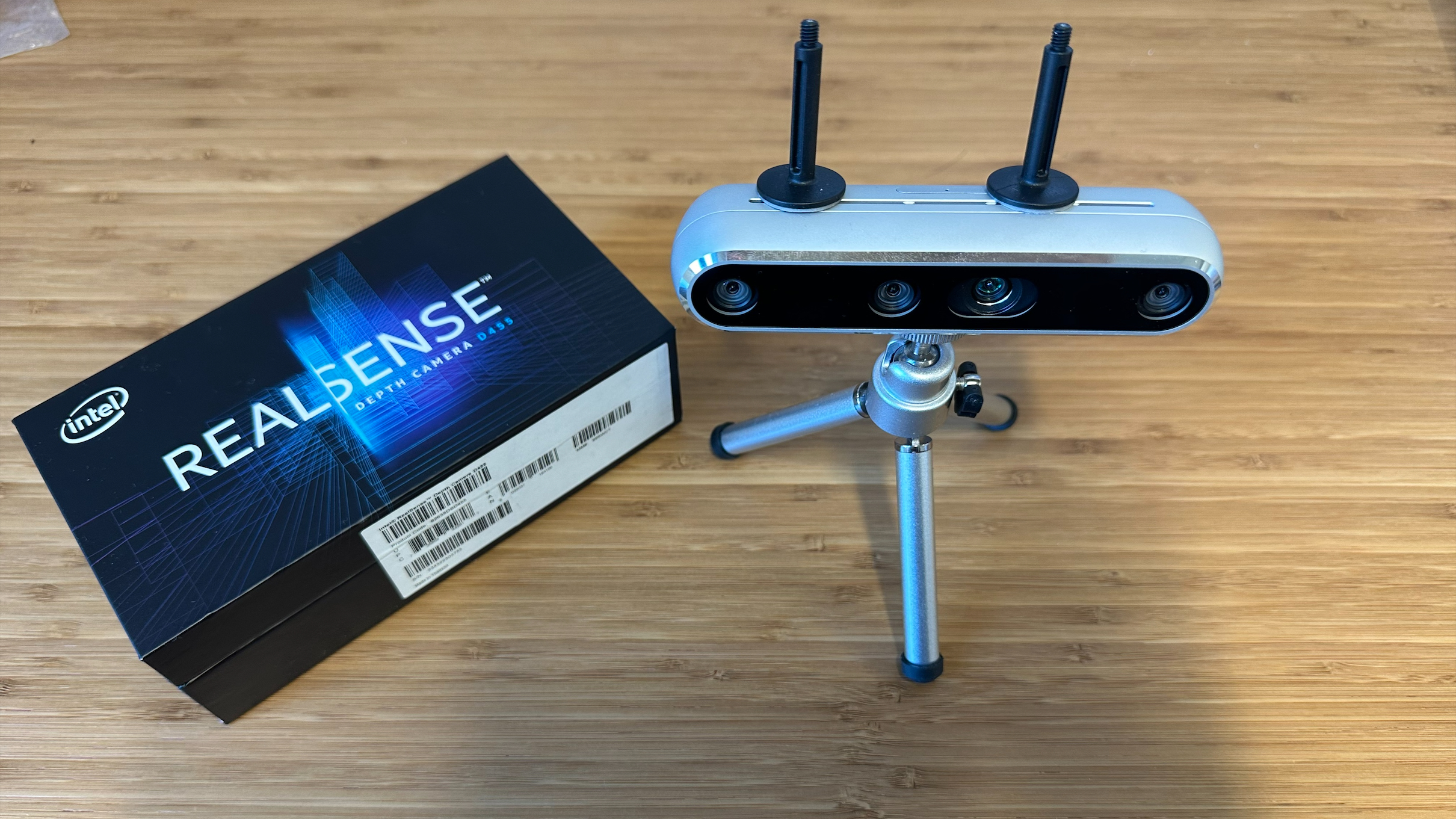} %
        \caption{ }
        \label{fig:d455}
    \end{subfigure}
    \hfill
    \begin{subfigure}[b]{0.48\linewidth}
        \centering
        \includegraphics[width=\linewidth]{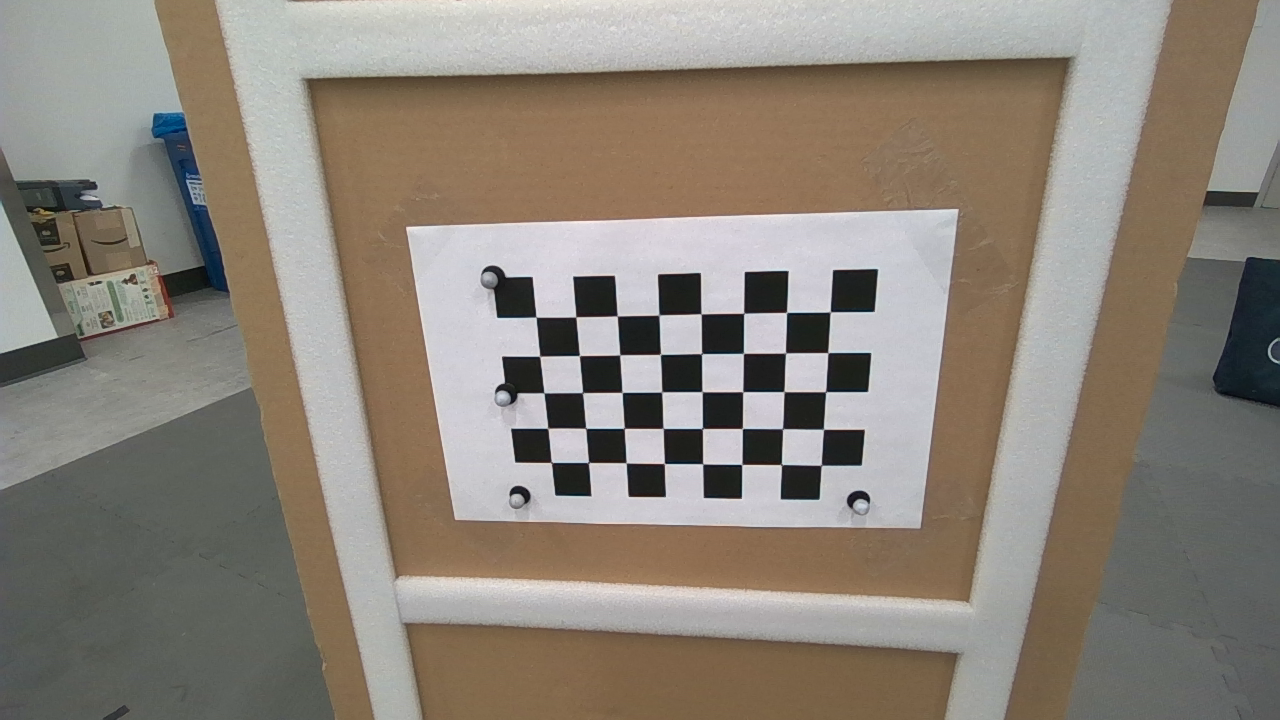} %
        \caption{ }
        \label{fig:calibration_board}
    \end{subfigure}

    \begin{subfigure}[b]{0.48\linewidth}
        \centering
        \includegraphics[width=\linewidth]{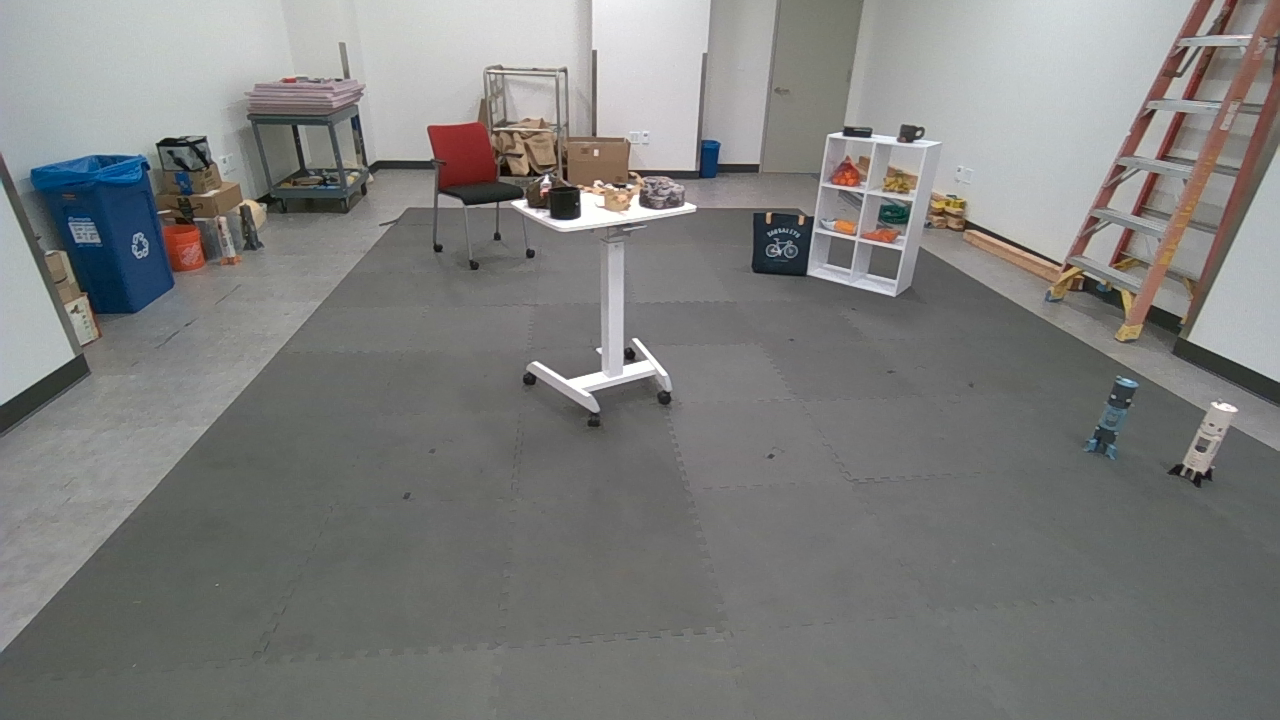} %
        \caption{ }
        \label{fig:scene1}
    \end{subfigure}
    \hfill
    \begin{subfigure}[b]{0.48\linewidth}
        \centering
        \includegraphics[width=\linewidth]{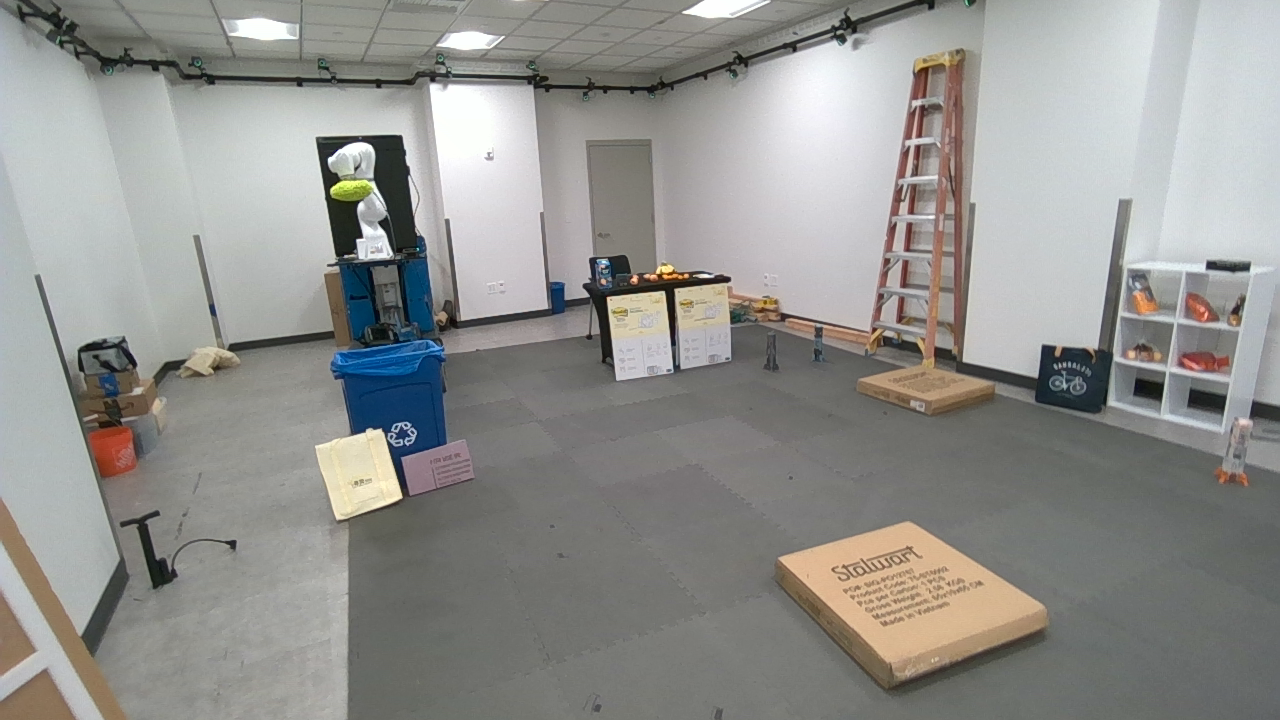} %
        \caption{ }
        \label{fig:scene2}
    \end{subfigure}
    
    \vspace{-0.2cm}
    
    \caption{(a) Intel RealSense D455 camera~\cite{intel_d455}. (b) Calibration board used to align camera reference frame with OptiTrack's rigid body frame. (c) Static scene 1. (d) Static scene 2.}
    \label{fig:mocap_dataset}
\end{figure}

\paragraph{Wild-SLAM MoCap Dataset}
This dataset comprises a total of 10 sequences of RGB-D frames featuring various moving objects as distractors, specifically designed for dynamic SLAM benchmarking. Although \project{} works with monocular inputs, aligned depth images are included to support the evaluation of other RGB-D baselines or future research. The RGB-D frames were captured using an Intel RealSense D455 camera (\figref{fig:d455}) at a resolution of $720 \times 1280$ and a frame rate of 30 fps. All image sequences in this dataset were recorded with a fixed exposure time. 
The dataset includes two distinct static environment layouts: 2 sequences were captured in one static scene (\figref{fig:scene1}), and the remaining 8 sequences were recorded in the other (\figref{fig:scene2}). A summary of each sequence is provided in
\tabref{tab:mocap_dataset}, while all sequences are presented in the video.
The room used for dataset collection was equipped with an OptiTrack motion capture (MoCap) system~\cite{optitrack}, consisting of 32 OptiTrack PrimeX-13 cameras, to provide the ground truth camera poses. The OptiTrack system operates at 120 fps. 

Inspired by~\cite{zhao2024egopressure}, to synchronize the OptiTrack system with the Intel RealSense D455 camera, 
we positioned the RealSense camera and one of the OptiTrack cameras to observe an iPhone. 
By switching the iPhone flashlight on and off, we identified the corresponding timestamps in the images captured by both devices that reflected the flashlight's state change. This allowed us to get the timestamp offset between the two devices.
To improve synchronization accuracy, the frame rate of the D455 was increased to 60 fps.
We switched the flashlight on and off before and after each sequence recording, obtaining four timestamp offset values. The timestamp offset per recording was calculated as their average. 
Across all sequences, the average standard deviation among the four timestamp offset values was 5.25 ms, while the time interval between consecutive frames in the captured sequence is 33.33 ms, highlighting the precision of the synchronization.

To track the pose of the D455 camera using the MoCap system, we attached four reflective markers to the camera, defining a rigid body. We then performed a calibration procedure using a calibration board (\figref{fig:calibration_board}) to determine the relative transformation between the MoCap coordinate system and the camera coordinate system for this rigid body.
Four reflective markers were carefully placed on the calibration board, enabling the MoCap system to track their 3D positions. These positions were then utilized to compute the locations of the grid corners on the board. Meanwhile, the corresponding 2D pixel coordinates of these grid corners in the camera frame were identified using the method described in~\cite{duda2018accurate}. Using this information, the camera poses in the MoCap coordinate system were determined, allowing us to compute the transformation between the rigid body and the camera frame.
The calibration process was repeated 19 times, with the camera and the calibration board positioned in different poses for each trial, resulting in 19 transformation matrices. The final transformation between the rigid body and the camera frame was computed by averaging the results across all trials. Specifically, the rotation component was averaged using the chordal \(L_2\) method~\cite{hartley2013rotation}. The average deviation between an individual estimated transformation matrix and the final averaged transformation is \(0.44^\circ\) for rotation and \(0.24\) cm for translation.

\begin{table*}[h!]
    \centering
    \resizebox{2\columnwidth}{!}
    {
    \begin{tabular}{lccccc}
        \toprule
        \textbf{Sequence Name}         & \textbf{Distractors} & \textbf{Static Environment} & \textbf{Number of Frames} & \textbf{Length of Trajectory [m]}  \\
        \midrule
        \texttt{ANYmal1} & ANYmal Robot & Scene 1 & \phantom{1}651 & \phantom{1}7.274 \\
        \texttt{ANYmal2} & ANYmal Robot & Scene 1 & 1210 & 11.567 \\
        \texttt{Ball} & Human, Basketball & Scene 2 & \phantom{1}931 & 11.759 \\
        \texttt{Crowd} & Human, Basketball, Bag & Scene 2 & 1268 & 14.189 \\
        \texttt{Person} & Human & Scene 2 & \phantom{1}986 & 10.354 \\
        \texttt{Racket} & Human, Racket & Scene 2 & \phantom{1}962 & 12.421 \\
        \texttt{Stones} & Human, Table, Bag, Gripper, Stone & Scene 2 & \phantom{1}962 & 12.421 \\
        \texttt{Table1} & Human, Table, Gripper, Stone & Scene 2 & \phantom{1}561 & \phantom{1}6.592 \\
        \texttt{Table2} & Human, Table, Gripper, Stone & Scene 2 & 1029 & 11.184 \\
        \texttt{Umbrella} & Human, Umbrella & Scene 2 & \phantom{1}458 & \phantom{1}4.499 \\
        \bottomrule
    \end{tabular}
    }
    \caption{\textbf{Overview of our WildGS-SLAM MoCap Dataset.} }

    \label{tab:mocap_dataset}
\end{table*}

\paragraph{Wild-SLAM iPhone Dataset}
To further assess performance in more unconstrained, real-world scenarios, we captured 7 sequences using an iPhone 14 Pro. These sequences comprise 4 outdoor and 3 indoor scenes, showcasing a variety of daily-life activities such as strolling along streets, shopping, navigating a parking garage, and exploring an art museum. Each sequence provides RGB images at a resolution of 1920 $\times$ 1280, accompanied by LiDAR depth images at 256 $\times$ 192 resolution. While \project{} only requires monocular inputs, the inclusion of LiDAR data facilitates the evaluation of RGB-D baselines and future research. All sequences are showcased in the supplementary video.

\paragraph{Discussion} Both datasets capture humans performing activities. The \textit{Wild-SLAM MoCap Dataset} was recorded in controlled environments, with explicit consent obtained from all participants for publishing, presenting, and sharing the data with the research community. In contrast, the \textit{Wild-SLAM iPhone Dataset} was captured in more unconstrained settings, where we had less control over the presence of bystanders in the scene. While consent was obtained from the primary individuals featured, additional people may occasionally appear in the background. In most cases, these individuals are positioned too far from the camera to be identifiable (occupying very few pixels). Additionally, in the Parking sequence, certain car license plates are visible. To ensure privacy, all sensitive regions, including faces and license plates, have been masked in the data. It is important to note that recordings were conducted in locations where capturing people in public spaces is legally permitted, provided the footage does not target individuals in a way that could be considered intrusive or harassing.

\section{Implementation Details}
\label{sec:implementation}

\subsection{\project{}}

\noindent \textbf{Two-Stage Initialization.}
We use the first 12 keyframes to run the DBA layer for tracking initialization. However, the uncertainty MLP $\mathcal{P}$ has not yet been trained to identify uncertain regions. Hence, we deactivate the uncertainty weight $\beta$ in \eqnref{eq:dba} during the first stage of initialization to obtain coarse camera poses. These initial poses are used for map initialization and training of $\mathcal{P}$. Subsequently, we perform a reduced number of iterations in the DBA layer, with uncertainty weighting activated, to refine the coarse keyframe camera poses from the first stage.

\noindent \textbf{Frame Graph Management.} 
We manage the frame graph as in \cite{teed2021droid} but enforce the insertion of a new keyframe every 8 frames, independent of the criterion in \cite{teed2021droid} (average optical flow to the last keyframe larger than a threshold).

\noindent \textbf{Disparity Regularization Mask $\boldsymbol{M}$.}
For each newly inserted keyframe \( i \), we project each of its connected keyframes $j$ in the frame graph, i.e., $ (i, j) \in E$, onto \( i \) using the metric depth \( \tilde{D}_{j} \) and calculate the multi-view depth consistency count as:
\footnotesize
\begin{equation}
\begin{aligned}
n_{i}(u,v) = \sum_{j| \left(i,j\right) \in E} \mathbbm{1}\biggl(&\frac{|\tilde{D}_{i}(u,v) - \tilde{D}_{j\rightarrow i}(u',v')|}{\tilde{D}_{j\rightarrow i}(u',v')} < \epsilon \\ 
&\land \cos\left(F_{i}\left(u,v\right), F_{j}\left(u',v'\right)\right) > \gamma \biggr)
\end{aligned}
\end{equation}
\normalsize where $\mathbbm{1}\left(\cdot\right)$ is the indicator function, $(u',v')$ is the pixel coordinate in $j$th frame that falls to $(u,v)$ when re-projected to frame $i$ using $\tilde{D}_{j}$, $\bomega_{i}$ and $\bomega_{j}$, $\tilde{D}_{j\rightarrow i}(u',v')$ is the projected depth from point $(u',v')$ to frame $i$, and $\epsilon$ is the relative depth threshold. The second condition is to filter out incorrect correspondences that have lower than a threshold $\gamma$ DINO feature cosine similarity.
The depth mask $M_{i}(u,v)$ is set to $0$ if $(u,v)$ has more than one valid correspondence in neighboring frames and $n_{i}(u,v)$ is less than a threshold.
\begin{table*}[h!]
    \centering
    \begin{tabular}{lccccc}
        \toprule
        \textbf{Method}         & \textbf{Input Type} & \textbf{Dynamic} & \textbf{Open Source} & \textbf{Prior Free} & \textbf{Scene Representation} \\
        \midrule
        
        \noalign{\vskip 1pt}
        \multicolumn{6}{l}{\cellcolor[HTML]{EEEEEE}\textit{Classic SLAM methods}} \\
        DSO~\cite{Engel2017PAMI}                     & RGB                 & \xmark                          & \cmark & \cmark & Sparse Point Cloud          \\
        ORB-SLAM2~\cite{Mur2017TRO}               & RGB                 & \xmark                          & \cmark & \cmark & Sparse Point Cloud          \\
        DROID-SLAM~\cite{teed2021droid}              & RGB                 & \xmark                          & \cmark & \cmark & -                           \\
        \hdashline
        \noalign{\vskip 1pt}
        \multicolumn{6}{l}{\cellcolor[HTML]{EEEEEE}\textit{Classic SLAM methods with dynamic environment handling}} \\
        Refusion~\cite{palazzolo2019iros}                & RGB-D               & \cmark                          & \cmark & \cmark & TSDF                        \\
        DynaSLAM (RGB)~\cite{bescos2018dynaslam}          & RGB                 & \cmark                          & \cmark & \xmark(S) & Sparse Point Cloud          \\
        DynaSLAM (N+G)~\cite{bescos2018dynaslam}          & RGB-D               & \cmark                          & \cmark & \xmark(S) & Sparse Point Cloud          \\
        \hdashline
        \noalign{\vskip 1pt}
        \multicolumn{6}{l}{\cellcolor[HTML]{EEEEEE}\textit{Static neural implicit and 3DGS SLAM methods}} \\
        NICE-SLAM~\cite{Zhu2022CVPR}               & RGB-D               & \xmark                          & \cmark & \cmark & Neural Implicit             \\
        MonoGS~\cite{matsuki2024gaussian}                  & RGB/RGB-D                 & \xmark                          & \cmark & \cmark & 3D Gaussian Splatting       \\
        SplatSLAM~\cite{sandstrom2024splat}               & RGB                 & \xmark                          & \cmark & \cmark & 3D Gaussian Splatting       \\
        \hdashline
        \noalign{\vskip 1pt}
        \multicolumn{6}{l}{\cellcolor[HTML]{EEEEEE}\textit{Concurrent neural implicit and 3DGS SLAM methods for dynamic scenes}} \\
        DG-SLAM~\cite{xu2024dgslam}                 & RGB-D               & \cmark                          & \xmark & \xmark(S) & 3D Gaussian Splatting       \\
        RoDyn-SLAM~\cite{jiang2024rodyn}              & RGB-D               & \cmark                          & \xmark & \xmark(S) & Neural Implicit             \\
        DDN-SLAM~\cite{li2024ddn}                & RGB/RGB-D               & \cmark                          & \xmark & \xmark(O) & Neural Implicit             \\
        DynaMoN (MS)~\cite{schischka2023dynamon}                 & RGB               & \cmark                          & \cmark & \cmark & Neural Implicit             \\
        DynaMoN (MS \& SS)~\cite{schischka2023dynamon}                 & RGB               & \cmark                          & \cmark  & \xmark(S) & Neural Implicit             \\
        \hdashline
        \noalign{\vskip 1pt}
        \multicolumn{6}{l}{\cellcolor[HTML]{EEEEEE}\textit{Recent feed-forward methods}} \\
        MonST3R~\cite{zhang2024monst3r}                 & RGB               & \cmark                          & \cmark & \cmark & Dense Point cloud           \\
        \hdashline
        {\textbf{\project{} (\ours)}} & RGB & \cmark & \cmark* & \cmark &  3D Gaussian Splatting \\
        \bottomrule
    \end{tabular}
    \caption{\textbf{Overview of Baseline Methods.} 
`Dynamic' indicates whether the method explicitly addresses dynamic scenes. 
`Open Source' specifies if a public implementation is available. 
`Prior Free' refers to not using class priors, where `O' represents object detection and `S' denotes semantic segmentation. In all our experiments, we employ the RGB mode of MonoGS~\cite{matsuki2024gaussian}.  }

    \label{tab:baseline_methods}
\end{table*}

\paragraph{Final Global BA}
After processing all the input frames, we incorporate a final global Bundle Adjustment (BA) module, similar to DROID-SLAM~\cite{teed2021droid}, to refine the keyframe poses. The frame graph construction follows the same approach as DROID-SLAM~\cite{teed2021droid}. For the DBA objective during tracking, we retain only the first term of \eqnref{eq:dba}, omitting the disparity regularization term, as sufficient multiview information is already available, and the uncertainty map has converged to a stable state. We include an ablation study in \secref{sec:add_results}.

\paragraph{Final Map Refinement}
After the final global BA, we perform a final refinement of the map using all keyframes, following the same strategy as Splat-SLAM~\cite{sandstrom2024splat} and MonoGS~\cite{matsuki2024gaussian}. In the final refinement, we fix the keyframe poses and optimize both the uncertainty MLP and 3D Gaussian map using \eqnref{eq:render_loss}.

\paragraph{Obtaining Non-keyframe Pose}
After completing the final global BA and map refinement, we conduct a motion-only bundle adjustment to estimate non-keyframe poses, similar to the approach in DROID-SLAM~\cite{teed2021droid}. During this optimization, we also deactivate the disparity regularization term in \eqnref{eq:dba}. These poses are further refined using an L1 RGB re-rendering loss, as employed in MonoGS~\cite{matsuki2024gaussian}, weighted by the uncertainty map.

\begin{figure*}[t!]
  \centering
  \footnotesize
  \setlength{\tabcolsep}{0.5pt}
  \newcommand{\sz}{0.16}
  \newcommand{\sza}{0.18} %
  \begin{tabular}{cc:ccc:c}

    \includegraphics[width=\sz\linewidth]{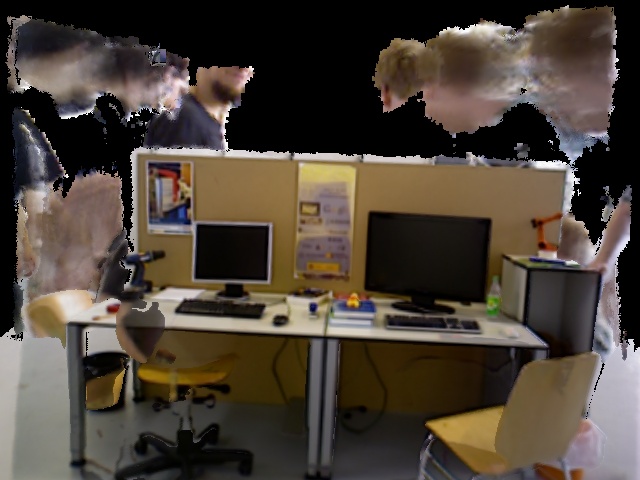} &
    \includegraphics[width=\sz\linewidth]{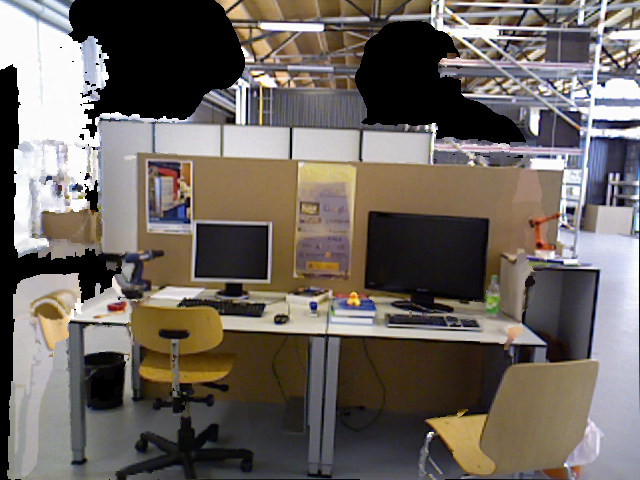} &
    \includegraphics[width=\sz\linewidth]{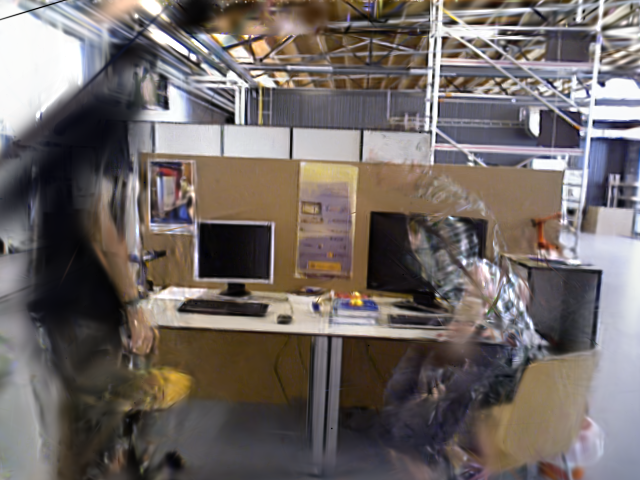} &
    \includegraphics[width=\sz\linewidth]{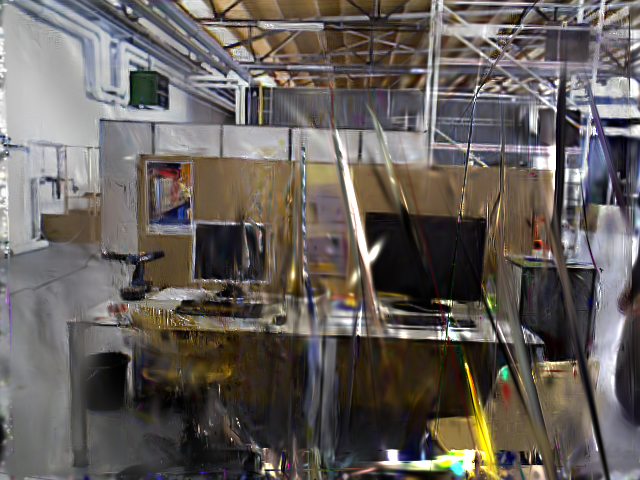} &
    \includegraphics[width=\sz\linewidth]{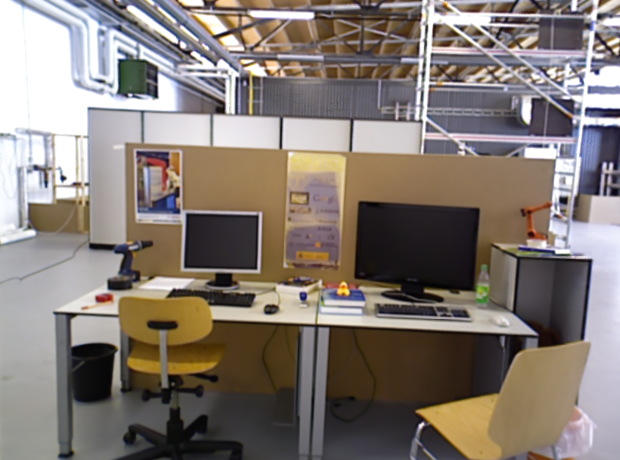}  &
    \includegraphics[width=\sz\linewidth]{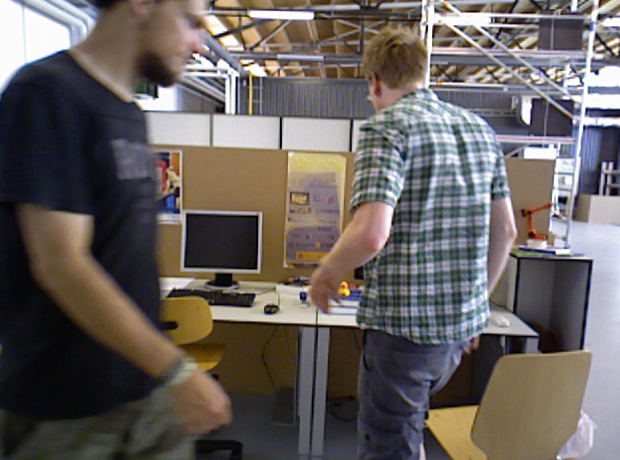} 
    \\
    \includegraphics[width=\sz\linewidth]{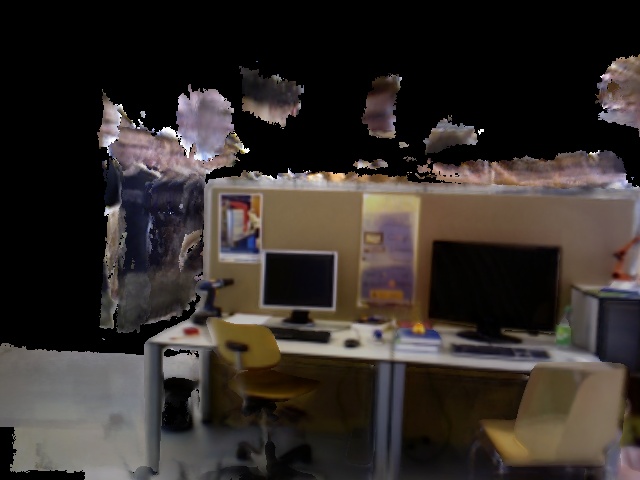} &
    \includegraphics[width=\sz\linewidth]{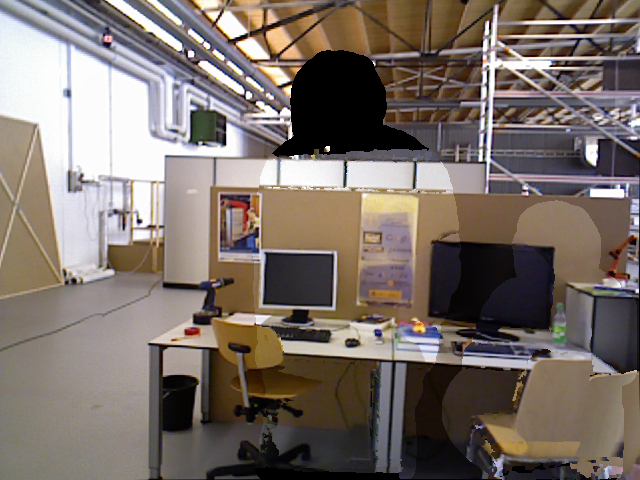} &
    \includegraphics[width=\sz\linewidth]{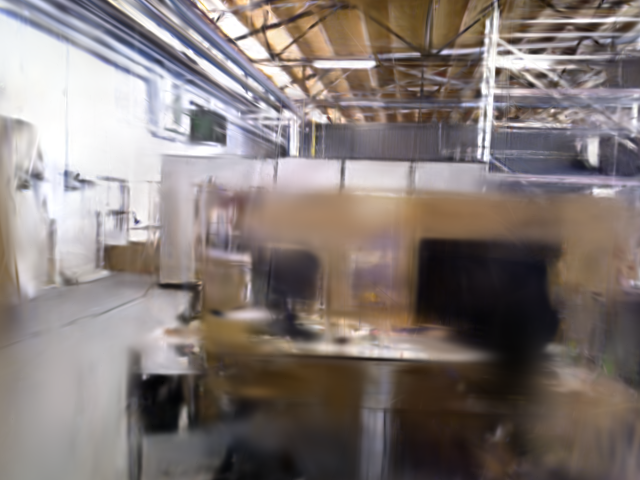} &
    \includegraphics[width=\sz\linewidth]{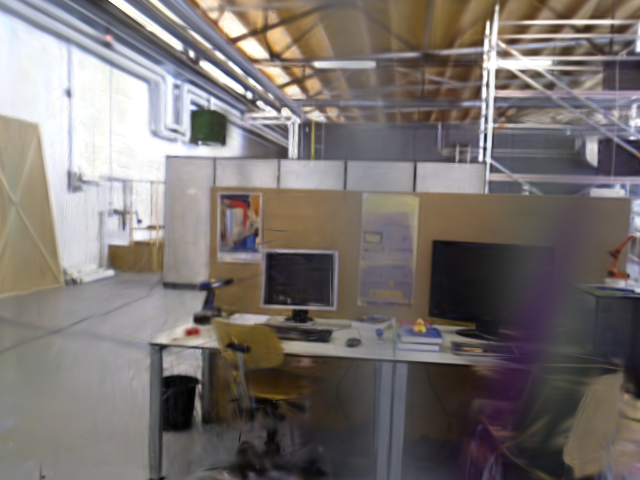} &
    \includegraphics[width=\sz\linewidth]{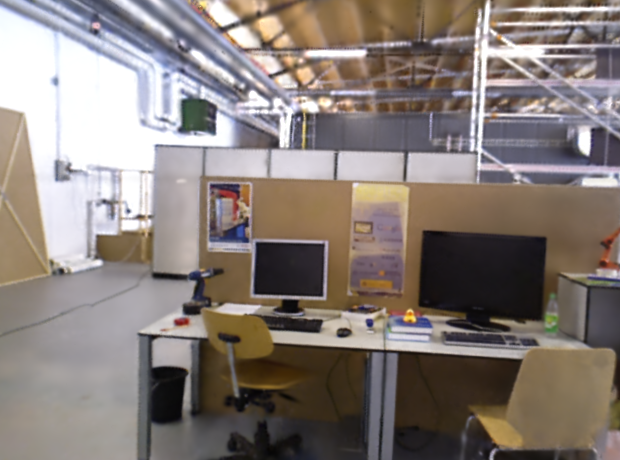}  &
    \includegraphics[width=\sz\linewidth]{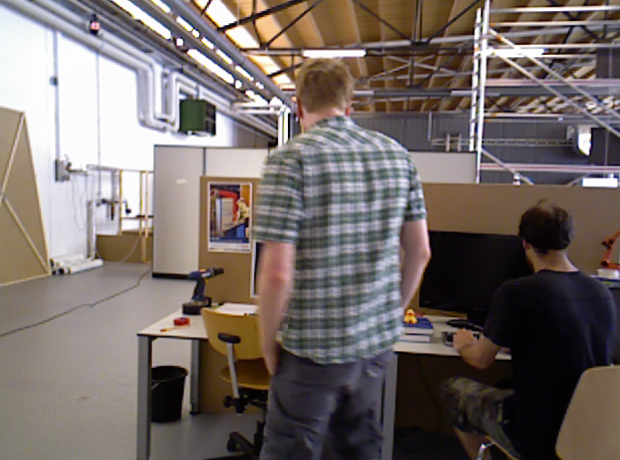} 
    \\
    {Refusion~\cite{palazzolo2019iros}}   & {DynaSLAM (N+G)~\cite{bescos2018dynaslam}} & {MonoGS~\cite{matsuki2024gaussian}} & {Splat-SLAM~\cite{sandstrom2024splat}} & {\textbf{\project{} (\ours)}}  &  {Input} \\
    \multicolumn{2}{c:}{\textbf{\textit{RGB-D input}}} & \multicolumn{3}{c:}{\textbf{\textit{Monocular input}}} & \\ 
     
  \end{tabular} 
  \vspace{-2mm}
  \caption{\textbf{\textit{Input} View Synthesis Results on TUM RGB-D Dataset~\cite{sturm2012benchmark}.} We show results on the \texttt{freiburg3\_walking\_static} (first row) and \texttt{freiburg3\_walking\_xyz} (second row) sequences. Our method produces substantially better rendering results.}
  
  \label{fig:tum_rendering}
\end{figure*} 

\begin{table*}[t]
\centering
\footnotesize
{
\begin{tabular}{lrrrrrrrrrr}
\toprule

Method & \texttt{f2/dp} & \texttt{f3/ss} & \texttt{fr3/sx} & \texttt{f3/sr} & \texttt{f3/shs} & \texttt{f3/ws} & \texttt{f3/wx} & \texttt{f3/wr} & \texttt{f3/whs} & Avg.\\
\midrule
\multicolumn{11}{l}{\cellcolor[HTML]{EEEEEE}{\textit{RGB-D}}} \\

Refusion~\cite{palazzolo2019iros}& 4.9* &  \rd 0.9 &4.0 & 13.2* & 11.0 & 1.7 & 9.9 & 40.6* & 10.4 & 10.73*\\
ORB-SLAM2~\cite{Mur2017TRO}& \fs 0.6 &  \nd 0.8 &  \rd 1.0 & 2.5 & 2.5 & 40.8 & 72.2 & 80.5 & 72.3 & 30.4\\
DynaSLAM (N+G)~\cite{bescos2018dynaslam}&  \nd 0.7* & 0.5* &1.5 & 2.7* & \rd 1.7 & \nd 0.6 & \rd  1.5 &  \nd 3.5 & 2.5& 1.7*\\
NICE-SLAM~\cite{Zhu2022CVPR} & 88.8
 & 1.6 & 32.0 & 59.1 & 8.6 & 79.8 & 86.5 & 244.0 & 152.0 & 83.6\\
DG-SLAM~\cite{xu2024dgslam} & 3.2 & -& \rd 1.0 & - & - & \nd 0.6 & 1.6 & 4.3 & - & - \\
RoDyn-SLAM~\cite{jiang2024rodyn}& -&-&-&-& 4.4 & 1.7&8.3&-&5.6&-\\
DDN-SLAM (RGB-D)~\cite{li2024ddn}& - & -& \rd 1.0&-& \rd 1.7&1.0&  \nd 1.4& \rd 3.9&2.3&-\\

\hdashline
\noalign{\vskip 1pt}
\multicolumn{11}{l}{\cellcolor[HTML]{EEEEEE}{\textit{Monocular}}} \\ 
DSO~\cite{Engel2017PAMI}& 2.2 & 1.7 & 11.5 & 3.7 & 12.4 & 1.5 & 12.9 & 13.8 & 40.7 & 11.1\\
DROID-SLAM~\cite{teed2021droid} & \fs 0.6 & \fs 0.5 &  \nd  0.9 & 2.2 & \fs 1.4 & 1.2 & 1.6 & 4.0 & 2.2 & \nd 1.62\\
MonoGS~\cite{matsuki2024gaussian}& 112.8 & 1.2 & 6.1 & 5.1 & 28.3 & 1.1 & 21.5 & 17.4 & 44.2 & 26.4\\
Splat-SLAM~\cite{sandstrom2024splat}& \nd  0.7 & \fs 0.5 & \nd   0.9 &  \nd 2.3 &  \nd  1.5 &  2.3 & \fs 1.3 & \rd 3.9 & 2.2 & 1.71\\
DynaMoN (MS)~\cite{schischka2023dynamon}   & \fs 0.6 & \fs 0.5 &  \nd  0.9 & \fs 2.1 & 1.9 & 1.4 &  \nd 1.4 & \rd 3.9 & \rd 2.0 & \rd 1.63\\
DynaMoN (MS\&SS) ~\cite{schischka2023dynamon} & \nd 0.7 & \fs 0.5 &  \nd  0.9 &  \rd 2.4 & 2.3 &  \rd 0.7 &  \nd 1.4 &  \rd 3.9 &  \nd 1.9 & \rd 1.63\\
DDN-SLAM (RGB)~\cite{li2024ddn}& - & -&1.3 &-&3.1&2.5&2.8&8.9&4.1&-\\
MonST3R-SW~\cite{zhang2024monst3r} & 51.6 & 2.4 & 28.2 & 5.4 & 36.5 & 2.2 & 27.3 & 13.6 & 19.8 & 20.8 \\
{\textbf{\project{} (\ours)}} &  \rd 1.4 & \fs 0.5 & \fs 0.8 &  \rd 2.4 & 2.0 & \fs 0.4 &\fs  1.3 & \fs 3.3 &\fs  1.6 & \fs 1.51\\ 
\bottomrule
\end{tabular}
}
\caption{\textbf{Tracking Performance on TUM RGB-D Dataset~\cite{sturm2012benchmark}} (ATE RMSE $\downarrow$ [cm]). Best results are highlighted as\colorbox{colorFst}{\bf first},\colorbox{colorSnd}{second}, and\colorbox{colorTrd}{third}.
For methods without complete scene coverage in the original reports, results obtained by running their open-source code are marked with `*'. If open-source code is unavailable, scenes without results are marked with `-'.
DynaSLAM (RGB)~\cite{bescos2018dynaslam} consistently fails to initialize or experiences extended tracking loss across all sequences and therefore cannot be included in this table.
} 
\label{tab:tum_tracking_full}
\end{table*}
 
\begin{figure*}[t!]
  \centering
  \footnotesize
  \setlength{\tabcolsep}{1.5pt}
  \newcommand{\sz}{0.155}
  \newcommand{\sza}{0.18} %
  \begin{tabular}{ccccccc}

  \raisebox{2.5\normalbaselineskip}[0pt][0pt]{\rotatebox[origin=c]{90}{Shopping}} &
  \includegraphics[width=\sz\linewidth]{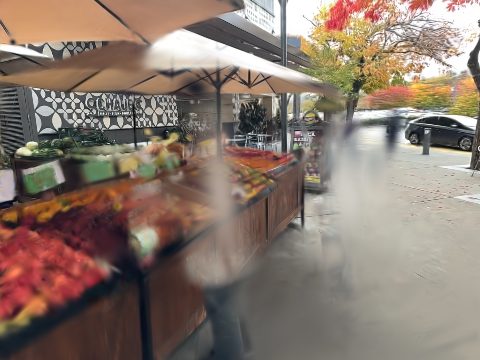} &
  \includegraphics[width=\sz\linewidth]{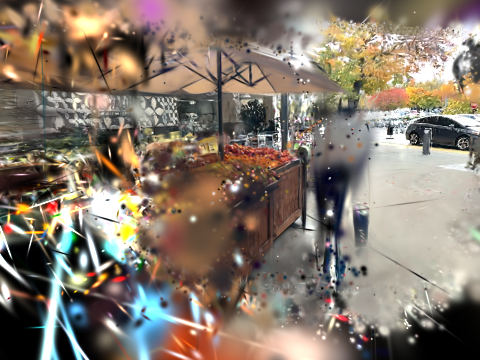} &
  \includegraphics[width=\sz\linewidth]{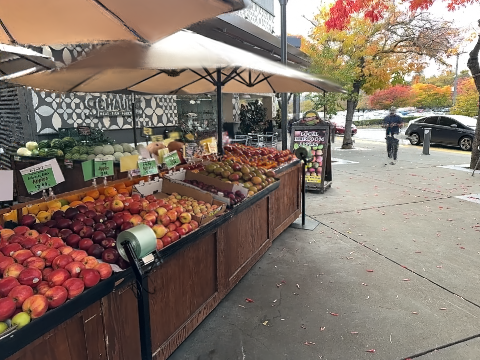} &
  \includegraphics[width=\sz\linewidth]{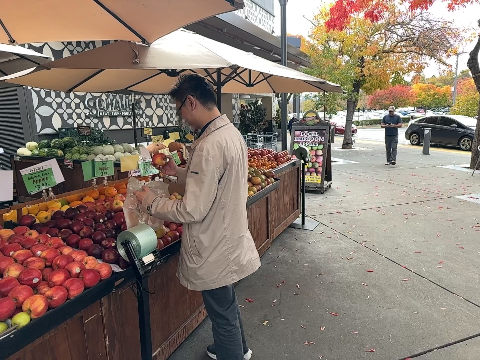} &
  \includegraphics[width=\sz\linewidth]{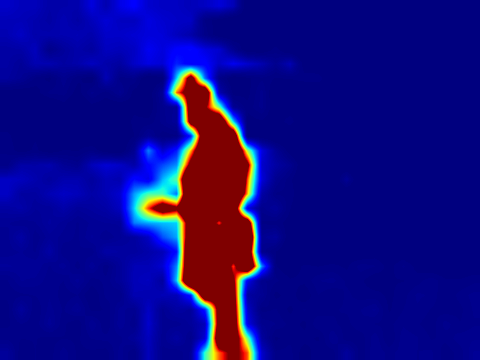}  &
  \includegraphics[width=\sz\linewidth]{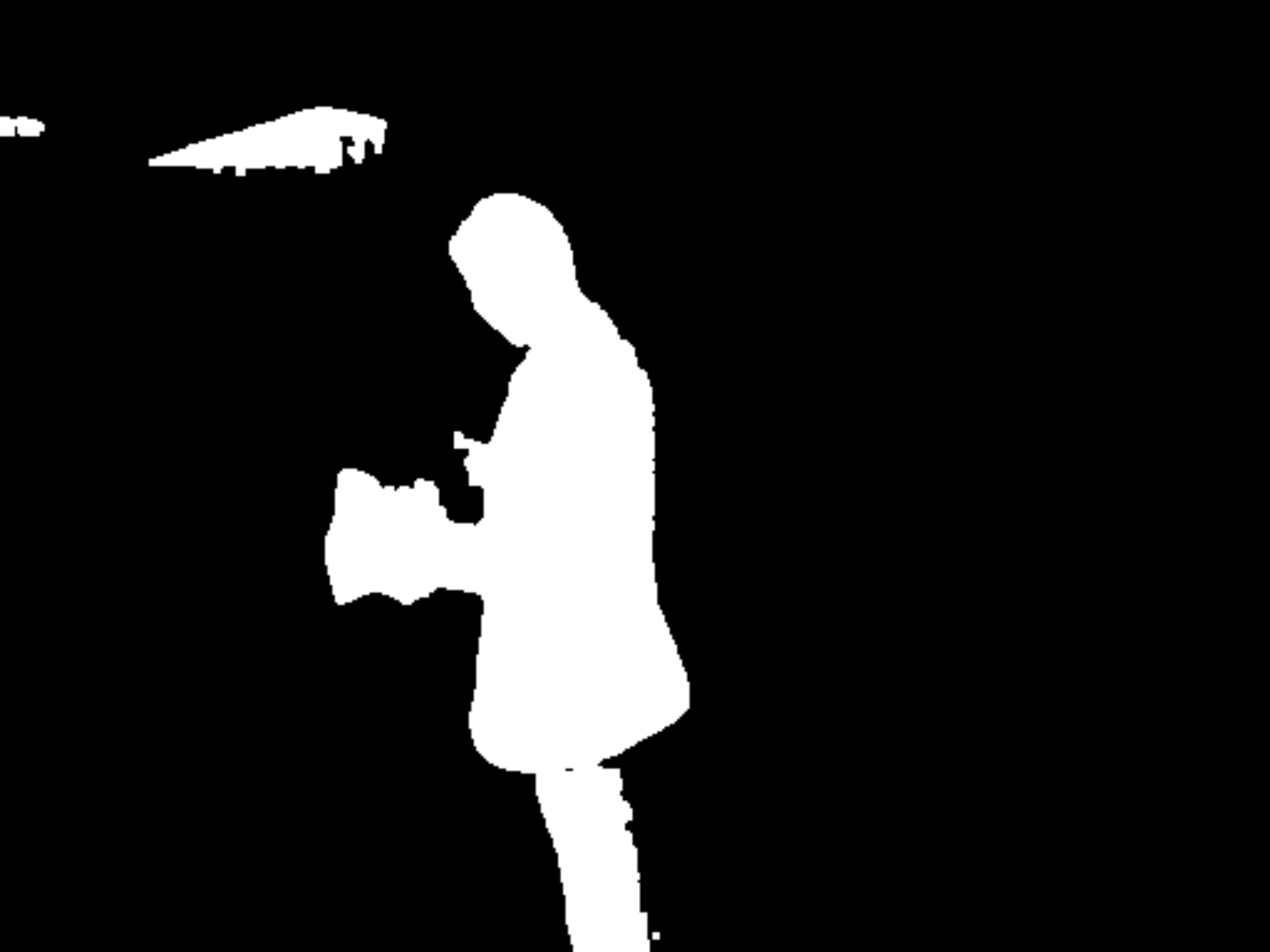} 
  \\
  \raisebox{2.5\normalbaselineskip}[0pt][0pt]{\rotatebox[origin=c]{90}{Wandering}} &
  \includegraphics[width=\sz\linewidth]{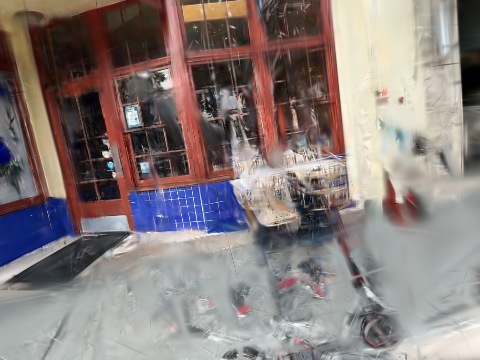} &
  \includegraphics[width=\sz\linewidth]{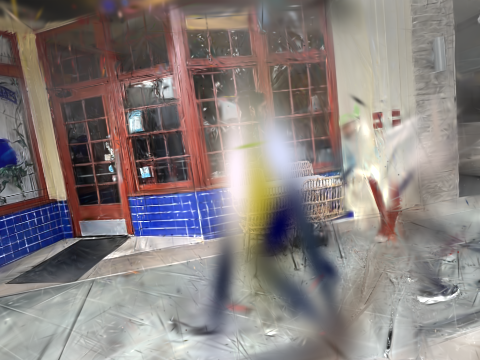} &
  \includegraphics[width=\sz\linewidth]{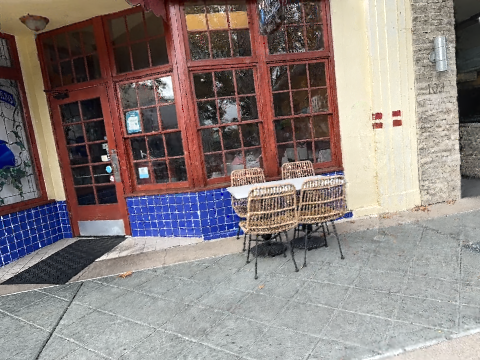} &
  \includegraphics[width=\sz\linewidth]{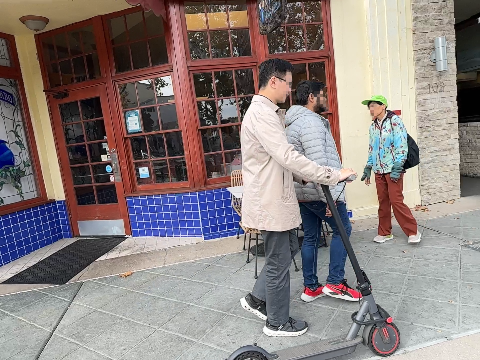} &
  \includegraphics[width=\sz\linewidth]{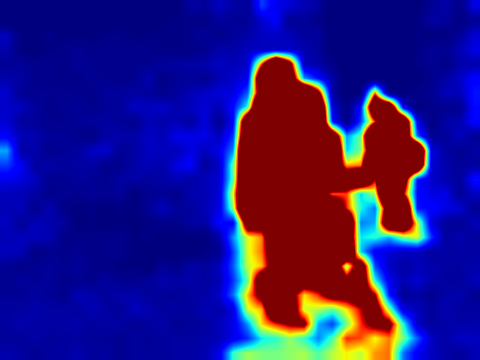}  &
  \includegraphics[width=\sz\linewidth]{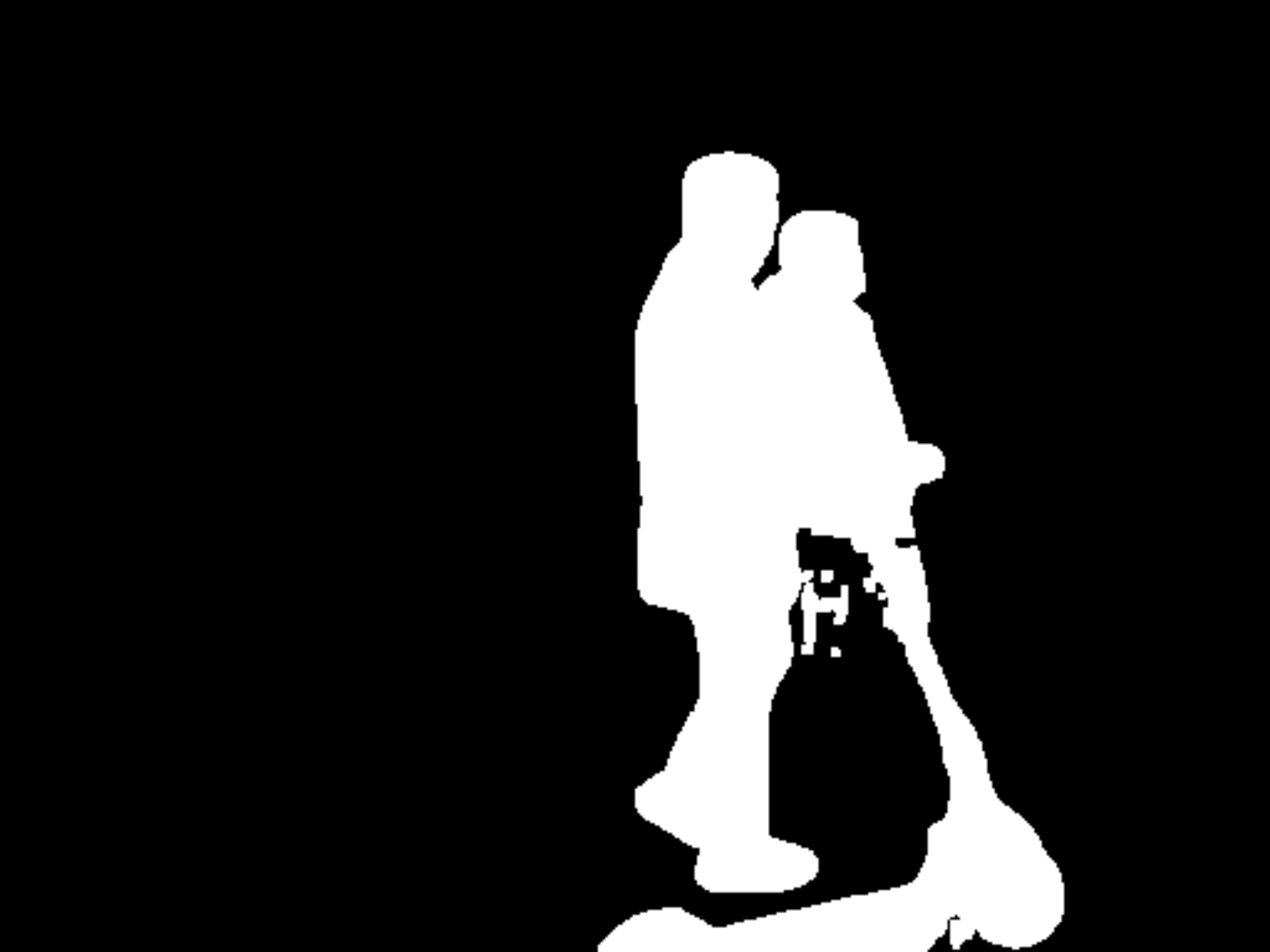} 
  \\
  \raisebox{2.5\normalbaselineskip}[0pt][0pt]{\rotatebox[origin=c]{90}{Wall}} &
  \includegraphics[width=\sz\linewidth]{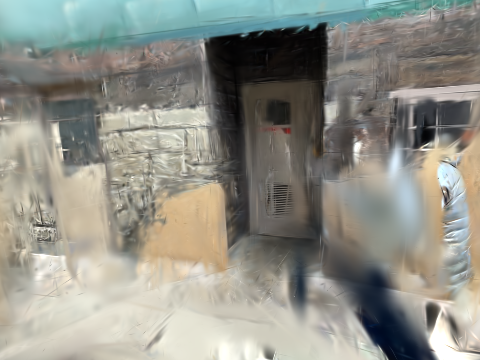} &
  \includegraphics[width=\sz\linewidth]{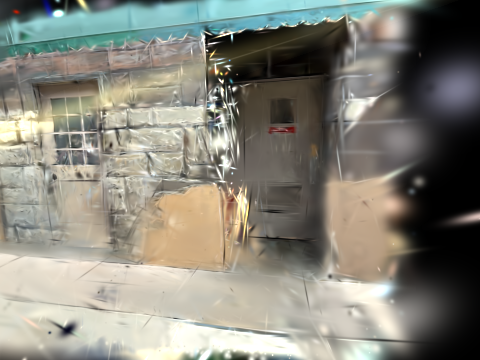} &
  \includegraphics[width=\sz\linewidth]{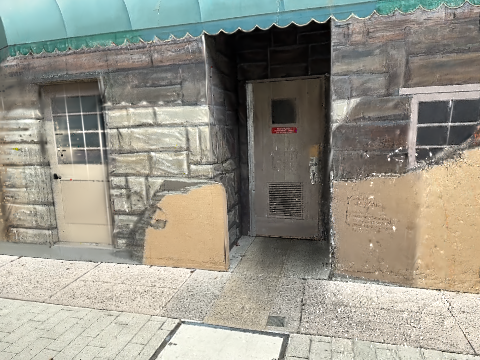} &
  \includegraphics[width=\sz\linewidth]{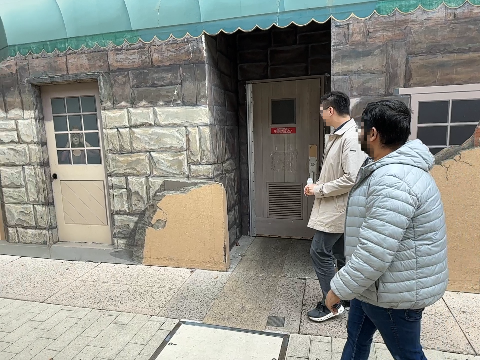} &
  \includegraphics[width=\sz\linewidth]{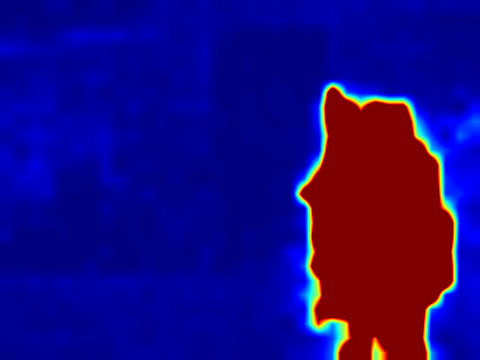}  &
  \includegraphics[width=\sz\linewidth]{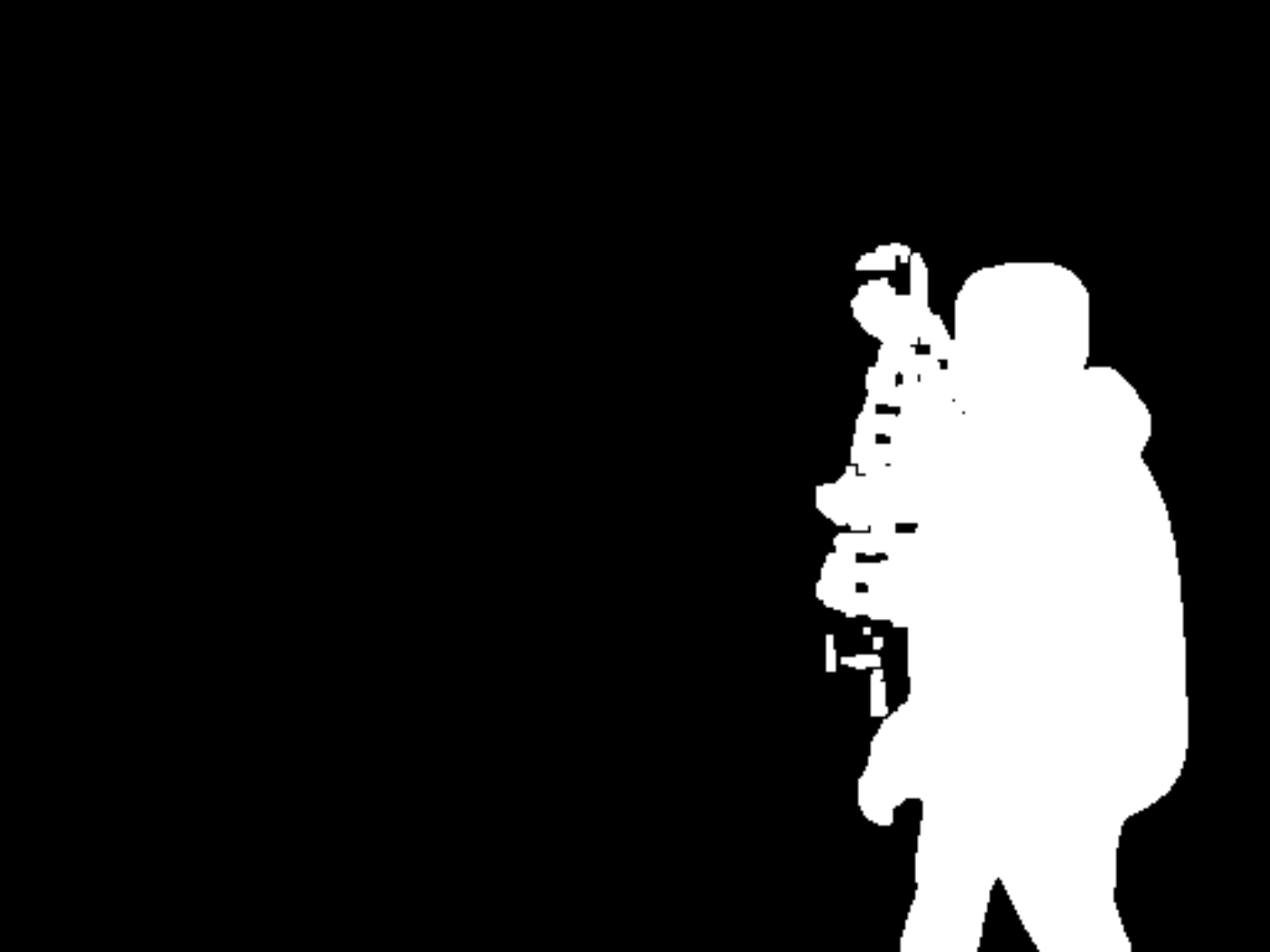} 
  \\
  &  {MonoGS~\cite{matsuki2024gaussian}} & {Splat-SLAM~\cite{sandstrom2024splat}} & {\textbf{\project{} (\ours)}}  &  {Input} & {Uncertainty $\beta$ (\textbf{\ours})} & {MonST3R~\cite{zhang2024monst3r} Mask }\\
     
  \end{tabular}  
 \vspace{-2mm}
  \caption{\textbf{Additional \textit{Input} View Synthesis Results on our Wild-SLAM iPhone Dataset.} 
  Faces are blurred to ensure anonymity.
  }
  \label{fig:iphone_rendering_more}
\vspace{-5pt}
\end{figure*}

\subsection{Baseline Details}

The characteristics of all baseline methods are presented in \tabref{tab:baseline_methods}.
Here we detail the source of each baseline method's results tabulated in the main paper and include the implementation details of MonST3R-SW (our sliding window extension of MonST3R~\cite{zhang2024monst3r}).

\paragraph{Details for \tabref{tab:bonn_tracking}}
For tracking performance on the Bonn RGB-D Dynamic Dataset~\cite{palazzolo2019iros}, results for ORB-SLAM2~\cite{Mur2017TRO}, NICE-SLAM~\cite{Zhu2022CVPR}, and DDN-SLAM~\cite{li2024ddn} are taken from the DDN-SLAM~\cite{li2024ddn} paper.  Results for DROID-SLAM~\cite{teed2021droid} are taken from the DynaMoN~\cite{schischka2023dynamon} paper. Results for DynaSLAM (N+G)~\cite{bescos2018dynaslam} and ReFusion~\cite{palazzolo2019iros} are taken from the ReFusion~\cite{palazzolo2019iros} paper. DG-SLAM~\cite{xu2024dgslam}, RoDyn-SLAM~\cite{jiang2024rodyn}, and DynaMoN~\cite{schischka2023dynamon} are not open-sourced by the time of submission, therefore we take results from their own paper. The results for DSO~\cite{Engel2017PAMI}, MonoGS~\cite{matsuki2024gaussian}, Splat-SLAM~\cite{sandstrom2024splat}, and MonST3R-SW~\cite{zhang2024monst3r} are obtained by running their open-source implementation. 

\paragraph{Details for \tabref{tab:tum_tracking}}
For tracking performance on the TUM RGB-D Dataset~\cite{sturm2012benchmark}, results for Refusion~\cite{palazzolo2019iros}, DG-SLAM~\cite{xu2024dgslam}, DynaSLAM (N+G)~\cite{bescos2018dynaslam}, RoDyn-SLAM~\cite{jiang2024rodyn}, and DDN-SLAM~\cite{li2024ddn} are based on data reported in their respective papers. 
Results for ORB-SLAM2~\cite{Mur2017TRO}, DROID-SLAM~\cite{teed2021droid}, and DynaMoN~\cite{schischka2023dynamon} are sourced from the DynaMoN~\cite{schischka2023dynamon} paper. For DSO~\cite{Engel2017PAMI}, NICE-SLAM~\cite{Zhu2022CVPR}, MonoGS~\cite{matsuki2024gaussian}, Splat-SLAM~\cite{sandstrom2024splat}, and MonST3R-SW~\cite{zhang2024monst3r} results were obtained by running their open-source code. 

\noindent \textbf{MonST3R-SW.} 
High VRAM usage is required for MonST3R~\cite{zhang2024monst3r}, making it impractical to process an entire sequence as input. Instead, we apply a sliding window approach, merging overlapping frames from consecutive windows to form a complete sequence. Specifically, we use a window of 30 frames with a stride of 3, as in the original paper, and maintain an overlap of 25 frames to ensure consistent alignment. We employ Sim(3) Umeyama alignment~\cite{umeyama1991least} to integrate each new window’s trajectory with the global trajectory.

\section{Additional Experiments}
\label{sec:add_results}

\paragraph{Time Analysis}
\tabref{tab:time_analysis} presents the average fps of our method and the baselines.
We also provide a fast version to support more efficient processing with minimal loss of accuracy by disabling low-impact processes and reducing iterations.
To be more specific, the modifications involve (i) removing the calculation of disparity regularization mask; (ii) optimizing the map $\mathcal{G}$ and the uncertainty MLP $\mathcal{P}$ every 5 keyframes; (iii) skipping the refinement of non-keyframe pose via re-rendering loss; (iv) decrease the number of iterations of final map refinement to 3000.
As shown in \tabref{tab:time_analysis}, the fast version still outperforms baselines by a clear margin with comparable runtime.

\begin{table}[h]
    \centering
    \vspace{-10pt}
    \footnotesize
    \newcommand{\sz}{0.195}
    \newcommand{\sza}{0.1729} 
    \setlength{\tabcolsep}{3pt}
    \resizebox{0.98\linewidth}{!}{
    \begin{tabular}{lrrrrrrrrr}
    \toprule
    \multirow{2}{*}{Dataset} & \multicolumn{2}{c}{MonoGS~\cite{matsuki2024gaussian}}  & \multicolumn{2}{c}{Splat-SLAM~\cite{sandstrom2024splat}} & \multicolumn{2}{c}{\textbf{Ours-full}} & \multicolumn{2}{c}{\textbf{Ours-fast}}\\
    \cmidrule(lr){2-3} \cmidrule(lr){4-5} \cmidrule(lr){6-7} \cmidrule(lr){8-9}
    & FPS $\uparrow$ & ATE $\downarrow$   & FPS $\uparrow$ & ATE $\downarrow$   & FPS $\uparrow$ & ATE $\downarrow$   & FPS $\uparrow$ & ATE $\downarrow$ \\
    \midrule
    Wild-SLAM & 2.41 & 47.99 & 2.44 & 8.71 &0.49 &0.46 & 1.96 & 0.48\\
    Bonn & 2.98 & 22.80 & 1.99  & - & 0.50  &2.31 & 2.13& 2.47\\
    \bottomrule
    \end{tabular}}
    \caption{\textbf{Running time evaluation.} For each dataset, we report the average FPS and RMSE of ATE [cm]. We logged the total running time to process a sequence and compute FPS by dividing the total number of processed frames by the total running time. \textit{Ours-full} is the full pipeline presented, while \textit{Ours-fast} is a fast version of \project{}.}
    \label{tab:time_analysis}
    \vspace{-12pt}
\end{table}

\paragraph{Rendering Results on TUM RGB-D Dataset~\cite{sturm2012benchmark}}
Our method effectively removes distractors, as illustrated in \figref{fig:tum_rendering}. ReFusion~\cite{palazzolo2019iros} struggles to fully eliminate distractors, leading to the presence of multiple ghosting artifacts. DynaSLAM (N+G)~\cite{bescos2018dynaslam} exhibits "black holes" due to insufficient multiview information for effective inpainting, while the regions it does manage to inpaint often suffer from noticeable whitish artifacts. MonoGS~\cite{matsuki2024gaussian} and Splat-SLAM~\cite{sandstrom2024splat} exhibit blurry and floating artifacts as they do not explicitly address dynamic environments.

\paragraph{Full Tracking Results on the TUM RGB-D Dataset~\cite{sturm2012benchmark}} We report our performance on the full TUM RGB-D Dataset~\cite{sturm2012benchmark} dynamic sequences in \tabref{tab:tum_tracking_full}. Our method performs the best on average. 

\paragraph{High Resolution Uncertainty Map}
\begin{figure}[t!]
  \centering
  \footnotesize
  \setlength{\tabcolsep}{0.5pt}
  \newcommand{\sz}{0.32}
  \newcommand{\sza}{0.18} %
  \begin{tabular}{cccc}
  \raisebox{2.5\normalbaselineskip}[0pt][0pt]{\rotatebox[origin=c]{90}{Parking}} &
  \includegraphics[width=\sz\linewidth]{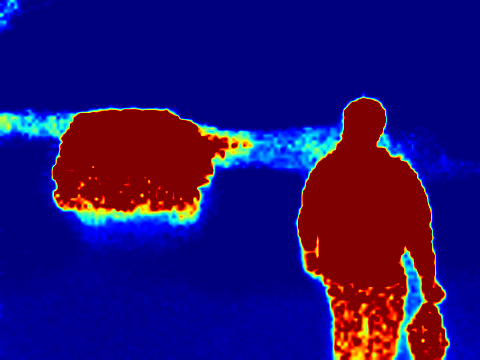} &
  \includegraphics[width=\sz\linewidth]{figs/outdoor/parking2/uncertainty.png} &
  \includegraphics[width=\sz\linewidth]{figs/outdoor/parking2/frame_00184.png}
  \\
  \raisebox{2.5\normalbaselineskip}[0pt][0pt]{\rotatebox[origin=c]{90}{Piano}} &
  \includegraphics[width=\sz\linewidth]{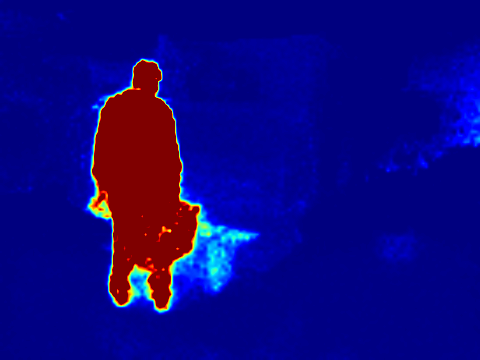} &
  \includegraphics[width=\sz\linewidth]{figs/outdoor/piano/uncertainty.png} &
  \includegraphics[width=\sz\linewidth]{figs/outdoor/piano/frame_00090.png}

  \\

  \raisebox{2.5\normalbaselineskip}[0pt][0pt]{\rotatebox[origin=c]{90}{Street}} &
  \includegraphics[width=\sz\linewidth]{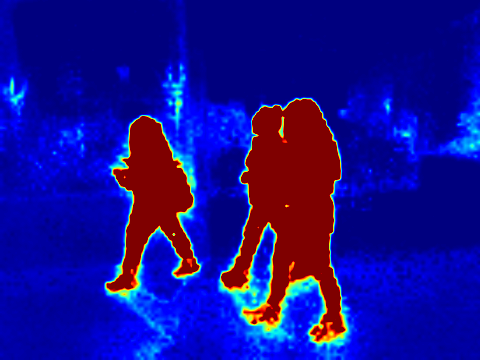} &
  \includegraphics[width=\sz\linewidth]{figs/outdoor/street1/uncertainty.png} &
  \includegraphics[width=\sz\linewidth]{figs/outdoor/street1/frame_00864_mosaic.png}

  \\
  
  &   {High Res. Uncer. Map} & {Uncer. Map} & {Input}\\
     
  \end{tabular}  
 \vspace{-2mm}
  \caption{\textbf{High Resolution Uncertainty Map.}
  }
  \label{fig:high_res_uncer}
\vspace{-15pt}
\end{figure} 
 In \figref{fig:high_res_uncer}, we present the visualization of high-resolution maps, as referenced in \figref{fig:iphone_rendering}. Achieving higher resolution and sharper uncertainty maps is possible, though it comes at the cost of computational efficiency.

\paragraph{More Results on our Wild-SLAM iPhone Dataset}
In addition to the results shown in \figref{fig:iphone_rendering} of the main paper, we provide additional results in \figref{fig:iphone_rendering_more}.

\begin{figure}[h!]
    \centering
    \begin{subfigure}[b]{0.11\textwidth}
        \includegraphics[width=\textwidth]{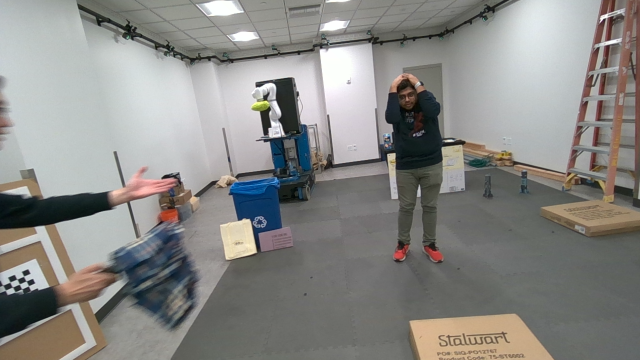}
        \captionsetup{labelformat=empty, font=scriptsize}
        \caption{RGB (F. 215)}
    \end{subfigure}
    \begin{subfigure}[b]{0.11\textwidth}
    \includegraphics[width=\textwidth]{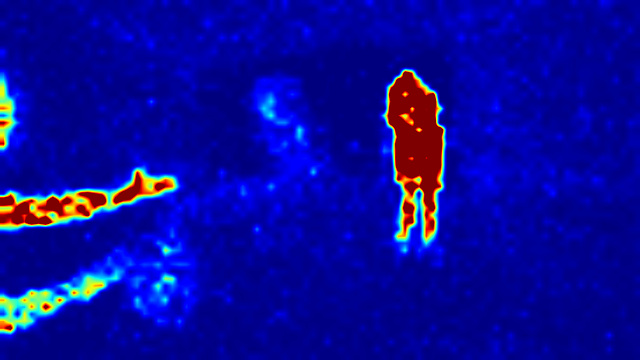}
        \captionsetup{labelformat=empty, font=scriptsize}
        \caption{Trained until F. 80}
    \end{subfigure}
    \begin{subfigure}[b]{0.11\textwidth}
        \includegraphics[width=\textwidth]{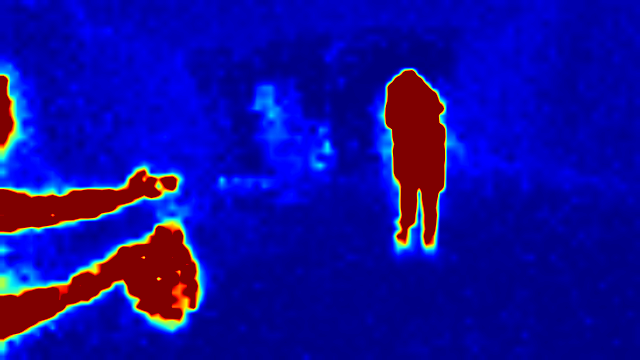}
        \captionsetup{labelformat=empty, font=scriptsize}
        \caption{Trained until F. 215}
    \end{subfigure}
    \begin{subfigure}[b]{0.11\textwidth}
        \includegraphics[width=\textwidth]{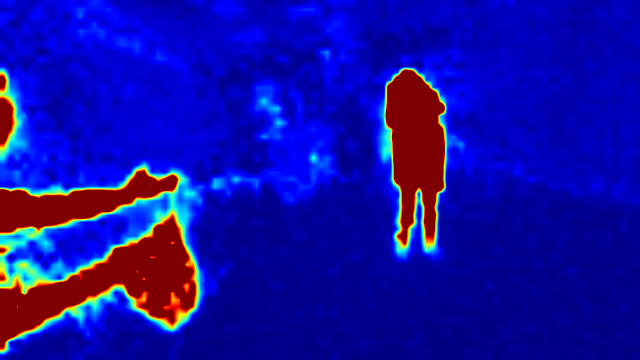}
        \captionsetup{labelformat=empty, font=scriptsize}
        \caption{Trained until F. 451}
    \end{subfigure}
    \vspace{-7.5pt}
    \caption{\textbf{Online Uncertainty Prediction.}}
    \label{fig:online}
    \vspace{-10pt}
\end{figure}

\paragraph{Online Uncertainty Prediction}
We visualize the online uncertainty prediction for frame 215---with MLP trained before, during, and after the umbrella enters the scene---in \figref{fig:online}.
Before (trained until frame 80), the MLP mainly classifies the moving human as a moving distractor since it has never seen the umbrella in the first 80 frames. As the umbrella enters the scene (frame 215), our uncertainty prediction module rapidly identifies it as a moving distractor due to the inconsistency between the Gaussian map and the frame 215. Moreover, the uncertainty estimate stabilizes shortly afterward (frame 451). 

\vspace{1pt} \noindent
\textbf{Pure Static Sequences.}
To demonstrate the robustness of our method, we also evaluate it on static sequences from the TUM RGB-D Dataset~\cite{sturm2012benchmark}, shown in \tabref{tab:tum_static_tracking}. Our approach performs on par with state-of-the-art monocular Gaussian Splatting SLAM methods, such as MonoGS~\cite{matsuki2024gaussian} and Splat-SLAM~\cite{sandstrom2024splat}.

\begin{table}[t]
  \centering
  \footnotesize
  \setlength{\tabcolsep}{1.6pt}          %
  \renewcommand{\arraystretch}{1.05}     %
    \resizebox{\linewidth}{!}{
  \begin{tabular}{@{\hspace*{1em}}lrrrrr}
    \toprule
    & \tt{fr1/desk} &  \tt{fr2/xyz} &  \tt{fr3/off} & Avg.\\
    \midrule
    \multicolumn{5}{l}{\cellcolor[HTML]{EEEEEE}{\textit{RGB-D}}} \\ 
    {ORB-SLAM2~\cite{Mur2017TRO}} & \fs 1.6  & \rd 0.4  & \fs 1.0 & \fs 1.0 \\ 
    {NICE-SLAM}~\cite{Zhu2022CVPR}           & 2.7 & 1.8 & 3.0 & 2.5\\
    \hdashline
    \noalign{\vskip 1pt}
    \multicolumn{5}{l}{\cellcolor[HTML]{EEEEEE}{\textit{Monocular}}} \\ 
    {DROID-SLAM~\cite{teed2021droid}} & \rd 1.8 & 0.5 & \rd 2.8 & \rd 1.7\\{MonoGS}~\cite{matsuki2024gaussian} & 3.8 & 5.2 & 2.9& 4.0 \\
    
    Splat-SLAM~\cite{sandstrom2024splat} & \fs 1.6 & \fs 0.2 & \nd 1.4 & \nd 1.1 \\
    {\textbf{\project{} (\ours)}} & \nd 1.7 & \nd 0.3& \nd 1.4 & \nd 1.1  \\
    \bottomrule
  \end{tabular}}
  \caption{\textbf{Tracking Performance on TUM RGB-D Dataset (Static)~\cite{sturm2012benchmark}} (ATE RMSE $\downarrow$ [cm]). Best results are highlighted as\colorbox{colorFst}{\bf first},\colorbox{colorSnd}{second}, and\colorbox{colorTrd}{third}.
  Results for {ORB-SLAM2~\cite{Mur2017TRO}} and {NICE-SLAM}~\cite{Zhu2022CVPR} are taken from NICE-SLAM~\cite{Zhu2022CVPR}. Results for {MonoGS}~\cite{matsuki2024gaussian} and Splat-SLAM~\cite{sandstrom2024splat} are taken from Splat-SLAM~\cite{sandstrom2024splat}. The results for {DROID-SLAM~\cite{teed2021droid}} are obtained by running their open-source code.
  }
    \label{tab:tum_static_tracking}
  \vspace{-12pt}
\end{table}

\vspace{1pt} \noindent
\textbf{Ablation Study on Disparity Regularization.}
\begin{table*}[t]
    \centering
    \footnotesize
    \newcommand{\sz}{0.195}
    \newcommand{\sza}{0.1729} 
    \setlength{\tabcolsep}{3pt}
    \resizebox{0.8\linewidth}{!}{
    \begin{tabular}{cccrrrrrr}
    \toprule
\multirow{2}{*}{\makecell{ }}& \multirow{2}{*}{\makecell{Disp. Reg. \\ Mask $M$}} & \multirow{2}{*}{\makecell{No Disp. Reg. \\ in Final Global BA}} & \multicolumn{2}{c}{\texttt{Wild-SLAM}} & \multicolumn{2}{c}{\texttt{Bonn}} & \multicolumn{2}{c}{\texttt{TUM}} \\
\cmidrule(lr){4-5} \cmidrule(lr){6-7} \cmidrule(lr){8-9}
& & &Before BA & After BA & Before BA & After BA & Before BA & After BA \\
\midrule
\textit{(i)} & \xmark & \xmark  &3.12 & 1.95 &4.34 & 2.56&2.17 &1.86\\
\textit{(ii)} & \cmark & \xmark  &\textbf{2.90} & 1.57 & 3.97& 2.56& \textbf{1.92} &1.69\\
\textit{(iii)} & \xmark & \cmark & 3.17 & \textbf{0.46} & 4.40 & 2.47 &2.16 & \textbf{1.55} \\
\textit{(iv)} & \cmark & \cmark  & 2.92 & \textbf{0.46} &\textbf{ 3.89} & \textbf{2.31} & 1.94 & 1.63 \\ 
\bottomrule
\end{tabular}}
\vspace{-2mm}
\caption{
\textbf{Ablation Study on Disparity Regularization} (ATE RMSE $\downarrow$ [cm]). For each dataset, we report the average tracking error before and after the final global BA. `Before BA' denotes before final global BA. `After BA' denotes after final global BA. 
}
\label{tab:ablation_depth_mask}
\vspace{-5pt}
\end{table*}

\tabref{tab:ablation_depth_mask} presents an ablation study evaluating the effects of (a) the disparity regularization mask $M$ used in DBA (\eqnref{eq:dba}) during on-the-fly capture and (b) the exclusion of the disparity regularization term in the final global BA.
Removing $M$ \textit{(rows iii and iv)} has minimal impact on the final global BA, as shown in the `After BA' results. However, the `Before BA' results are significantly degraded, highlighting the multi-view inconsistencies in monocular predictions.
Excluding the disparity regularization term in the final global BA \textit{(rows ii and iv)} has no effect on the `Before BA' results (minor deviations are expected due to randomness and initialization) but leads to improved `After BA' performance. This improvement is attributed to the availability of multiple views in the final global BA, which refines depth accuracy compared to monocular predictions. The best results are achieved when $M$ is applied and the disparity regularization term is excluded during the final global BA \textit{(row iv)}, validating our design choices.

\vspace{1pt} \noindent
\textbf{Ablation Study on Distractor Estimation.}
\begin{table}[t]
    \centering
    \footnotesize
    \newcommand{\sz}{0.195}
    \newcommand{\sza}{0.1729} 
    \setlength{\tabcolsep}{3pt}
    \resizebox{0.8\linewidth}{!}{
    \begin{tabular}{lrrrr}
    \toprule
& \texttt{Wild-SLAM} & \texttt{Bonn} & \texttt{TUM} & \\
\midrule
MonST3R Mask  & 2.60 & 2.58 & 1.80 \\
YOLOv8 + SAM Mask &  3.06 & 2.37 &  1.65\\
{\textbf{\project{} (\ours)}}  & \textbf{0.46} & \textbf{2.31} &  \textbf{1.63} \\ 
\bottomrule
\end{tabular}}
\vspace{-2mm}
\caption{\textbf{Ablation Study on Distractor Estimation.} (ATE RMSE $\downarrow$ [cm]). For each dataset, we report the average tracking error. 
}
\label{tab:ablation_distractor_estimation}
\vspace{-5pt}
\end{table}

We compare various distractor estimation methods and utilize their resulting distractor masks for tracking in WildGS-SLAM. For the MonST3R mask, we aggregate masks from multiple runs because MonST3R supports only a limited number of images per run. The YOLOv8 + SAM mask corresponds to (c) in \tabref{tab:ablation}; we include it here for a clearer comparison. As shown in \tabref{tab:ablation_distractor_estimation}, our method consistently outperforms others, as other approaches struggle to produce accurate enough masks, particularly on our Wild-SLAM dataset, which features diverse and complex distractors.

\vspace{1pt} \noindent
\textbf{Ablation Study on Pretrained Models.}
\begin{table}[t]
    \centering
    \footnotesize
    \newcommand{\sz}{0.195}
    \newcommand{\sza}{0.1729} 
    \setlength{\tabcolsep}{3pt}
    \resizebox{\linewidth}{!}{
    \begin{tabular}{llrrrrr}
    \toprule
    \multirow{2}{*}{Depth Estimator} & \multirow{2}{*}{DINOv2 model} & \multicolumn{2}{c}{\texttt{Wild-SLAM}} & \texttt{Bonn} & \texttt{TUM} \\
    \cmidrule(lr){3-4} \cmidrule(lr){5-5} \cmidrule(lr){6-6}
    & & PSNR $\uparrow$ & ATE $\downarrow$ & ATE $\downarrow$ & ATE $\downarrow$ \\
    \midrule
    DPTv2~\cite{yang2024depth} & Original~\cite{oquab2023dinov2} & 20.56 & 0.47 & 2.36 & 1.76 \\
    DPTv2~\cite{yang2024depth} & Finetuned\cite{yue2025improving}& 20.57 & 0.47 & 2.41 & 1.66 \\ 
    Metric3D V2~\cite{hu2024metric3d} & Original~\cite{oquab2023dinov2} & \textbf{20.58} & 0.52 & \textbf{2.31} & \textbf{1.61} \\
    Metric3D V2~\cite{hu2024metric3d} & Finetuned\cite{yue2025improving}& \textbf{20.58} & \textbf{0.46} & \textbf{2.31} & 1.63 \\ 
    \bottomrule
    \end{tabular}}
    \vspace{-2mm}
    \caption{\textbf{Ablation Study on Different Pretrained Models}. For the Wild-SLAM dataset, we report the novel view synthesis results (PSNR $\uparrow$) and the tracking error (ATE RMSE $\downarrow$ [cm]). For the Bonn and TUM datasets, we report the average tracking error (ATE RMSE $\downarrow$ [cm]).}
    \label{tab:ablation_pretrained_model}
    \vspace{-5pt}
\end{table}

We conduct an ablation study on the pretrained models, namely the depth estimator and the feature exactor DINOv2 model, as presented in \tabref{tab:ablation_pretrained_model}. Both novel view synthesis and tracking evaluations confirm that Metric3D V2~\cite{hu2024metric3d}, combined with the finetuned DINOv2 model~\cite{yue2025improving}, achieves the best overall performance, validating our design choices.

\paragraph{Failure Cases}
\begin{figure}[t!]
  \centering
  \footnotesize
  \setlength{\tabcolsep}{0.5pt}
  \newcommand{\sz}{0.32}
  \newcommand{\sza}{0.18} %
  \begin{tabular}{cccc}
  \raisebox{2.5\normalbaselineskip}[0pt][0pt]{\rotatebox[origin=c]{90}{Shopping}} &
  \includegraphics[width=\sz\linewidth]{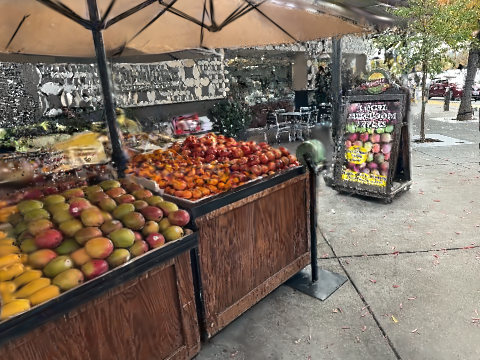} &
  \includegraphics[width=\sz\linewidth]{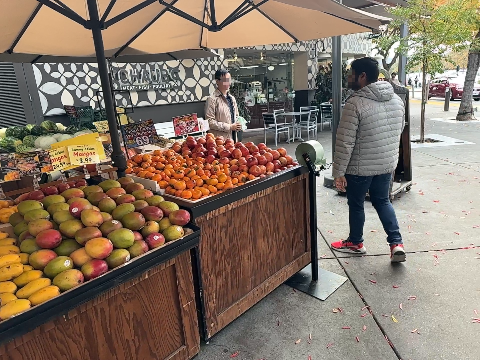} &
  \includegraphics[width=\sz\linewidth]{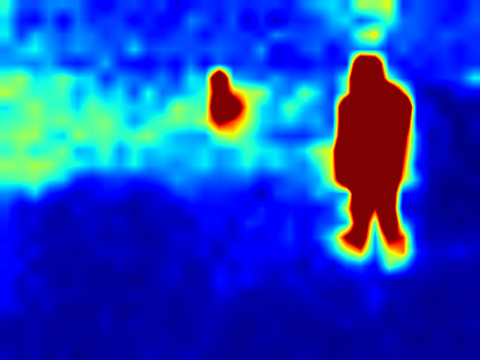} 
  \\
  \raisebox{2.5\normalbaselineskip}[0pt][0pt]{\rotatebox[origin=c]{90}{Wandering}} &
  \includegraphics[width=\sz\linewidth]{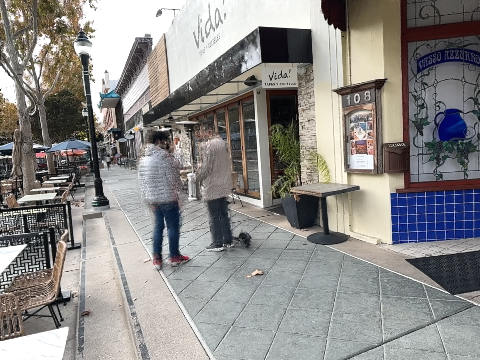} &
  \includegraphics[width=\sz\linewidth]{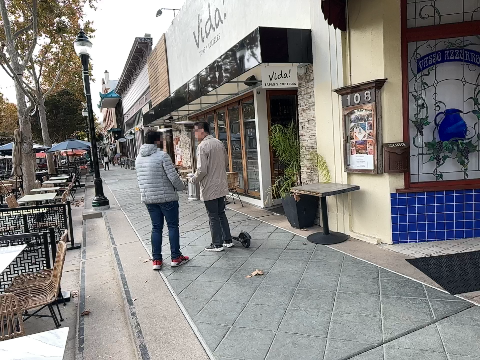} &
  \includegraphics[width=\sz\linewidth]{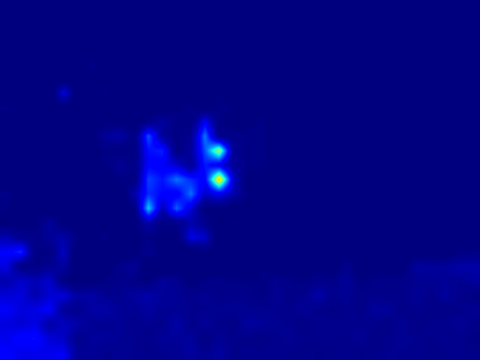} 
  \\
  &   {\textbf{\project{} (\ours)}}  &  {Input} & {Uncertainty $\beta$ (\textbf{\ours})} \\
     
  \end{tabular}  
 \vspace{-2mm}
  \caption{\textbf{Failure Cases.} In shopping dataset, patterns on the wall is incorrectly regarded as medium uncertainty because of the difficulty of reconstructing the complicated textures. In wandering, humans are not removed due to the lack of observation of the static scene. Faces are blurred to ensure anonymity.
  }
  \label{fig:failure_case}
\vspace{-5pt}
\end{figure} 

In \figref{fig:failure_case}, we present two failure cases of our method. 
In the first case, while our method successfully removes dynamic objects, it struggles to reconstruct the complex background, leading to a high SSIM loss in \eqnref{eq:uncer_loss}. Therefore, the high SSIM loss drives the uncertainty prediction to incorrectly assign higher uncertainty to static regions.

In the second case, the dynamic objects remain stationary in some of the frames and, since all frames are captured from roughly the same camera direction, no earlier frames are available to observe the static scene without the dynamic objects. As a result, the system assigns lower uncertainty to these regions and mistakenly reconstructs the dynamic objects.

\end{document}